\documentclass[twocolumn,switch]{article}

\usepackage{preprint}

\usepackage[utf8]{inputenc}             
\usepackage[T1]{fontenc}                
\usepackage{url}                        
\usepackage{booktabs}                   
\usepackage{nicefrac}                   
\usepackage{microtype}                  
\usepackage{amsmath, amsthm, amssymb, amsfonts}
\usepackage{widetext}
\usepackage{wrapfig2}
\usepackage{subcaption}
\usepackage{float}
\usepackage{array}
\usepackage{tabularx}
\usepackage[ruled]{algorithm2e}
\usepackage{graphicx}
\usepackage[numbers,square]{natbib}
\usepackage{doi}
\usepackage{titlesec}
\usepackage{hyperref}                   
\usepackage[capitalize,nameinlink]{cleveref}       
\usepackage{cuted}
\usepackage{dblfloatfix}
\usepackage{placeins}

\titlespacing\section{0pt}{12pt plus 3pt minus 3pt}{1pt plus 1pt minus 1pt}
\titlespacing\subsection{0pt}{10pt plus 3pt minus 3pt}{1pt plus 1pt minus 1pt}
\titlespacing\subsubsection{0pt}{8pt plus 3pt minus 3pt}{1pt plus 1pt minus 1pt}

\DeclareMathOperator{\tr}{tr}
\DontPrintSemicolon

\makeatletter
\let\origsubref\subref
\renewcommand{\subref}[1]{\textbf{\origsubref{#1}}}
\makeatother

\crefformat{subfigure}{Figure~#2\thefigure.\alph{subfigure}#3}
\Crefformat{subfigure}{Figure~#2\thefigure.\alph{subfigure}#3}

\title{
Active Transfer Bagging---A New Approach for Accelerated Active Learning 
Acquisition of Data by Combined Transfer Learning and Bagging Based Models
}

\date{}

\usepackage{authblk}

\author[1\thanks{These authors contributed equally}]{Vivienne Pelletier}
\author[2{\protect\footnotemark[1]}]{Daniel J. Rivera}
\author[2]{Obinna Nwokonkwo}
\author[2]{Steven A. Wilson}
\author[1,2\thanks{Corresponding author: \texttt{cmuhich@asu.edu}}]{%
    Christopher L. Muhich
}
\affil[1]{
    Materials Science \& Engineering\\
    School for the Engineering of Matter, Transport, \& Energy\\
    Arizona State University, 551 E. Tyler Mall, Tempe Arizona, 85287, USA
}
\affil[2]{
    Chemical Engineering\\
    School for the Engineering of Matter, Transport, \& Energy\\
    Arizona State University, 551 E. Tyler Mall, Tempe Arizona, 85287, USA
}

\hypersetup{
pdftitle={Active Transfer Bagging},
pdfsubject={},
pdfauthor={Vivienne Pelletier},
pdfkeywords={transfer learning, downselection, coresets, active learning},
}

\makeatletter
\let\orig@footnotetext\footnotetext
\gdef\saved@footnotes{}
\renewcommand\footnotetext[2][]{%
  \g@addto@macro\saved@footnotes{\orig@footnotetext[#1]{#2}}%
}
\makeatother

\begin{document}

\twocolumn[%
\begin{@twocolumnfalse}

\maketitle

\begin{abstract}
Modern machine learning has achieved remarkable success on many problems, but this
success often depends on the existence of large, labeled datasets. While active
learning can dramatically reduce labeling cost when annotations are expensive, early
performance is frequently dominated by the initial seed set, typically chosen at
random. In many applications, however, related or approximate datasets are readily
available and can be leveraged to construct a better seed set. We introduce a new
method for selecting the seed data set for active learning, Active-Transfer Bagging
(ATBagging). ATBagging estimates the informativeness of candidate data point from
a Bayesian interpretation of bagged ensemble models by comparing in-bag and
out-of-bag predictive distributions from the labeled dataset, yielding an
information-gain proxy. To avoid redundant selections, we impose feature-space
diversity by sampling a determinantal point process (DPP) whose kernel uses Random
Fourier Features and a quality-diversity factorization that incorporates the
informativeness scores. This same blended method is used for selection of new data
points to collect during the active learning phase. We evaluate ATBagging on four
real-world datasets covering both target-transfer and feature-shift scenarios (QM9,
ERA5, Forbes 2000, and Beijing PM\textsubscript{2.5}). Across seed sizes
$n_\text{seed} = $ 10--100, ATBagging improves or ties early active learning and
increases area under the learning-curve relative to alternative seed subset
selection methodologies in almost all cases, with strongest benefits in low-data
regimes. Thus, ATBagging provides a low-cost, high reward means to initiating active
learning-based data collection.
\end{abstract}


\vspace{1.0cm}

\end{@twocolumnfalse}
]

\makeatletter
\begingroup
  \renewcommand{\thefootnote}{\fnsymbol{footnote}}
  \renewcommand{\@makefnmark}{\hbox to \z@{$^{\@thefnmark}$\hss}}
  \long\def\@makefntext#1{%
    \parindent 1em\noindent
    \hbox to 1.8em{\hss $\m@th ^{\@thefnmark}$}#1%
  }
  \saved@footnotes 
\endgroup
\makeatother

\section{Introduction}

The reliance of modern machine learning methodologies on large, fully labeled datasets,
i.e. datasets with a target $y_i$ associated with each element of the domain $x_i$,
remains a fundamental bottleneck in their application to new problems. The expense of
dataset construction is primarily incurred in the labelling process, leading to an
abundance of unlabeled datasets \citep{Roh2021, Emam2021} particularly when labelling
requires consultation with experts, e.g. doctors classifying medical imagery, or when
the prediction target is experimentally or computationally demanding to
acquire.\citep{Rdsch2023,Lawley2024,Keith2021,Zhang2021} This expensive data collection
process must be repeated if a different target is desired, even when the underlying
labels are consistent and the problems are related, i.e. different levels of numerical
rigor in a simulation, or a new experimental measurement of the same underlying system.
Ameliorating these obstacles requires a rigorous approach towards dataset efficiency,
ensuring that labelling and learning expenses are incurred on the fewest representative
elements possible.

A common approach to address the expense of data collection is active learning. Active
learning is an iterative process where unlabeled data points are selected for
acquisition following some criterion, e.g. model uncertainty, model disagreement, or
expected change, thereby prioritizing data collection efforts towards the most
informative points\citep{Tharwat2023}. However, the quality and effectiveness of this
approach is limited by the model's ability to approximate the acquisition criterion from
the currently available data set. Thus, the rate of model improvement in the early
stages of active learning is dominated by the information content in the model
initiation, or seed, data.\citep{Dligach2011} The seed data is typically chosen via
uniform random sampling of the unlabeled pool.\citep{Chandra2021,Hu2010} Although
random sampling is straightforward and not biased by human intuition, it carries no
guarantees about the informativeness or representativeness of the resulting seed data
set.

We posit that this early inefficiency may be mitigated by recognizing that there often
already exist labels of a different prediction target of the data set which are
correlated with the target of interest, or at least that there are correlated targets
that are easily acquired.\citep{Rakotonirina2025,Brust2020} In such cases the
information provided by these proxy labels should be able to guide the creation of the
model data set seed by finding a transferable, informative subset of the proxy dataset.
However, this information is rarely, if at all, used.\citep{Dong2022} Therefore, our
objective is to leverage the information inherent in proxy datasets to select a seed
subset for labelling such that a model trained on the subset gains a broad understanding
of the target’s response over the feature space and has well-tuned uncertainty estimates
and thus decreases the data sampling during active learning.

A related concept, a coreset, comes from the field of computational geometry, and is
defined as a weighted subset such that the loss of any model evaluated on the coreset
approximates the loss of the full dataset.\citep{Bachem2015} Coresets are designed to
reproduce a specific loss function generally, for all possible models or model
parameters; our goal, conversely, is to determine an optimal subset for transferring to
a new loss function, such that the predictive performance of a model with optimal
parameters trained on the subset is close to that of the same model trained on a full
dataset. The coreset approach is most useful in settings where a large, fully labelled
dataset exists but training a model on it is prohibitively expensive. Thus, it is not
concerned with the transferability of such a subset to different targets and has not
been applied to such tasks.

There remains a major knowledge gap as to how to transfer information from a known
dataset to a new, related learning task via the creation of an optimal subset to seed
active learning with the new dataset. We call this concept an active transfer learning
task and propose that an effective subset for this purpose should maximize two
properties: informativeness, i.e. the set of points whose inclusion most affects the
model’s predictive distribution, and heterogeneity, i.e. the points which collectively
span the feature space. 

The first of these has previously been addressed by coreset generation methodologies and
is quantifiable via properties such as influence functions,\citep{Koh2020} which
estimates the effect of reweighing a training point on model loss or parameters, and
sensitivity scores,\citep{Braverman2021} which quantify how much a training point could
contribute to the overall loss across potential model parameter sets. Subset
heterogeneity is rarely addressed in the context of dataset pruning, but is often
considered in the context of (batch) active learning, where many approaches aim to
diversify the selected batch by maximizing the determinant of a covariance
matrix,\citep{Kirsch2019} or impose a sampling strategy which iteratively chooses the
points with the most distinct uncertainty representation, such as Largest Cluster
Maximum Distance.\citep{Holzmller2023} However, these diversifying approaches focus on
the model’s predictive uncertainty space rather than the data’s feature space, which we
hypothesize will be of greater importance for a transfer task. 

While these properties have been addressed separately in different contexts, we combine
them into one method that ensures both informative and heterogenous seed subsets for
initializing active learning procedures from a transferred data set. Specifically, this
work introduces the Active Transfer Bagging (ATBagging) method which combines
informativeness scores derived from a Bayesian interpretation of bagged ensemble models
with the heterogeneity enforced by determinantal point processes to determine the
optimal active learning seed data. In experiments with real world datasets, the
ATBagging method is a facile and inexpensive preprocessing step that seeds any active
learning method, improves early model uncertainty estimates and accelerates the rate of
information acquisition during active learning. 

\section{Methods}
\label{sec:methods}

The proposed ATBagging method contains two major components: 1) the quantification of
informativeness and 2) the imposition of subset heterogeneity. In this section we first
describe the target class of problems which ATBagging addresses, then we will discuss
the mathematics of the two components, and finally we will discuss the real-world
datasets and the associated experiments used in benchmarking ATBagging.

\subsection{Problem Statement}

We consider the transfer learning scenario in which one has access to a source dataset,
$\mathcal{D}$, that consists of labeled data points $\mathcal{D}={(x_i,y_i )}_i^N$,
where $x_i\in X$ represents a feature vector drawn from sampling distribution $x\sim
p(X)$ and $y_i\in Y$ is the corresponding label from the target space $Y$. Additionally,
there is a desired target dataset, $\mathcal{D}'$, which consists of not-yet-acquired
new labels $y_i'\in Y'$ that are correlated with the original labels $Y$ but are
expensive or difficult to acquire. The $x_i'$ of $\mathcal{D}'$ may be the same as the $x_i$
of $\mathcal{D}$, or they may be drawn from a different sampling distribution, $x_i'\sim
p'(X)$. Our goal is to construct a small representative subset $\mathcal{S} \subset
\mathcal{D}$, such that when training a model to learn the map $f':X'\to Y'$ using only
$\mathcal{S}$, its predictive performance approximates the performance of the same model
trained on the full dataset $\mathcal{D}'$.

\subsection{Informativeness Scores}

We define data point informativeness as the information gain of the predictive
distribution $p(Y_*|X_*)$ due to observing the datum $(x,y)$,
\[
\text{IG}_{Y,X}(Y,x)=\text{KL}\left(p(Y_*|X_*,x,y)\parallel p(Y_*|X_*)\right)
\]

given by the Kullback-Leibler (KL) divergence\citep{Kullback1951} between the posterior
predictive distribution $p(Y_*|X_*,x,y)$ and the prior predictive distribution
$p(Y_*|X_*)$. This definition is oriented towards the predictive information and is
a similar definition to the EPIG active learning method.\citep{BickfordSmith2023}
$\text{IG}_{Y,X}$ quantifies how much the distribution of the model's predictions,
$p(Y_*|X_*)$, over a set of test inputs, $X_*$, changes upon incorporation of new
observation pairs $(x,y)$ into the model training set. Therefore, $X_*$ should be chosen
to be representative of the $X$ set features on which the model will ultimately be
applied. Simply using the $X$ set features of the transfer problem, or a subset thereof
if it is very large, is a practical approach and we adopt it in this work. However, the
creation of full probabilistic models of these distributions for the inclusion and
exclusion of every data point would be prohibitively expensive. Instead, a powerful
approximation can be found in the Bayesian interpretation of bagged ensemble models. 

A bagged ensemble model is a model $\mathcal{M}=\{m_i\}_i^M$, comprised of $M$ submodels
of weak learners, $m_i$, trained on bootstrapped samples of the training
dataset.\citep{Breiman1996} The utility of bagged ensemble models for our purpose comes
from interpretating them through an approximate Bayesian lens, where the parameters
learned by each submodel $m_i$ are interpreted as Monte Carlo draws from the
distribution of model parameters, $p(\theta)$, and therefore their predictions
$\tilde{y}_*=m_i(x_*)$ are interpreted as draws from the posterior predictive
distribution $\tilde{y}_*\sim p(y_*|x_*,\theta)p(\theta)$. When $M$ is chosen to be
sufficiently large (see \Cref{sec:appendixA}), there exists with very high probability
a partitioning of the submodels into two non-empty subsets whose bootstrapped samples
either do or do not contain the training point $(x_i,y_i)$, called in-bag and out-of-bag
models, respectively, and denoted by $\mathcal{M}_\text{ib}\subset \mathcal{M}$ and
$\mathcal{M}_\text{oob}\subset \mathcal{M}$. Both $\mathcal{M}_\text{ib}$ and
$\mathcal{M}_\text{oob}$ are themselves bagged ensemble models for the datasets
$\mathcal{D}$ and $\mathcal{D}\setminus\{(x_i,y_i)\}$ respectively, and therefore their
predictions can be interpreted as draws from the posterior predictive distributions
$p(Y_*|X_*,x,y)$ and $p(Y_*|X_*)$, the exact distributions required for the definition
of information gain.

The approximate samples from these distributions which this interpretation affords
allows us to extend this framework and calculate an approximation to the information
gain formula. The resulting formula is:  

\begin{widetext}
\[
\text{KL}\left(p(Y_*|X_*,x,y) \parallel p(Y_*|X_*)\right)
=
\frac{1}{2}\left[
    \tr(\Sigma^{-1}_\text{oob}\Sigma_\text{ib})
    + (\mu_\text{oob} - \mu_\text{ib})^\top
    \Sigma^{-1}_\text{oob}
    (\mu_\text{oob}-\mu_\text{ib})
    - n
    - \ln{\frac{\det{\Sigma_\text{oob}}}{\det{\Sigma_\text{ib}}}}
\right]
\]
\end{widetext}

Full mathematical details of the derivation are provided in \Cref{sec:appendixB}.

\subsection{Heterogeneity}

For the purposes of an active learning seed subset, we propose that diversity in feature
space is important, especially early in the process. However, this cannot come at the
expense of informativeness. We therefore use Determinantal Point Processes (DPPs)
\citep{Kulesza2012} to balance informativeness and heterogeneity. DPPs provide a method
for randomly selecting from a set of points such that more heterogenous subsets are
given a higher probability of being sampled, with heterogeneity determined via an
imposed similarity measure. DPPs are chosen for this application because the sampling
distribution of a DPP is highly customizable; it is possible to emphasize and
deemphasize the marginal probability of inclusion of each point individually, and to
reduce the heterogeneity bias via weakening the similarity measure. By altering the
weight given to these two properties, it is possible to tune the methodology with prior
knowledge of the specific application. Even with large datasets, generating samples of
a desired size from a DPP can be efficient, thus allowing the entire methodology to be
performed in a matter of minutes for datasets of 10-100k labels.

The DPP can be constructed from a matrix $L_{ij}$ whose entries encode the correlations
between all $(i,j)$ point pairs in the superset, which is called the $L$-ensemble. The
probability that a subset $\mathcal{S}$ will be sampled from an $L$-ensemble is given by

\[
p(\mathcal{S})=\frac{\det L_\mathcal{S}}{L_\mathcal{S} + I}
\]

where $L_\mathcal{S}$ is the matrix created by selecting only the $\mathcal{S}$ set of
rows and columns of $L$, and $I$ is the identity matrix. This formulation emphasizes
sampling of dissimilar points, as the sampling is proportional to the determinant of the
correlation matrix. The $L$-ensemble formulation of a DPP allows the incorporation of
the previously determined informativeness scores via the quality-diversity factorization
of $L$ proposed by \citet{Kulesza2012}, while still explicitly considering
heterogeneity. This method is chosen over other means of enforcing heterogeneity, i.e.
performing biased sampling of groups of spatially clustered data, as it allows
a principled way of tuning the importance of informativeness and diversity via
alteration of the $L$ matrix construction.

The construction of the DPP and sampling from it require careful consideration. While
any correlation structure may be used to construct the $L$ matrix, in our application
$L$ is constructed via the Random Fourier Features (RFF) approximation to the squared
exponential kernel,\citep{Rahimi2007} where $k(x,y)=\exp(-\frac{|x-y|^2}{\ell})$ is
approximated by $\phi(X)^\top\phi(y)$, with 

\[
\phi(x)=\sqrt{2/R}\left[cos(\omega_r^\top x+b_r)\right]_{r=1}^R
\]

and $\omega_r \sim \mathcal{N}(0,\frac{1}{\ell}I)$ and $b_r \sim \text{Uniform}[0,2\pi]$
for a predetermined number of features $R$. This approach was utilized for three
reasons: first, it is found to be quicker than calculating the kernel explicitly as it
allows the use of fast dot product code. Second, it allows easy application of the
quality-diversity factorization of $L_{ij}=q_i \phi_i^\top \phi_j q_j$ by scaling each
point's $\phi(x_i)$ by their informativeness score. Third, and most importantly, this
affords us the ability to utilize a fast DPP sampling algorithm which we propose in
\Cref{sec:appendixC}.

\subsection{ATBagging Algorithm}
\label{sec:algo}

\begin{figure*}
\centering
\includegraphics[width=0.7\linewidth]{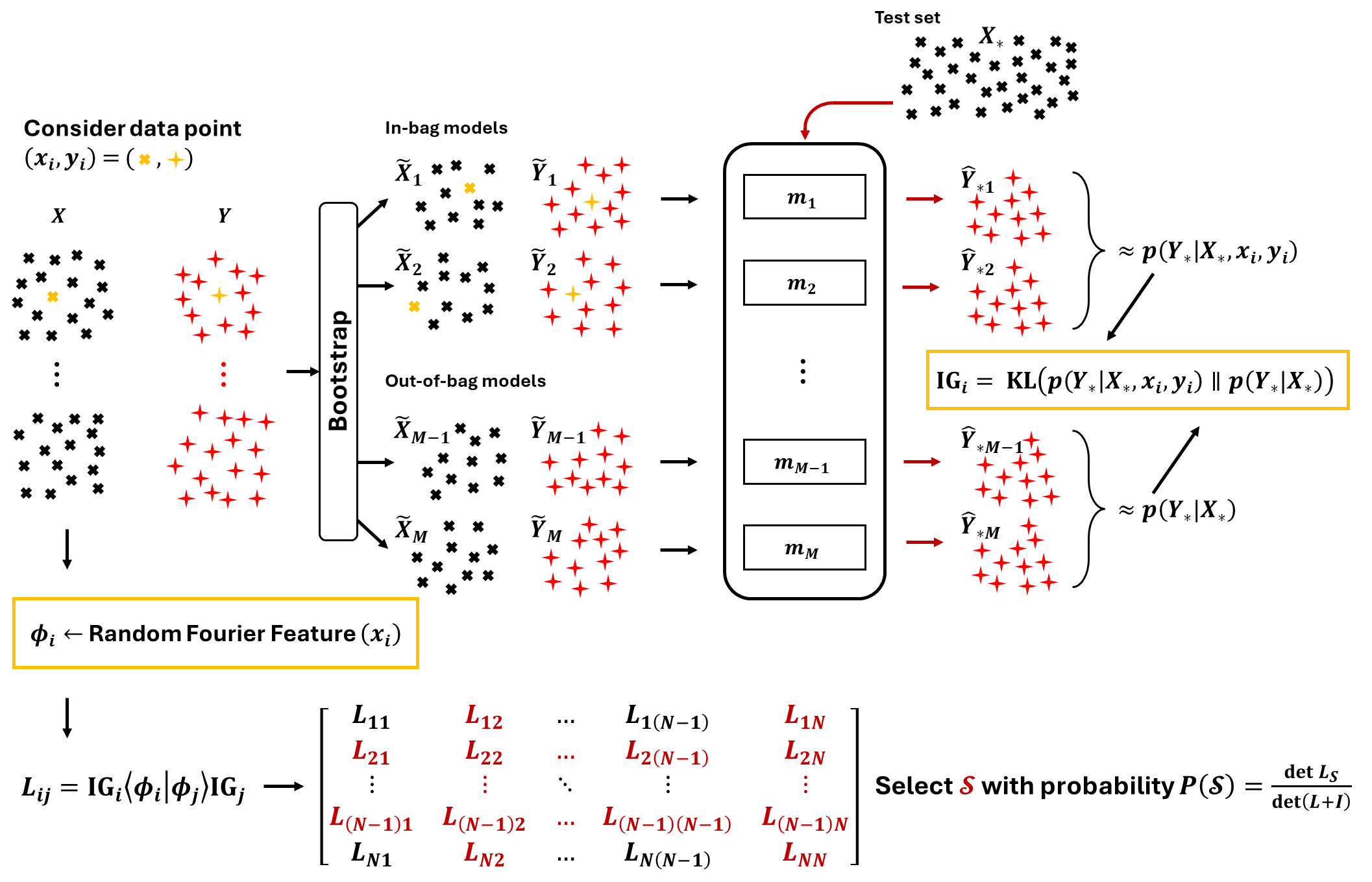}
\caption{Information flow of the knowledge gain and heterogeneity operators}
\label{fig:dataflow}
\end{figure*}

The combination of information score and heterogeneity is implemented via \Cref{alg:1},
and is shown schematically in \Cref{fig:dataflow}.

\begin{algorithm}[!h]
\caption{Transferable Subset Creation}\label{alg:1}
\KwData{%
    A dataset $\mathcal{D}=\{(x_i,y_i)\}_i^N$, a bagged ensemble with submodels
    $\mathcal{M} = \{m_i\}_i^M$, a test set $X_*$, a desired subset size $k$
}
\KwResult{A transferable subset $\mathcal{S}$}
\ForEach{$x_i \in \mathcal{D}$}{
    $\mathcal{M}_\text{ib} \gets \{m \in \mathcal{M} | x_i \in \tilde{X}_m\}$\;
    $\mathcal{M}_\text{oob} \gets \{m \in \mathcal{M} | x_i \notin \tilde{X}_m\}$\;
    Compute $\mu_\text{ib}, \mu_\text{oob}, \Sigma_\text{ib}, \Sigma_\text{oob}$ 
    using $\mathcal{M}_\text{ib}, \mathcal{M}_\text{oob}$ evaluated on the test set $X_*$
    (see \Cref{sec:appendixB})\;
    $\text{IG}_i \gets \text{KL}$ divergence between in-bag and out-of-bag
    distributions\;
    $\phi_i \gets \text{RandomFourierFeature}(x_i)$\;
}
Initialize matrix $L$\;
$L_{ij} \gets \text{IG}_i\phi_i^\top\phi_j\text{IG}_j$\;
$\mathcal{S} \gets \text{DPPSampler}(L)$\;
\end{algorithm}

\subsection{Data Sources}
\label{sec:data_sources}

We demonstrate the capability of the ATBagging approach to improve machine learning data
collection by evaluating its performance on four real-world datasets chosen to address
different types of potential applications: 1) the ERA5-Land Hourly Weather dataset,
\citep{Muoz-Sabater2021} 2) the Multi-XC QM9 molecular energy dataset,
\citep{Nandi2023} 3) the Forbes 2000 market evaluation dataset,\citep{Forbes2025} and
4) the Beijing PM\textsubscript{2.5} dataset.\citep{Liang2015} These data sets were
chosen to represent a wide range of possible applications, with varying degrees of $X$
and $Y$ transferability. Additionally, the use of collected rather than synthetic data
more accurately represents use cases where sampling error or bias may exist. We
partition these datasets into two types according to the transfer problem they address:
target-transfer problems and feature-shift problems. Target-transfer problems address
situations where the feature set is constant across the transfer (same $X$’s different
$Y$’s), while feature-shift problems address cases where both the $X$ and $Y$ sets may
be different between explored and unexplored domains, with an assumed correlation
between them.

The two datasets used to evaluate target-transfer problems are the ERA5-Land Hourly and
Multi-XC QM9 datasets. These problems are representative of cases where additional
(likely at additional cost) characterization of points is possible, but it is desirable
to minimize the cost of additional analysis. 

{
\renewcommand\tabularxcolumn[1]{m{#1}} 
\renewcommand{\arraystretch}{1.4}
\begin{table*}[!t]
	\caption{Summary of Data Sets Used}
	\centering
    \begin{tabularx}{\textwidth}{
        >{\raggedright\arraybackslash}m{0.12\textwidth} 
        >{\raggedright\arraybackslash}m{0.12\textwidth} 
        >{\raggedright\arraybackslash}m{0.12\textwidth} 
        >{\raggedright\arraybackslash}m{0.12\textwidth} 
        >{\raggedright\arraybackslash}X
    }
		\toprule
        Data Set & Type & $X$ & $Y$ & Description of Task \\
		\midrule
        QM9      & Target transfer & Same   & Highly correlated & 
        Selecting transfer subset from dataset with computationally cheap approximation
        of target to high accuracy calculation \\
        ERA5     & Target transfer & Same   & Somewhat correlated & 
        Selecting transfer subset from pre-existing dataset when a new target is desired
        for the same input set \\
        Forbes   & Feature shift   & Different distributions & Same type &
        Selecting initial subset to label from newly acquired unlabeled input set which
        is drawn from a different sampling distribution than the original dataset, but
        the target remains the same property \\
        PM\textsubscript{2.5} & Feature shift & Different distributinos & Different type &
        Selecting initial subset to label from newly acquired unlabeled input set which
        is drawn from a different sampling distribution than the original dataset, and
        the target is a different property \\
		\bottomrule
	\end{tabularx}
	\label{tab:table1}
\end{table*}
}

The ERA5-Land Hourly dataset\citep{Muoz-Sabater2021} (N$\approx$300,000) contains six
features of surface meteorological measurements. We select the two values related to
precipitation as the prediction targets to evaluate the performance of transfer
problems. This is an exemplar case of problems where a large dataset has already been
collected for one target, total precipitation, but researchers now wish to model a new
property of the system, surface runoff. Specifically, we frame this problem as
identifying a fixed, small number of locations where meteorologists should measure the
new property to gain the most performant and robust model for future predictions. 

The Multi-XC QM9 molecular energy database\citep{Nandi2023} compiles chemical
descriptors of \textasciitilde{}134,000 molecules and their energies as calculated by an
array of different accuracy quantum chemical methods. This dataset is selected to
evaluate target-quality transferability problems, where the true target value is
expensive to acquire, but less expensive, poorer, approximations of the target exist. We
select the two targets in this test case to be different in the quality of quantum
mechanical approximation used. The transfer target is chosen to be calculations that are
roughly 100 times more computationally expensive to acquire, M06-2X functional with
a triple-zeta polarized basis, than the low-level method, LDA(VWN) functional with
a single-zeta basis. We frame the problem as identifying which molecules the researchers
should select for explicit calculation with the expensive method to best predict
high quality molecular energies from molecular descriptors. 

The two datasets chosen to evaluate domain-shift transfer performance were the Beijing
PM\textsubscript{2.5} dataset and the Forbes dataset. Both these datasets represent
problems where the input set is expanded with new unlabeled data of the same type
(containing the same features) from a different sampling distribution, i.e. the same
features are measured for new data which is collected from a context that is expected to
be meaningfully different from that of the original data. For these problems, the subset
chosen by our method is determined via identifying the unlabeled points which are
closest to the selected points from the original dataset. This is done via either
Mahalanobis distance for fully numerical data or \emph{k}-prototypes distance for mixed
numerical-categorical data.\citep{Mahalanobis2018,Huang1998}

The Beijing PM\textsubscript{2.5} dataset\citep{Liang2015} (N$\approx$40,000) dataset
contains a variety of meteorological measurements which are used to predict the
PM\textsubscript{2.5} concentration (fine particulates which are less than 2.5 micron in
diameter) in Beijing from 2010 to 2014. With this dataset, neither the type of features
nor the target property change, but the distributions from which the input set is drawn
does. This domain split is achieved by dividing the dataset before and after 2012, which
is intended to capture the evolution of Chinese pollution regulations. The transfer
learning problem is framed as where in Beijing and when during the year it is best to
collect meteorological data to maximize the predictive capability for city-wide
particulate matter based on information available from previous years.

The Forbes market evaluation dataset\citep{Forbes2025} provides the sales, profits,
assets, industry, location, and market value for just over 1500 companies worldwide. The
domain split for this problem is constructed by dividing the companies into two classes,
Eastern (Asian) and Western (American and European) using Western companies as the
available domain and the other as the transfer domain. As these are disjoint divisions
(no company is considered both Eastern and Western), the transfer subset is constructed
by identifying the closest target companies to the source companies selected by the
down-selection method. Thus, the resulting problem is framed as where to best collect
information from Asian companies to maximize the predictive performance of their market
value, based on information available about Western companies. 

\subsection{Alternative Methods}

To evaluate the effectiveness of our methodology we assess it against three alternative
seed data set selection methods. The first is naïve uniform random sampling, which aims
to mimic the sampling distribution in feature space. This method takes no information
from the existing data. The second is PCA grid sampling, in which the data is
transformed via principal component analysis to remove linear correlations and grouped
into voxels on a regularly spaced grid, from which points are sampled randomly within
each voxel. This ensures feature-space diversity in the resulting subset. This again
does not translate target information from the existing data. Finally, we compare
performance against a loss-driven coreset methodology, in which data points are selected
via importance sampling with probabilities driven by their influence on the model’s loss
function, a common approach among simple coreset methodologies.\citep{Mirzasoleiman2020}
This prioritizes high-impact points to maximize informativeness but does not consider
heterogeneity. These baselines span the range of strategies including the naïve
approach, an approach addressing only heterogeneity, and an approach addressing only
informativeness.

\subsection{Performance Evaluation \& Metrics}

The performance of these methods was evaluated in two related tasks, the downselection
of representative subsets of the source dataset and the performance of these subsets for
transfer-active learning. We report accuracy, defined as a trial's $r^2$ divided by the
maximum $r^2$ achieved on any trial for that dataset. To summarize the performance of
the method across subset sizes we introduce an accuracy vs.\ subset size curve,
normalized such that perfect integrated accuracy is 1. We report the performance as the
normalized area under the learning curve (NAULC). Additionally, the proportion of trials
in which ATBagging outperformed its competitors is reported, with credible intervals
provided via a beta-binomial model of the pairwise comparison data with a Jeffreys
prior.\citep{Chen2008}

\begin{figure*}[t]
\centering
\includegraphics[width=0.7\linewidth]{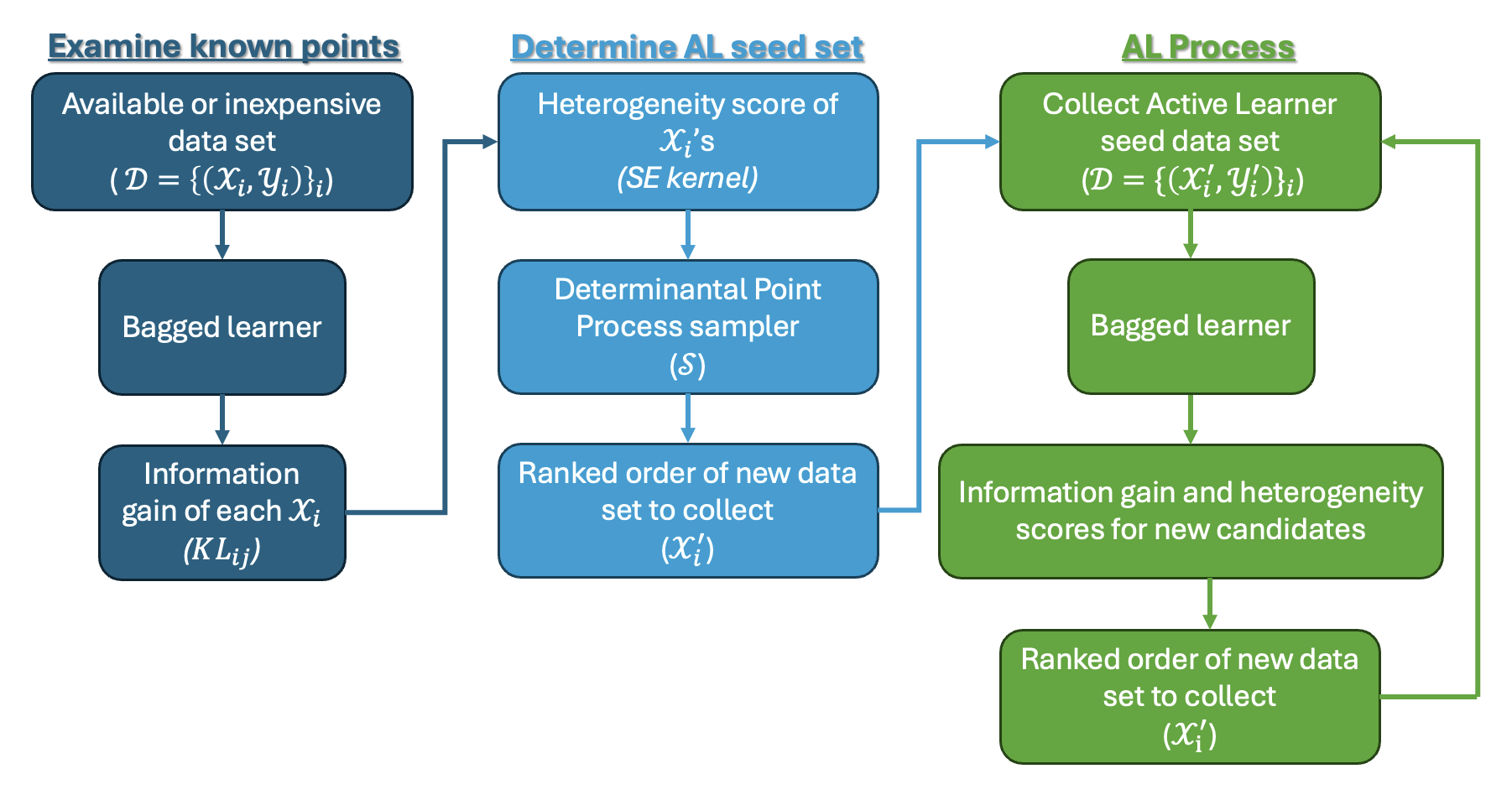}
\caption{Overview of the three key steps in ATBagging learning approach.}
\label{fig:schematic}
\end{figure*}

\section{Results \& Discussion}

The goal of ATBagging is use information from available or low collection cost data to
seed and drive an active learner. This proceeds through three steps: 1) examination of
known points and calculation of their information gain, 2) determination of active
learning seed set from the information gain and heterogeneity and 3) active learning of
the desired target data. This workflow is shown schematically in
\Cref{fig:schematic}. We test the performance of each of these processes with the
data sets described in \cref{sec:data_sources}. The results of our tests are
presented in three subsections. First, we discuss the details of how the methodology was
applied, describing the model architecture, active learning algorithm, and justifying
the choice of model hyperparameters. Then, we evaluate the quality of the subsets chosen
by our method against those of the alternative methods, by first investigating their
performance as downselected distillations of the source datasets, followed by their
quality as seed subsets for transfer-active learning.

\subsection{Experimental Setup}

We have chosen to evaluate ATBagging with a bagged ensemble model of 100 decision trees
(a random forest regressor, RFR)\citep{Hastie2009} paired with query-by-committee active
learning,\citep{BurbidgeRobertandRowland2007} with all evaluations being replicated 15
times. The methodology is agnostic to the characteristics of the base model in the
bagged ensemble, but decision trees were chosen due to their flexibility and simplicity.
Particularly, they can model mixed categorical and continuous features with ease while
having low computational demand. Additionally, prior work has demonstrated that, in the
active learning context, a bagged ensemble (a random forest) of 100 decision trees
performed better than an ensemble of 10 random forests of 10 trees
each.\citep{Kirsch2022}

\begin{figure}
    \centering
    \begin{subfigure}[t]{0.49\textwidth}
        \caption{}
        \label{fig:parity:a}
        \centering
        \includegraphics[width=0.8\textwidth]{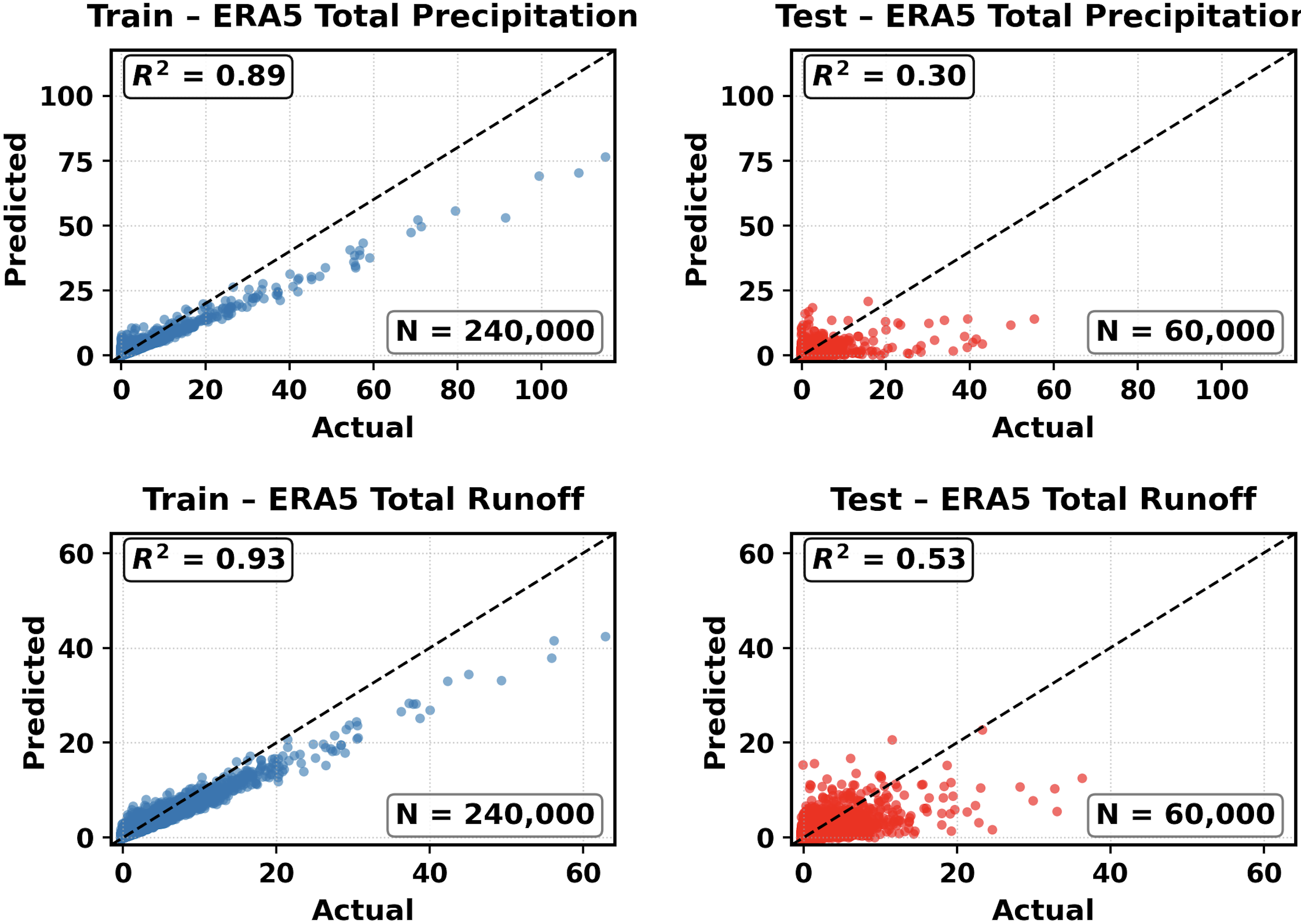}
    \end{subfigure}
    \begin{subfigure}[t]{0.49\textwidth}
        \caption{}
        \label{fig:parity:b}
        \centering
        \includegraphics[width=0.8\textwidth]{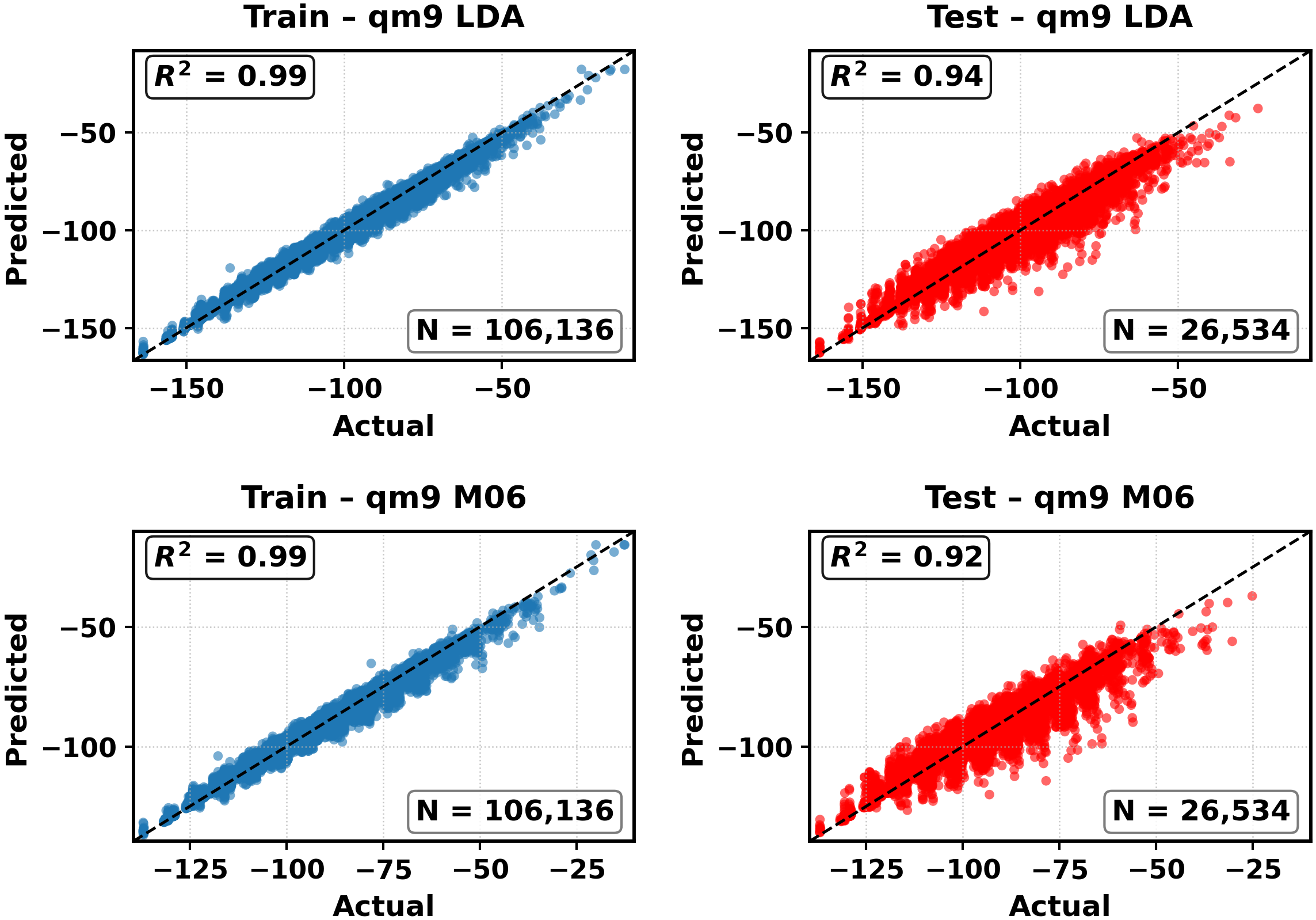}
    \end{subfigure}
    \begin{subfigure}[t]{0.49\textwidth}
        \caption{}
        \label{fig:parity:c}
        \centering
        \includegraphics[width=0.8\textwidth]{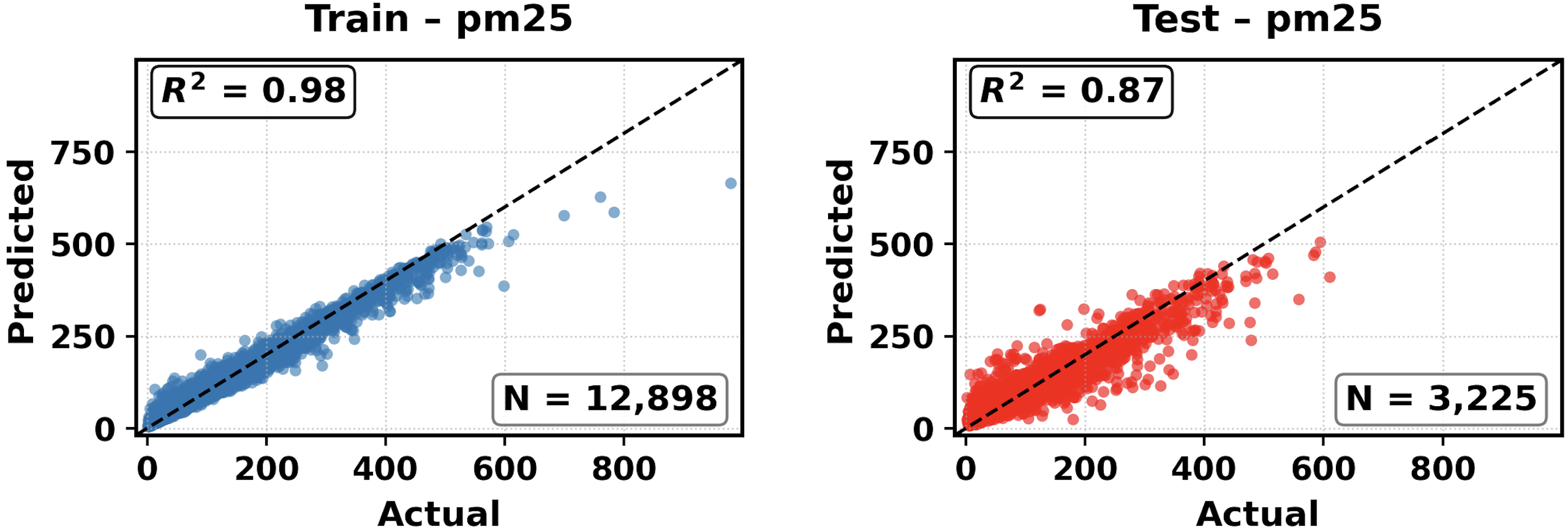}
    \end{subfigure}
    \begin{subfigure}[t]{0.49\textwidth}
        \caption{}
        \label{fig:parity:d}
        \centering
        \includegraphics[width=0.8\textwidth]{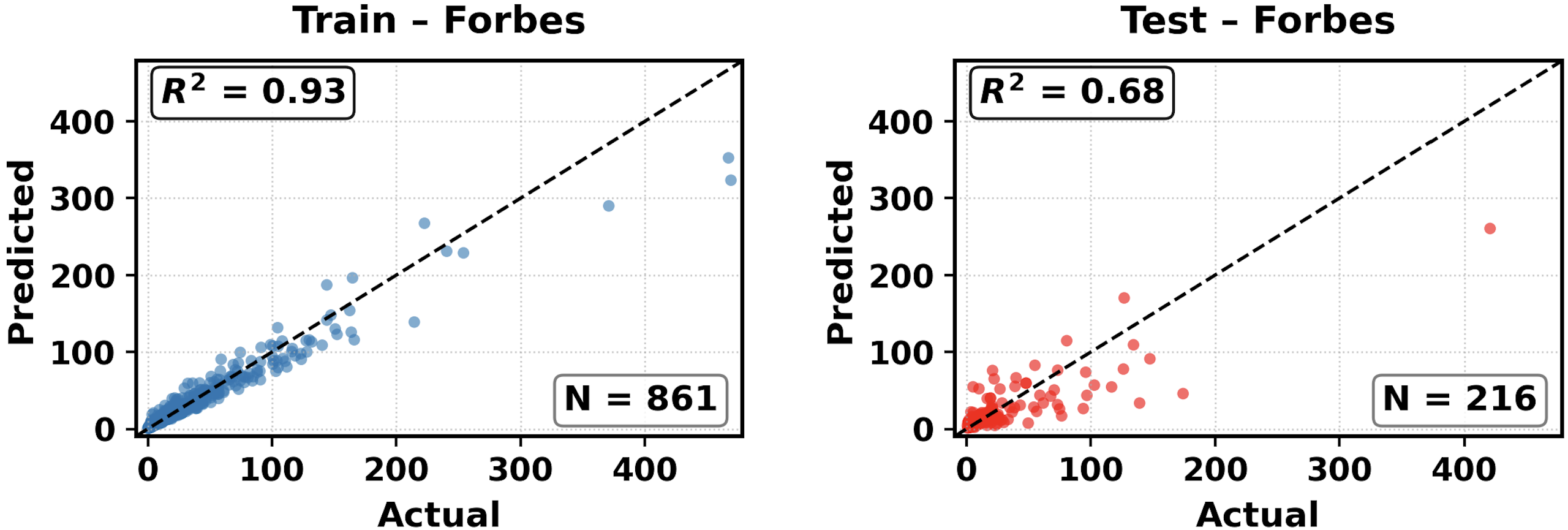}
    \end{subfigure}
    \caption{%
    Parity plots demonstrating the capability of the exemplar Scikit-learn RFR model to
    regress the selected datasets. Each subfigure corresponds to a different dataset
    utilized in the numerical experiments, with target-transfer datasets appearing
    twice, once for each target. The left column (blue) represents the training set
    performance (80\% of the data) and the right column (red) represents test set (20\%
    of the data). \subref{fig:parity:a} the ERA5 dataset with source
    target total precipitation and transfer target total runoff, \subref{fig:parity:b}
    the QM9 dataset with source target LDA(VWN)-SZP energies and transfer target
    M06-2X-TZP energies, \subref{fig:parity:c} the PM\textsubscript{2.5} dataset with
    target PM\textsubscript{2.5} particle concentration, and \subref{fig:parity:d} the
    Forbes 2000 dataset with target market value.
    }
    \label{fig:parity}
\end{figure}


We choose query-by-committee\citep{BurbidgeRobertandRowland2007} as our active learning
methodology as it is a simple, popular approach which is applicable to both regression
and classification tasks. Differences in the capabilities of seed set creation methods
will be most pronounced at small subset sizes, and problems which require these small
subset sizes, ones where label acquisition is extremely expensive, are the applications
where this methodology would see the most utility. Therefore, active learning tests are
seeded with a small subset size of $n_\text{seed} =$ 10.

\subsection{Model Justification \& Performance Characterization}

To demonstrate the feasibility of the RFR for the test datasets, we train a model on
each dataset and dataset split to be used to evaluate the ATBagging method (e.g. all
source and transfer datasets). We employ the python package
Scikit-learn\citep{Pedregosa2011} to train the RFR model. The entire available dataset
is split into an 80/20\% train and test set for RFR optimization. \Cref{fig:parity}
shows the parity plot for each RFR model and each dataset considered. We highlight the
generally excellent performance of the RFR models on the QM9, PM\textsubscript{2.5} and
Forbes datasets. The RFR model performs worse on the ERA5 weather dataset, but the 0.3
and 0.5 $r^2$ values are considered strong enough indication that the model is indeed
learning, therefore the dataset was included as an example of the methodology’s
performance on a difficult learning task.

\subsection{Downselection}

The ability of the different data selection methods to create representative subsets of
each source dataset was evaluated. Across datasets, the heterogeneity-aware methods
(ATBagging, PCA) retained more source accuracy than informativeness-only (loss
coreset). ATBagging was the best or tied for the best performing method in three of the
four datasets across all subset sizes tested and was always the best performing method
at the smallest subset sizes ($N_\text{tr} =$ 10, 60).

\begin{figure}
\centering
    \begin{subfigure}[t]{0.49\textwidth}
        \caption{}
        \label{fig:era_down:a}
        \centering
        \includegraphics[width=\textwidth]{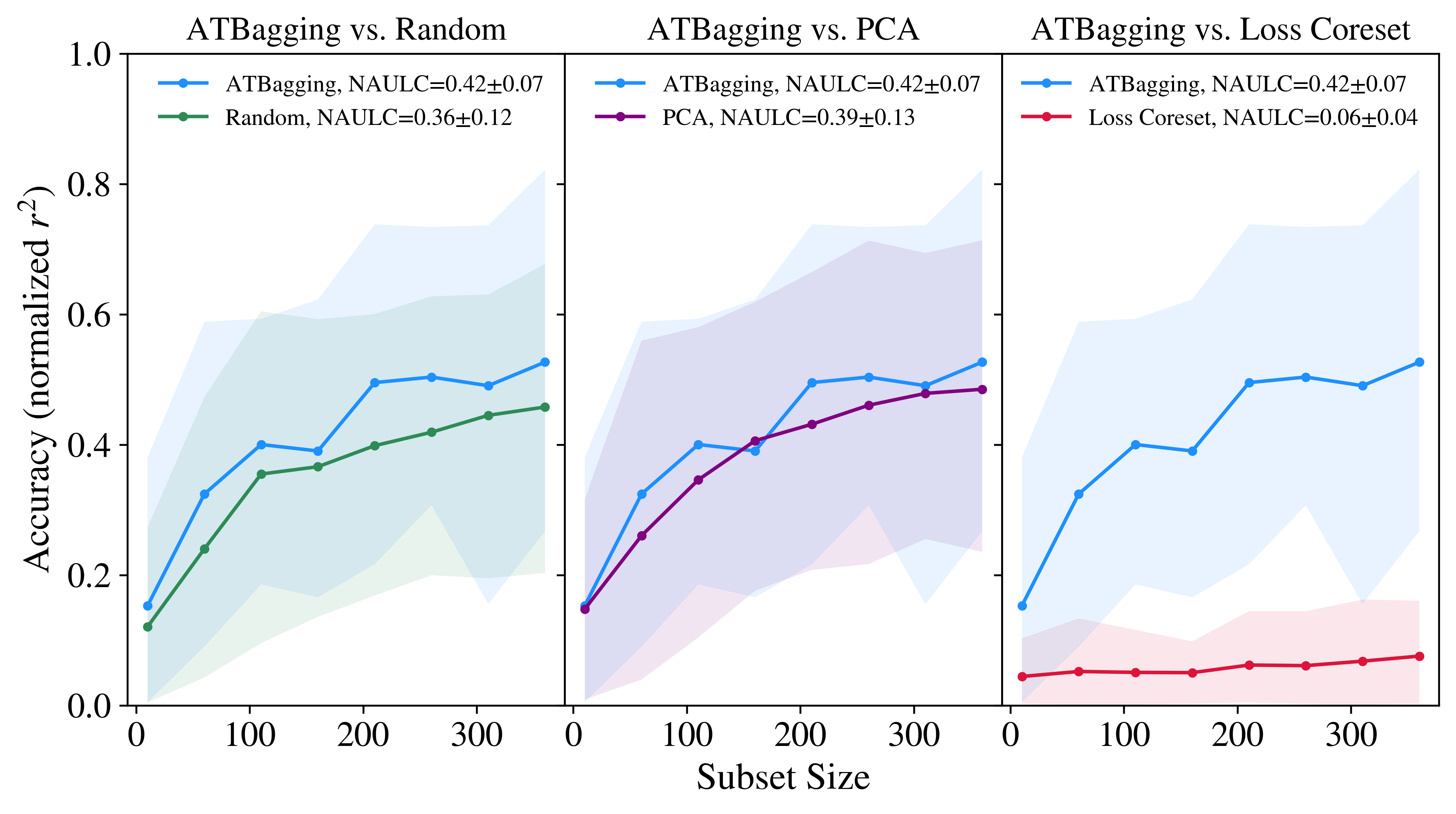}
    \end{subfigure}\hfill
    \begin{subfigure}[t]{0.49\textwidth}
        \caption{}
        \label{fig:era_down:b}
        \centering
        \includegraphics[width=\textwidth]{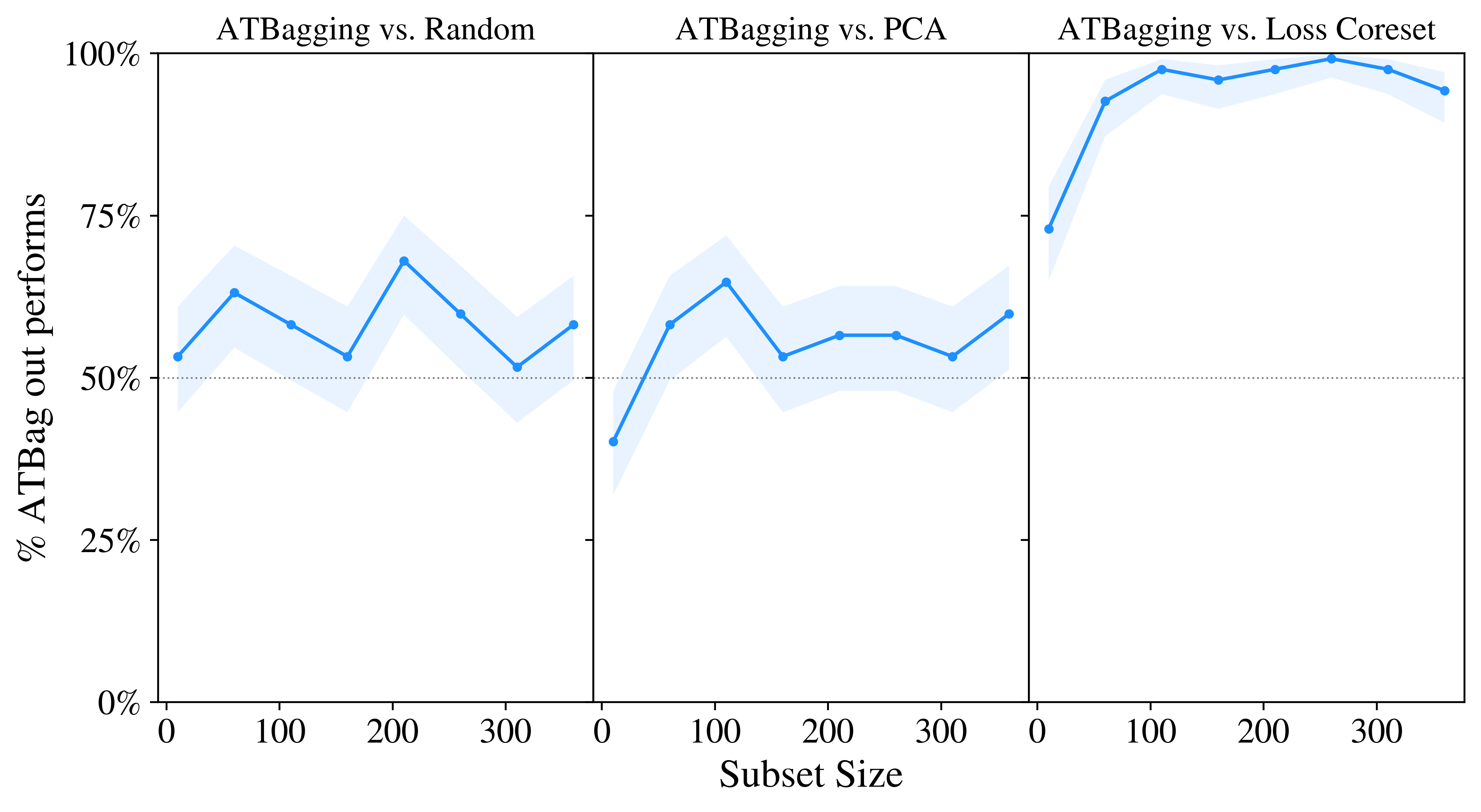}
    \end{subfigure}
    \caption{%
    Performance of the subset selection methods for dataset downselection on the ERA5
    dataset. 
    \subref{fig:era_down:a} Comparison of subset accuracy for those generated by
    ATBagging to those of the three alternative methods. Mean performance over the
    replicates is shown as the solid line, with 90\% high density intervals shown by the
    shaded regions. 
    \subref{fig:era_down:b} The percentage of trials in which ATBagging outperforms the
    indicated alternative method. The shaded regions represent 90\% credible intervals
    from a beta-binomial posterior of the pairwise comparison data.
    }
    \label{fig:era_down}
\end{figure}

\subsubsection{Downselection Performance: ERA5}

The ERA5 dataset was the most difficult of the datasets for the model to learn, and
therefore the results on this dataset are characterized via high variance across trials,
as seen in the 90\% density intervals shown in \Cref{fig:era_down:a}. Despite this,
ATBagging achieves higher or the same retention of the full dataset training accuracy
than the alternatives across all subset sizes tested. The methods that account for the
heterogeneity of the selected subset (ATBagging, PCA, and random sampling) outperform
the loss coreset method, which prioritizes the informativeness of the selected data
points only. These heterogeneity-driven sampling methods retain between 15--50\%,
15--46\%, and 13--41\%, of the full training accuracy on average for ATBagging, PCA, and
random sampling, respectively, across repeated experiments in the subset size range
examined ($N_\text{tr} =$ 10--350), while the loss coreset method consistently retains
only < 5\% across all subset sizes, as shown in \Cref{fig:era_down:a}. In the pairwise
comparison of ATBagging with the alternatives, it outperformed the Random and PCA
subsets moderately but consistently, with the proportion of trials in which ATBagging
out-performed these two remaining between 50--65\% for most subsets sizes, as shown in
\Cref{fig:era_down:b}. Due to the poor performance of the loss coreset method,
ATBagging outperformed it in nearly all trials with $n >$ 10, and in 75\% of trials with
$N_\text{tr} =$ 10.

\subsubsection{Downselection Performance: QM9}

\begin{figure}
    \centering
    \begin{subfigure}[t]{0.49\textwidth}
        \caption{}
        \label{fig:qm9_down:a}
        \centering
        \includegraphics[width=1.0\textwidth]{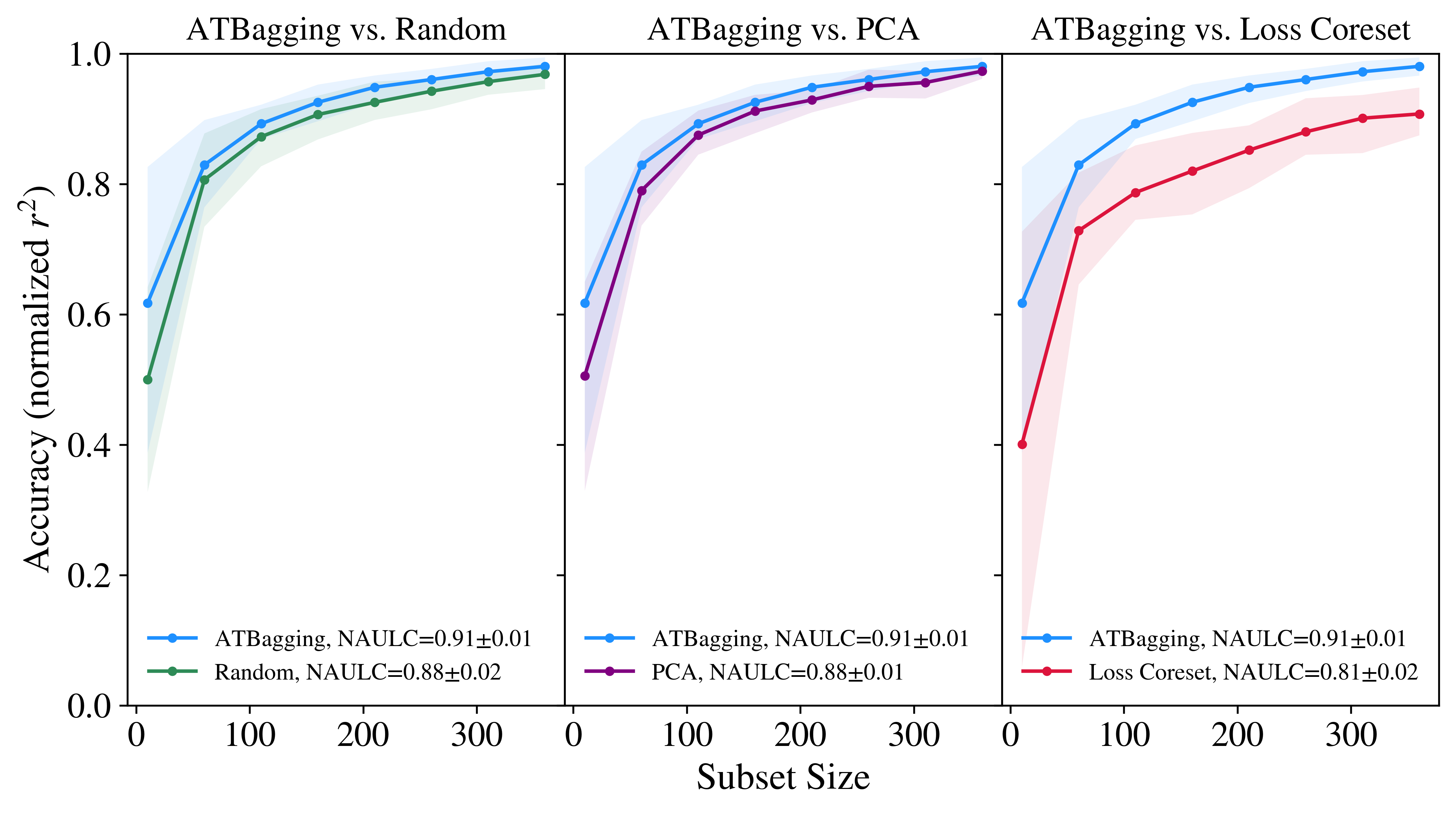}
    \end{subfigure}
    \begin{subfigure}[t]{0.49\textwidth}
        \caption{}
        \label{fig:qm9_down:b}
        \centering
        \includegraphics[width=1.0\textwidth]{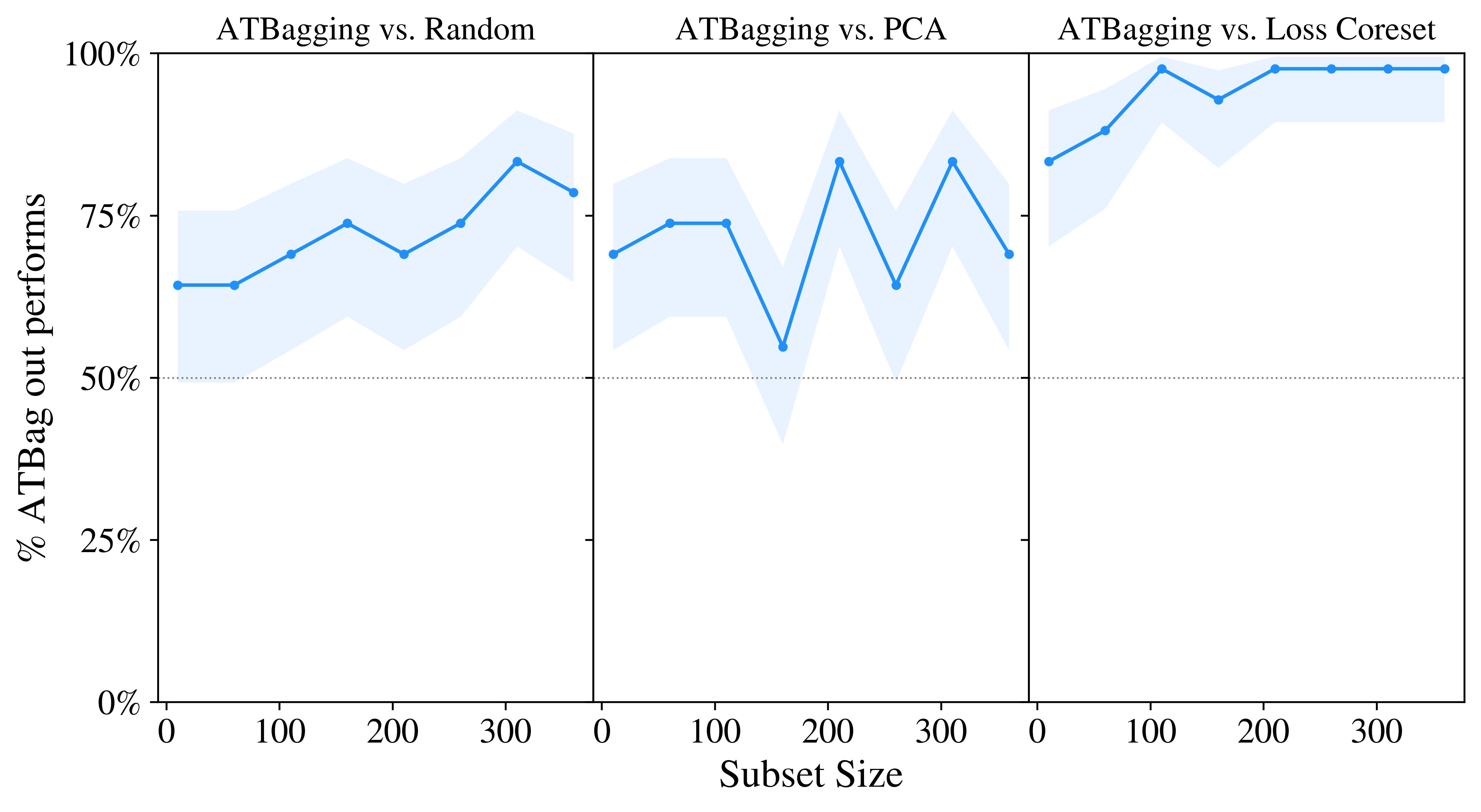}
    \end{subfigure}
    \caption{%
    Performance of the subset selection methods for dataset downselection on the
    QM9 dataset.
    \subref{fig:qm9_down:a} Comparison of subset accuracy for those generated by
    ATBagging to those of the three alternative methods. Mean performance over the
    replicates is shown as the solid line, with 90\% high density intervals shown by the
    shaded regions. 
    \subref{fig:qm9_down:b} The percentage of trials in which ATBagging outperforms the
    indicated alternative method. The shaded regions represent 90\% credible intervals
    from a beta-binomial posterior of the pairwise comparison data.
    }
    \label{fig:qm9_down}
\end{figure}


The QM9 dataset was found to pose a far simpler regression task, with subsets
demonstrating far higher accuracies as well as far less across-trial variance compared
to ERA5. At size $N_\text{tr} =$ 10, ATBagging subsets demonstrated a 61\% accuracy on
average, with the random, PCA, and loss coresets achieving 50\%, 51\%, and 40\%
respectively, as shown in \Cref{fig:qm9_down:a}. Again, like with the ERA5 dataset, the
methods which explicitly account for heterogeneity are the best performing, while the
loss coreset method demonstrates the worst performance, despite performing far better
than on ERA5. Even with these higher accuracies and reduced variance, the ATBagging
subsets outperformed the alternatives on average for every subset size tested, indeed
being the best subset in nearly all trials. In the pairwise comparisons, ATBagging
handily outperforms the alternatives on this dataset, demonstrating greater accuracy
than the alternatives across trials by 65\%--98\%, as shown in \Cref{fig:qm9_down:b}.

\subsubsection{Downselection Performance: Forbes}

\begin{figure}
    \centering
    \begin{subfigure}[t]{0.49\textwidth}
        \caption{}
        \label{fig:forbes_down:a}
        \centering
        \includegraphics[width=1.0\textwidth]{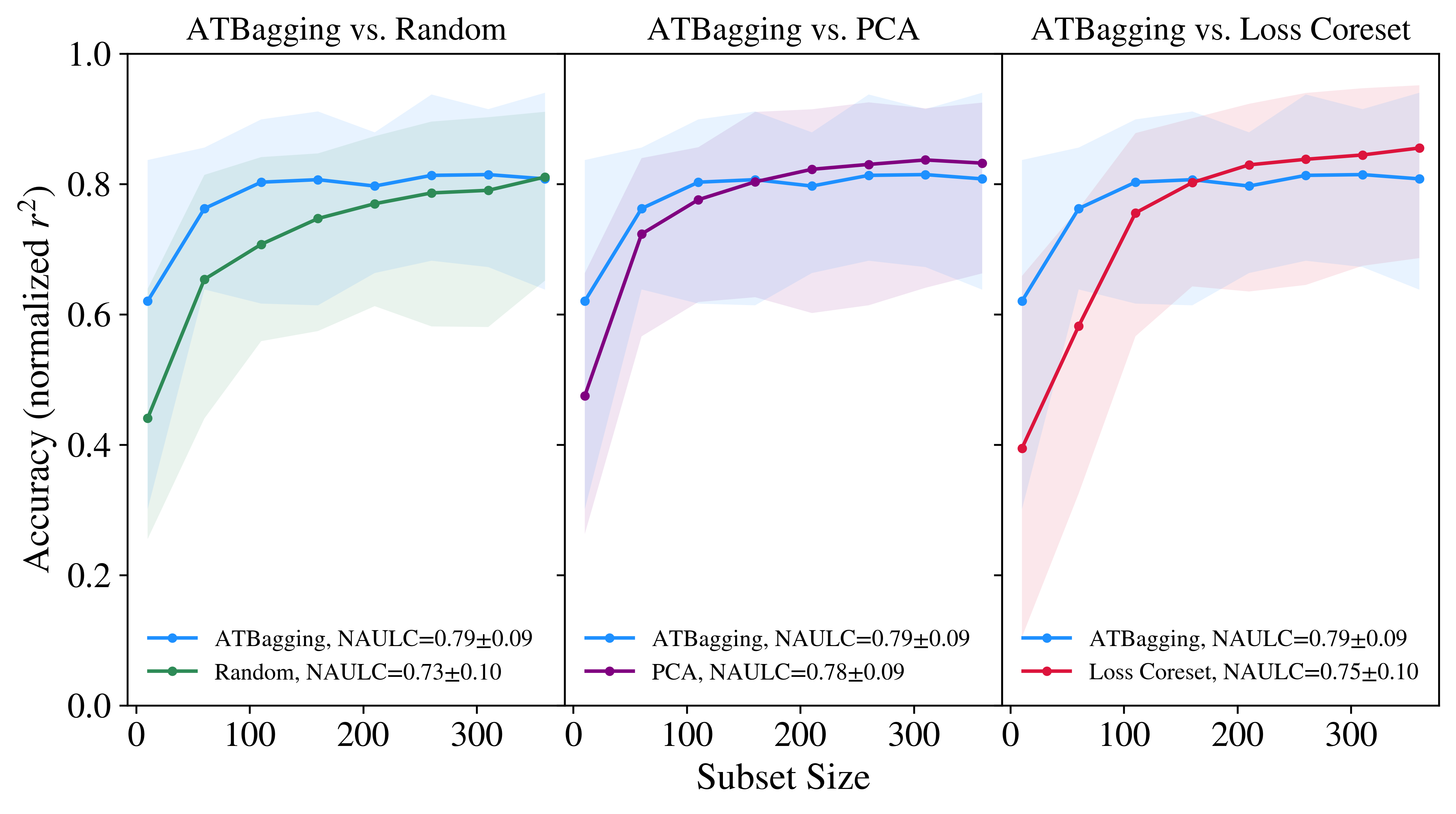}
    \end{subfigure}\hfill
    \begin{subfigure}[t]{0.49\textwidth}
        \caption{}
        \label{fig:forbes_down:b}
        \centering
        \includegraphics[width=1.0\textwidth]{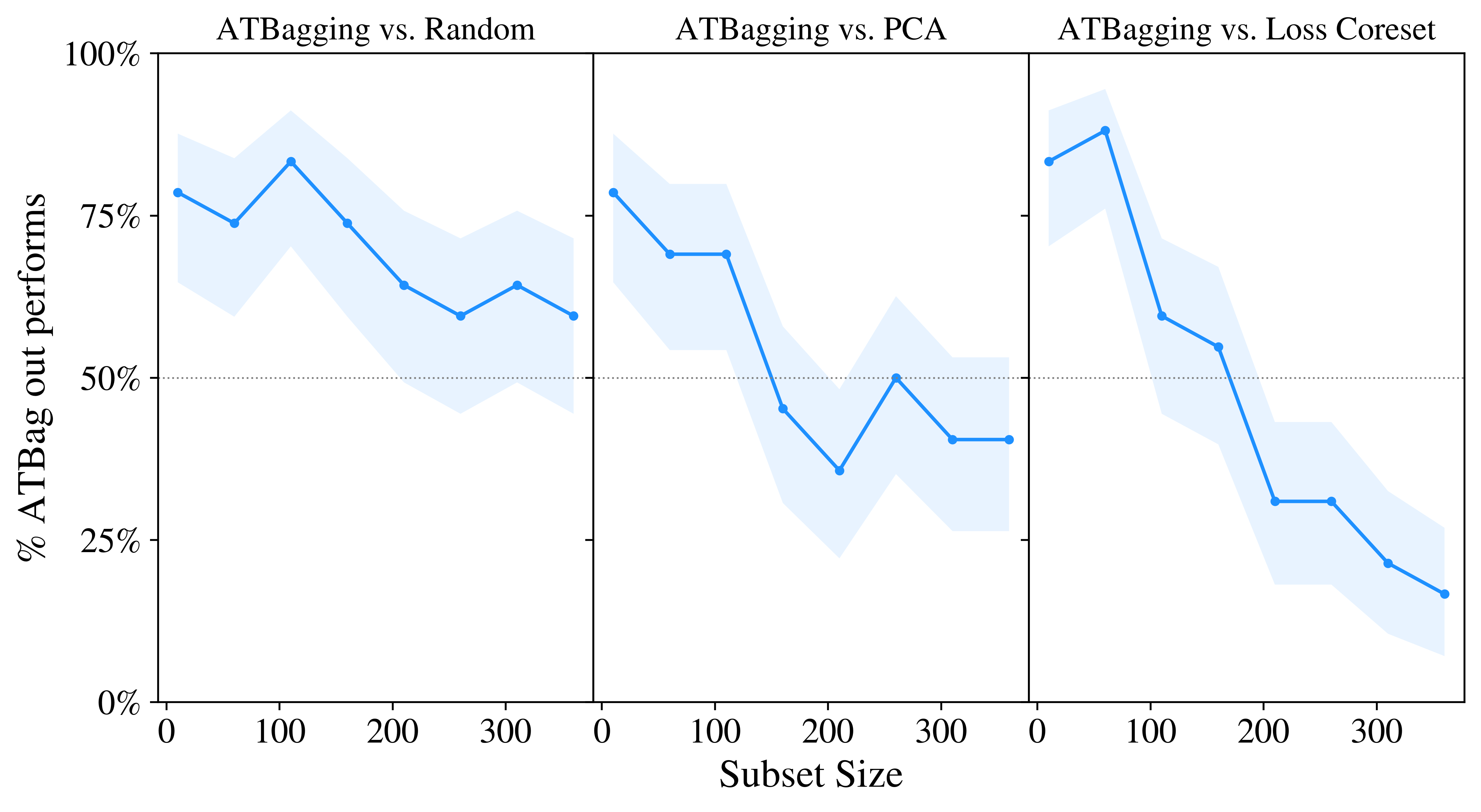}
    \end{subfigure}
    \caption{%
    Performance of the subset selection methods for dataset downselection on the
    Forbes 2000 dataset.
    \subref{fig:forbes_down:a} Comparison of subset accuracy for those generated by
    ATBagging to those of the three alternative methods. Mean performance over the
    replicates is shown as the solid line, with 90\% high density intervals shown by the
    shaded regions. 
    \subref{fig:forbes_down:b} The percentage of trials in which ATBagging outperforms
    the indicated alternative method. The shaded regions represent 90\% credible
    intervals from a beta-binomial posterior of the pairwise comparison data.
    }
    \label{fig:forbes_down}
\end{figure}


The Forbes dataset differs from the first two datasets in that it is the first where
ATBagging does not outperform or tie across all subset sizes. In this case, ATBagging
reaches a plateau in accuracy early, at $N_\text{tr} = $ 100, from which it does not
improve, as shown in \Cref{fig:forbes_down:a}. Prior to this plateau, for the small
subset sizes, ATBagging outperforms all the alternatives handily, demonstrating
75\%--80\% outperformance in the pairwise comparisons, as shown in
\Cref{fig:forbes_down:b}, due to its +0.1--0.2 difference in accuracy. For subsets
larger than these small sizes the alternative methods also plateau similarly, reaching
higher accuracies than ATBagging in the case of the PCA and loss coreset methods, while
it was lower for the random subset. These results can be understood by considering that
this dataset's features are mixed categorical and numerical, and thus the
distance-driven kernel similarity which powers the heterogeneity-enforcing DPP in the
ATBagging method is rendered less powerful due to the ambiguous ``distance'' between
categorical features. PCA ignores these small categorical dimensions as not contributing
strongly to the variance, and the loss coreset method does not consider these at all.
ATBagging is not rendered useless by this limitation: it still shows consistently
stronger pairwise performance than random subsets. Moreover, the balance between
heterogeneity and informativeness in ATBagging is controlled by a tunable
hyperparameter; however, to keep comparisons consistent across datasets, we use the
default setting throughout.

\subsubsection{Downselection Performance: PM\textsubscript{2.5}}

\begin{figure}
    \centering
    \begin{subfigure}[t]{0.49\textwidth}
        \caption{}
        \label{fig:pm25_down:a}
        \centering
        \includegraphics[width=1.0\textwidth]{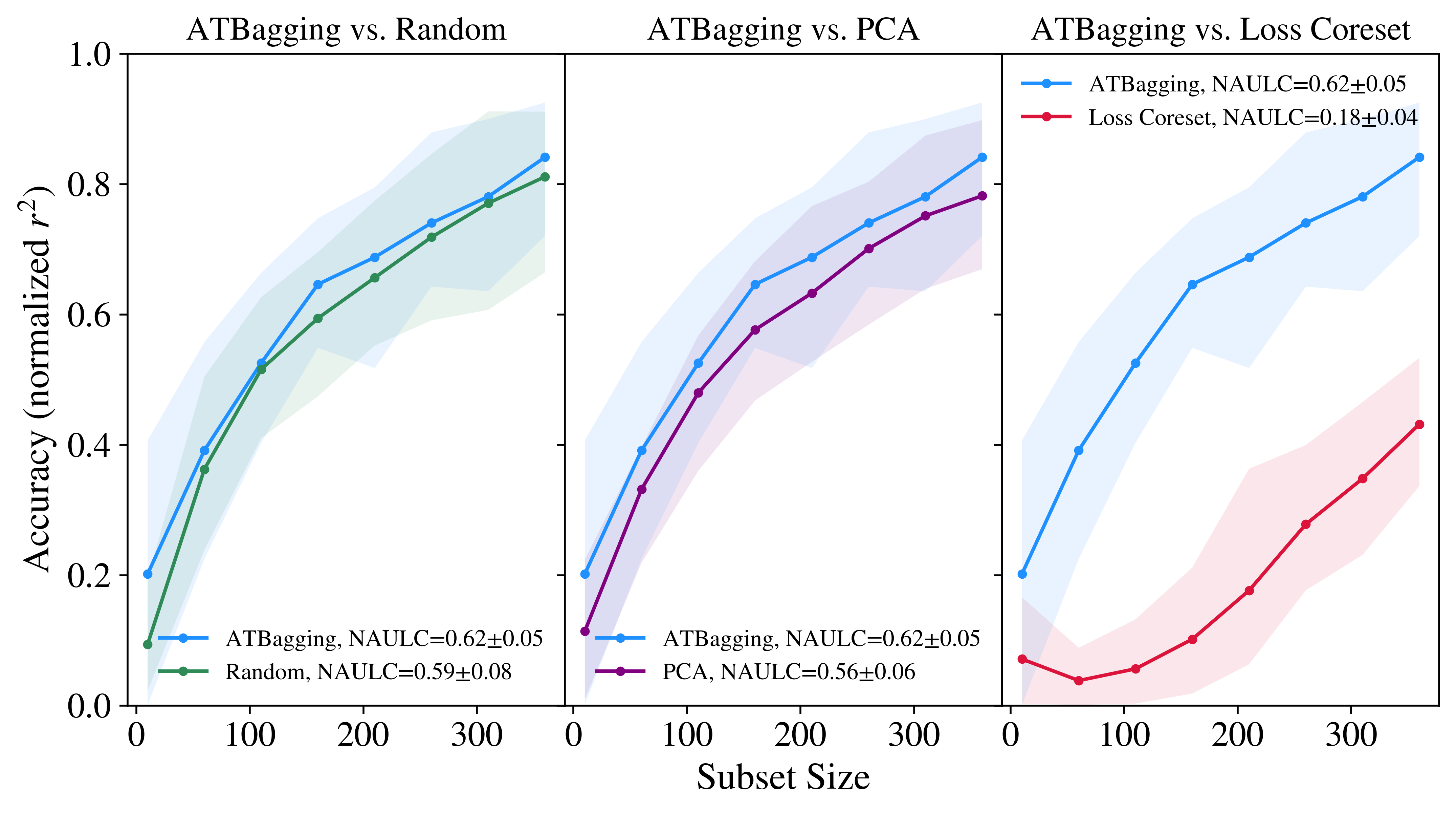}
    \end{subfigure}\hfill
    \begin{subfigure}[t]{0.49\textwidth}
        \caption{}
        \label{fig:pm25_down:b}
        \centering
        \includegraphics[width=1.0\textwidth]{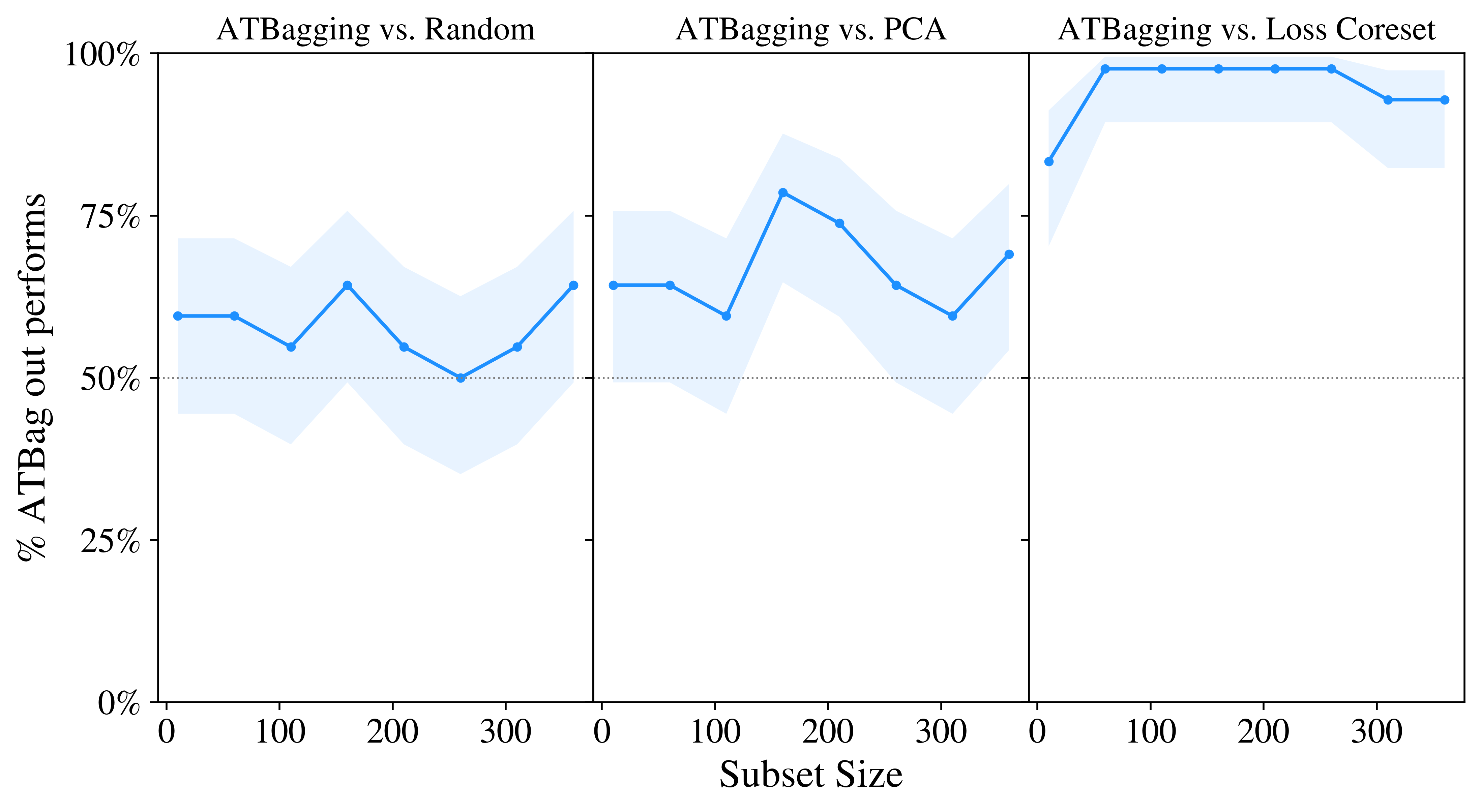}
    \end{subfigure}
    \caption{%
    Performance of the subset selection methods for dataset downselection on the
    PM\textsubscript{2.5} dataset. 
    \subref{fig:pm25_down:a} Comparison of subset accuracy for those generated by
    ATBagging to those of the three alternative methods. Mean performance over the
    replicates is shown as the solid line, with 90\% high density intervals shown by the
    shaded regions. 
    \subref{fig:pm25_down:b} The percentage of trials in which ATBagging outperforms the
    indicated alternative method. The shaded regions represent 90\% credible intervals
    from a beta-binomial posterior of the pairwise comparison data.
    }
    \label{fig:pm25_down}
\end{figure}


Finally, the PM\textsubscript{2.5} dataset generates similar behavior to the ERA5
dataset, in which the heterogeneity-enforcing methods perform well, with ATBagging
performing the best overall. Also, like ERA5 the loss coreset method performs poorly,
demonstrating accuracies < 10\% for subsets up to $N_\text{tr} = $ 100, as shown in
\Cref{fig:pm25_down:a}. Like all other datasets, ATBagging performs best at the
smallest subset sizes, with an advantage of \textasciitilde{}0.1 in accuracy above the
alternatives at $N_\text{tr} =$ 10, which is realized as a 60--80\% rate of
outperformance in the pairwise comparisons at that size. ATBagging’s pairwise comparison
outperformance against the random and PCA subsets ranges between 50--78\% across subset
sizes and is near 100\% against loss coreset subsets, as shown in
\Cref{fig:pm25_down:b}.

\subsubsection{Overall Performance in Data Set Downselection}

Our analysis reveals the capability of the ATBagging method to select performant small
subsets from real-world datasets of differing types from different domains, while
consistently out-performing alternative methods. We find that this outperformance is
largely irrespective of the subset size, being violated only in a situation where our
specific implementation of kernel similarity does not map nicely to the data (i.e.
categorical features). Even in this scenario, ATBagging achieved a higher rate of
success compared to alternatives on the smallest subset sizes, indicative of its high
utility for low-data scenarios where the cost of data acquisition might preclude the
availability of larger subset sizes.

\FloatBarrier
\subsection{Transfer \& AL Performance}

With the quality of the subsets produced by ATBagging demonstrated on the source
datasets, we examine performance of the seed subsets for transfer active learning. This
is tested for seed subset sizes of $n_\text{seed} = $ 10, 50, 100 on the four source
datasets.

\subsubsection{Transfer Seed Subset Performance: QM9}

\begin{figure}
    \centering
    \begin{subfigure}{0.49\textwidth}
        \caption{}
        \label{fig:qm9_trans:a}
        \centering
        \includegraphics[width=1.0\textwidth]{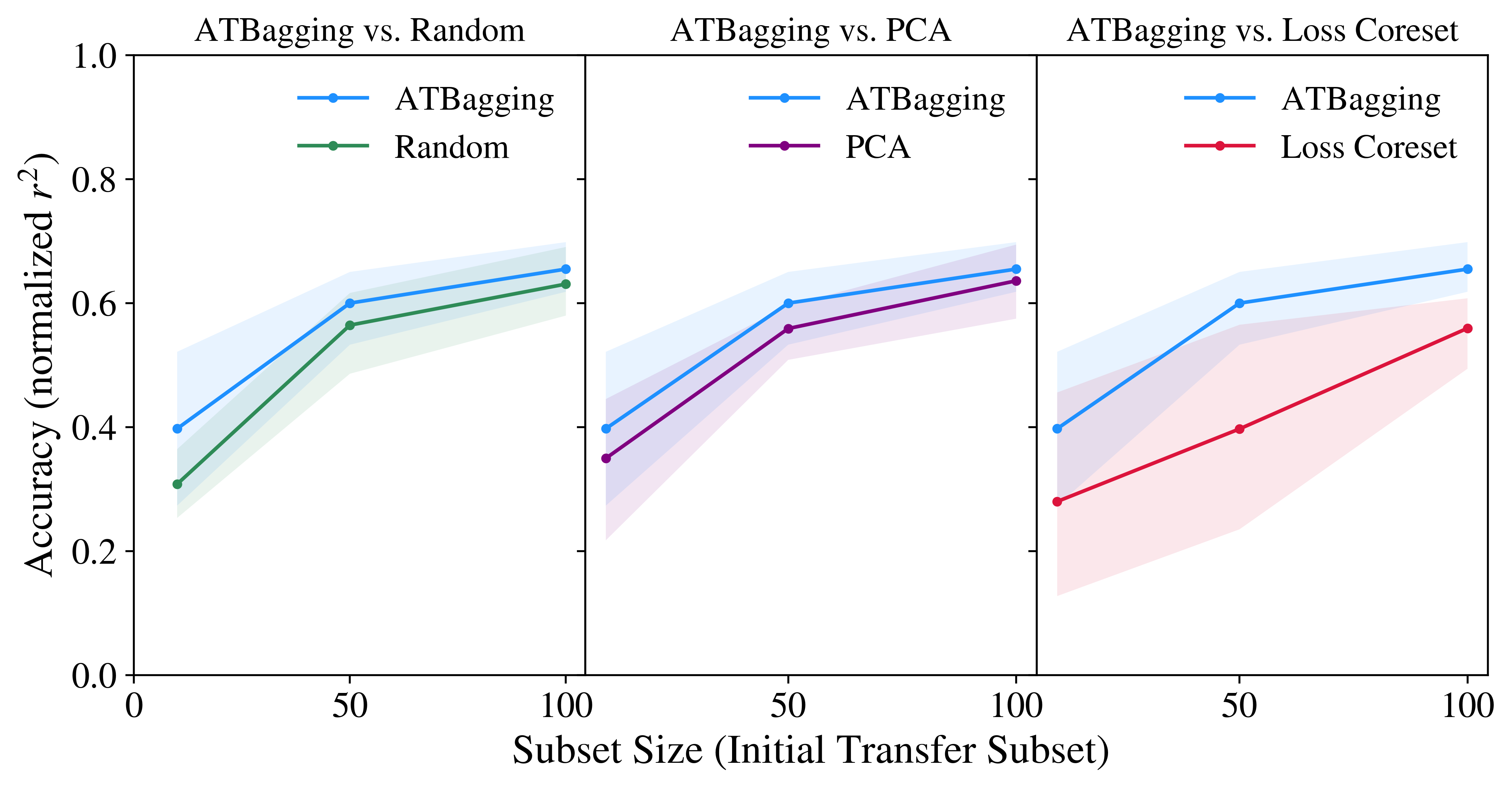}
    \end{subfigure}
    \begin{subfigure}[t]{0.49\textwidth}
        \caption{}
        \label{fig:qm9_trans:b}
        \centering
        \includegraphics[width=1.0\textwidth]{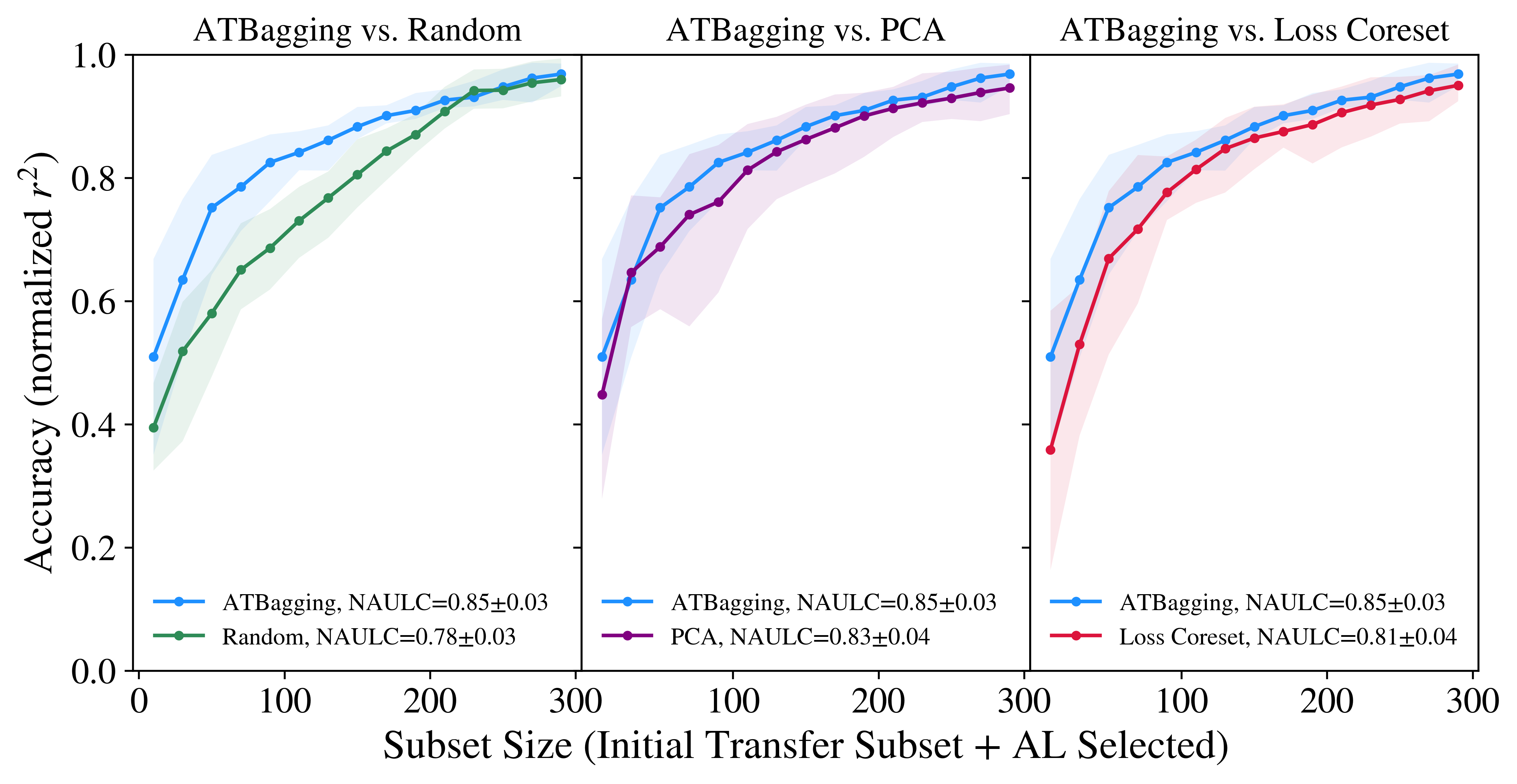}
    \end{subfigure}
    \begin{subfigure}[t]{0.49\textwidth}
        \caption{}
        \label{fig:qm9_trans:c}
        \centering
        \includegraphics[width=1.0\textwidth]{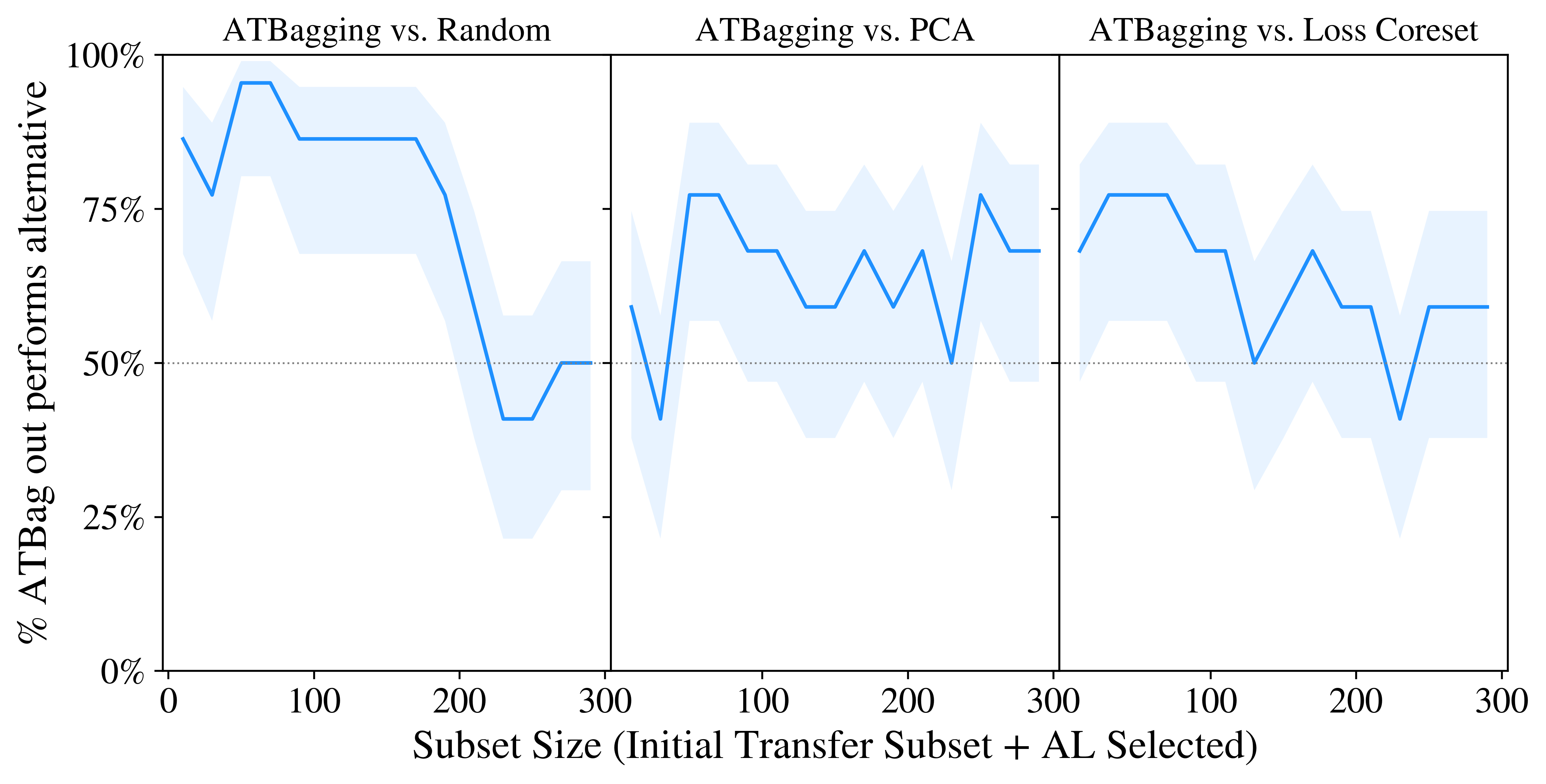}
    \end{subfigure}
    \caption{%
    Transfer and active learning performance for the seed subset creation methods on
    the QM9 dataset. 
    \subref{fig:qm9_trans:a} The initial transfer performance (ITP) of RFR models
    trained on the seed subsets generated by ATBagging and the three alternative methods
    for seed subset sizes of 10, 50, and 100.
    \subref{fig:qm9_trans:b} Active learning curves starting from $n_\text{seed}$ of 10
    growing with $m_\text{collect}$ of 20 to a subset of size 290. Mean accuracy is
    shown as solid lines, with 90\% high density intervals shown by the shaded regions.
    The legend reports the NAULC (mean$\pm$std.dev.) of the curves. 
    \subref{fig:qm9_trans:c} Pairwise model comparisons, the percentage of trials from
    the AL trials in \subref{fig:qm9_trans:b} in which ATBagging outperformed the
    indicated alternative method. 90\% credible intervals are shown in the shaded
    regions.
    }
    \label{fig:qm9_trans}
\end{figure}

The task for the QM9 dataset was to generate an active learning seed subset of candidate
molecules for performing an expensive high accuracy quantum chemical calculation from
a dataset containing only a computationally cheap approximation of the more expensive
desired target. In this task, the domain dataset remained unchanged (i.e. the set of
molecules is the same for both the source and transfer tasks). The initial transfer
performance (ITP) on different transfer subsets is shown in \Cref{fig:qm9_trans:a}.
For all three transferred subset sizes, $n_\text{seed} = $10, 50 and 100, active
learning starting from the ATBagging seeds out-perform those from the random, PCA, and
loss coreset methods.

\Cref{fig:qm9_trans:b} shows the result of active learning from the smallest transfer
set, $n_\text{seed} = $ 10, up to a training set size of 290 via 14 AL acquisition steps
each with an acquisition size ($m_\text{collect}$) of 20. Such a small $n_\text{seed}$
was chosen as it demonstrates the most difficult situation for a transfer subset and is
where differences between methodologies will be most apparent. The active learning
curves for the $n_\text{seed} = $ 50 and 100 subsets are included in
\Cref{sec:appendixD}. The seed subsets chosen via ATBagging resulted in the highest
accuracy of the resulting models, with a NAULC of 0.85±0.03 versus the 0.78±0.03,
0.83±0.04, and 0.81±0.04 of subsets generated using the random, PCA, and loss coreset
methods, respectively. Thus, ATBagging provides best models with the least data.  

The greatest difference in performance is apparent earliest in the active learning
process, when the AL acquired training set size ($N_\text{tr}$) is below approximately
150 data points, after which the models converge towards similar performance. The quick
learning is also seen in the pairwise comparison plot, where ATBagging substantially
outperforms random and loss coreset prior to the $N_\text{tr} = $ 150 mark, as shown in
\Cref{fig:qm9_trans:b}. PCA demonstrates performance on par with ATBagging at low
collected data set sizes, but ATBagging predictions are better after $N_\text{tr} = $ 50
points collected, and maintains a modest improvement in accuracy and a smaller
confidence interval until $N_\text{tr} = $ 300. This finding demonstrates that the
quality of a subset for active learning purposes is not just the accuracy of the model
which the subset produces but also the quality of the uncertainty estimates which it
produces.

\subsubsection{Transfer Seed Subset Performance: PM\textsubscript{2.5}}

\begin{figure}
    \centering
    \begin{subfigure}{0.49\textwidth}
        \caption{}
        \label{fig:pm25_trans:a}
        \centering
        \includegraphics[width=1.0\textwidth]{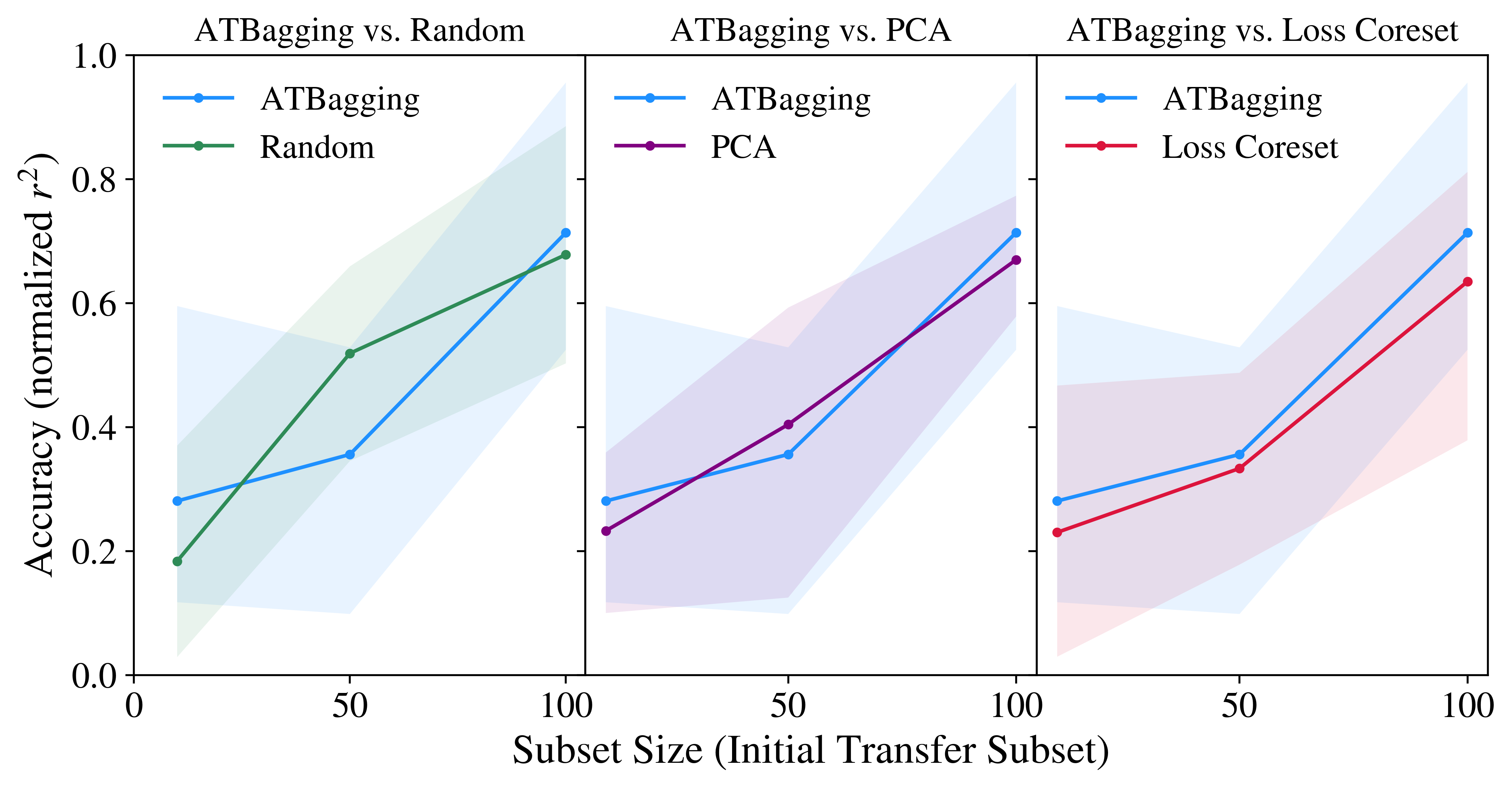}
    \end{subfigure}
    \begin{subfigure}[t]{0.49\textwidth}
        \caption{}
        \label{fig:pm25_trans:b}
        \centering
        \includegraphics[width=1.0\textwidth]{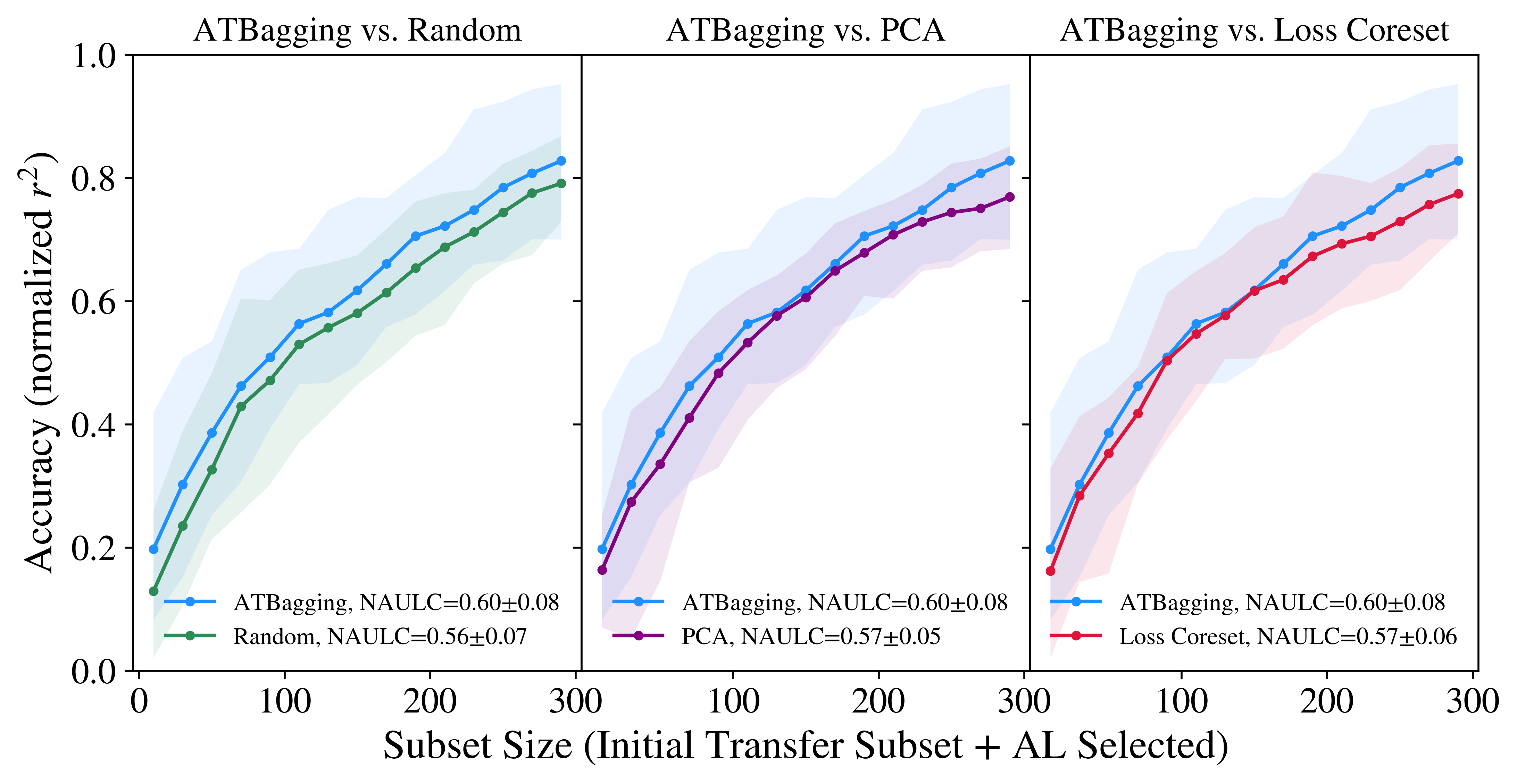}
    \end{subfigure}
    \begin{subfigure}[t]{0.49\textwidth}
        \caption{}
        \label{fig:pm25_trans:c}
        \centering
        \includegraphics[width=1.0\textwidth]{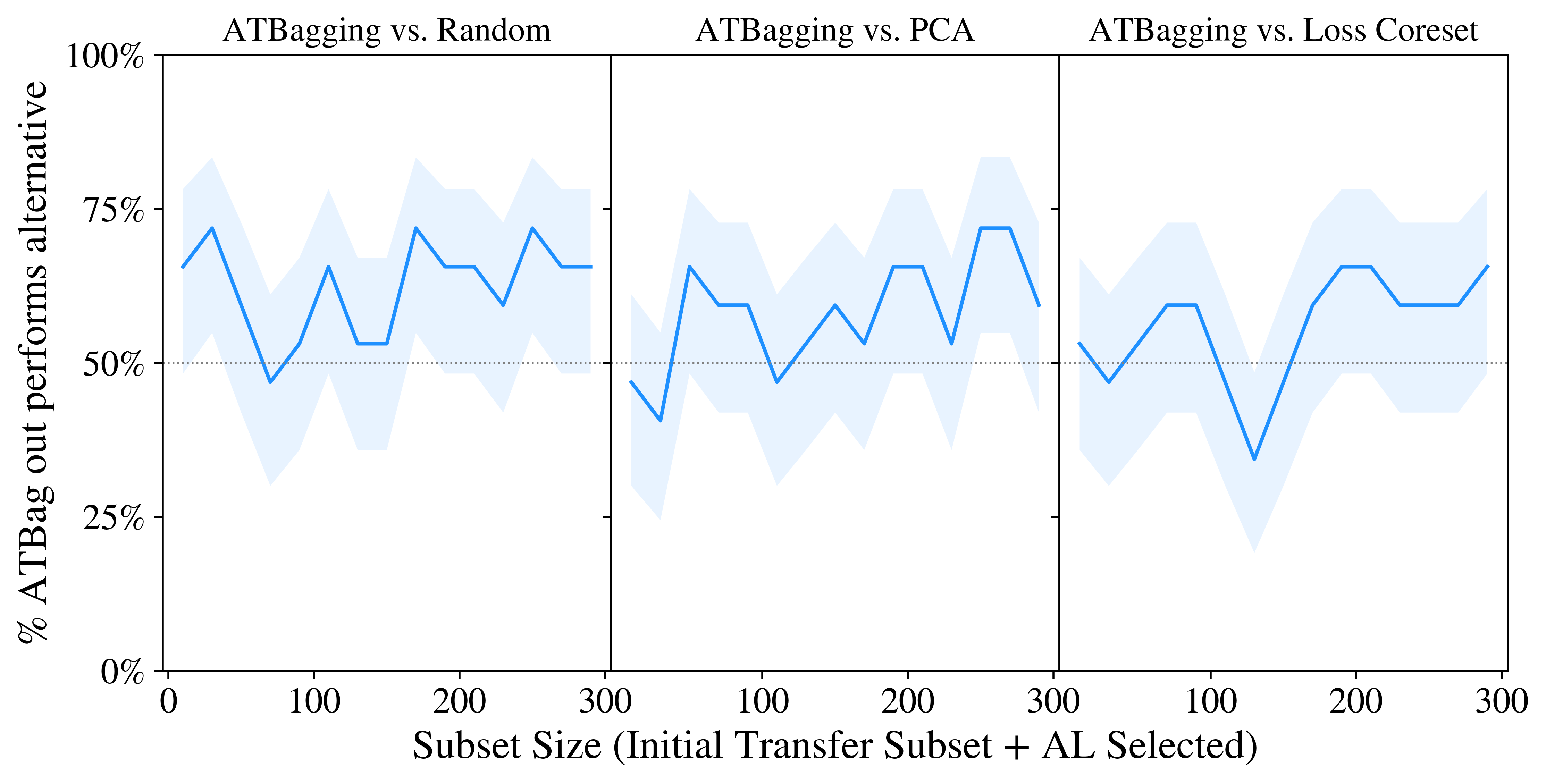}
    \end{subfigure}
    \caption{%
    Transfer and active learning performance for the seed subset creation methods on
    the PM\textsubscript{2.5} dataset. 
    \subref{fig:pm25_trans:a} The initial transfer performance (ITP) of RFR models
    trained on the seed subsets generated by ATBagging and the three alternative methods
    for seed subset sizes of 10, 50, and 100.
    \subref{fig:pm25_trans:b} Active learning curves starting from $n_\text{seed}$ of 10
    growing with $m_\text{collect}$ of 20 to a subset of size 290. Mean accuracy is
    shown as solid lines, with 90\% high density intervals shown by the shaded regions.
    The legend reports the NAULC (mean$\pm$std.dev.) of the curves. 
    \subref{fig:pm25_trans:c} Pairwise model comparisons, the percentage of trials from
    the AL trials in \subref{fig:pm25_trans:b} in which ATBagging outperformed the
    indicated alternative method. 90\% credible intervals are shown in the shaded
    regions.
    }
    \label{fig:pm25_trans}
\end{figure}

The transfer task for the PM\textsubscript{2.5} dataset is a feature shift task, in
which a model trained on a dataset of particulate matter readings from a trial period is
used to select a set of locations and times to record particulate matter in the future.
This is an example of a feature shift task, where the set of input points are
potentially different between the source and transfer. ITP (\Cref{fig:pm25_trans:a})
shows a modest advantage towards ATBagging at $n_\text{seed} = $ 10 and 100 compared to
random and PCA subsets, and at all sizes compared to loss coreset subsets. At
$n_\text{seed} = $ 50, both random subset selection and PCA outperform ATBagging in ITP.
It is important to note that the ATBagging $n_\text{seed} = $ 50 subsets outperformed
the PCA and random $n_\text{seed} = $ 50 subsets on the source domain task, indicating
that downselection performance does not ensure transfer performance.

ATBagging achieves a modestly higher AL performance than the other methods for the
$n_\text{seed} = $ 10 subsets. It has an NAULC of 0.60$\pm$0.08 compared to
0.57$\pm$0.05 for PCA and loss coreset subsets, and 0.56$\pm$0.07 for random subsets, as
shown in \Cref{fig:pm25_trans:b}. This is evident in the AL learning curve, which
shows the ATBagging subsets averaging slightly above those of the other methods, with
loss coreset achieving similar performance earlier in the AL process ($N_\text{tr}
< $ 150). This is also apparent in the pairwise comparisons
(\Cref{fig:pm25_trans:c}), where ATBagging maintains a roughly 60\% lead on average
against the random and PCA subsets. A modest lead is also seen in the pairwise
comparisons against the loss coreset subsets, but the 50\% line remains within the 90\%
credible intervals for the full learning trajectory.

\begin{figure}
    \centering
    \begin{subfigure}{0.49\textwidth}
        \caption{}
        \label{fig:forbes_trans:a}
        \centering
        \includegraphics[width=1.0\textwidth]{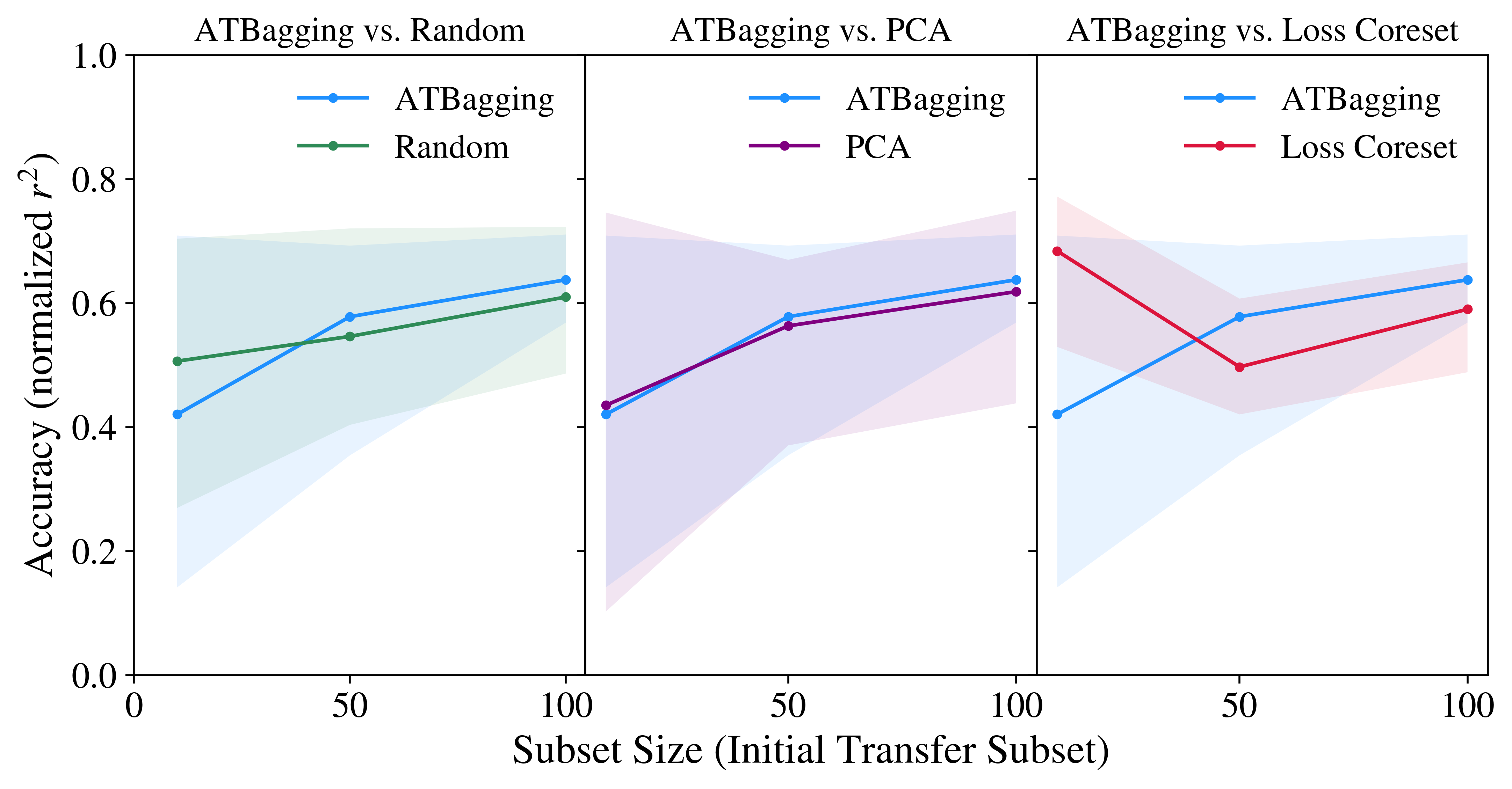}
    \end{subfigure}
    \begin{subfigure}[t]{0.49\textwidth}
        \caption{}
        \label{fig:forbes_trans:b}
        \centering
        \includegraphics[width=1.0\textwidth]{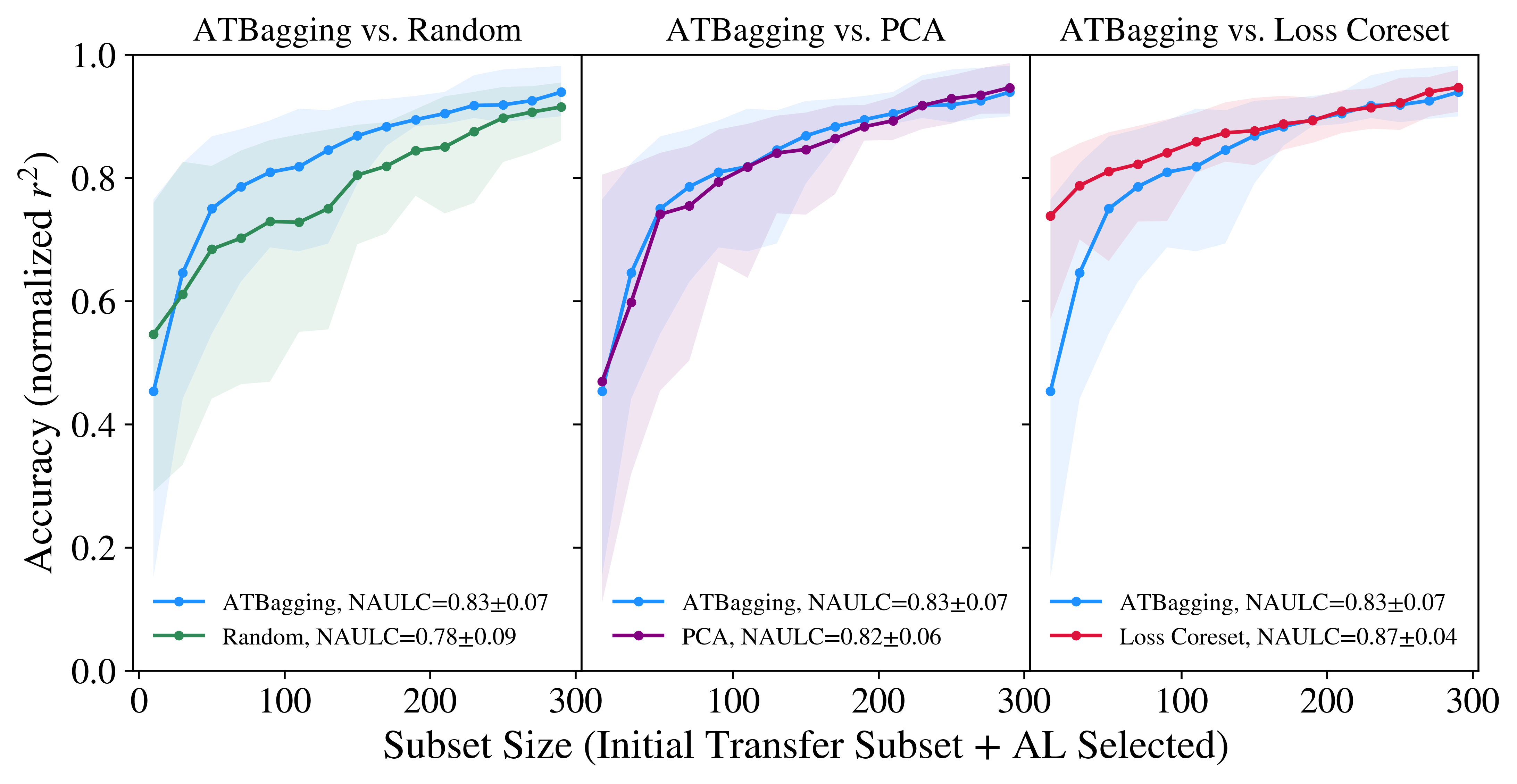}
    \end{subfigure}\hfill
    \begin{subfigure}[t]{0.49\textwidth}
        \caption{}
        \label{fig:forbes_trans:c}
        \centering
        \includegraphics[width=1.0\textwidth]{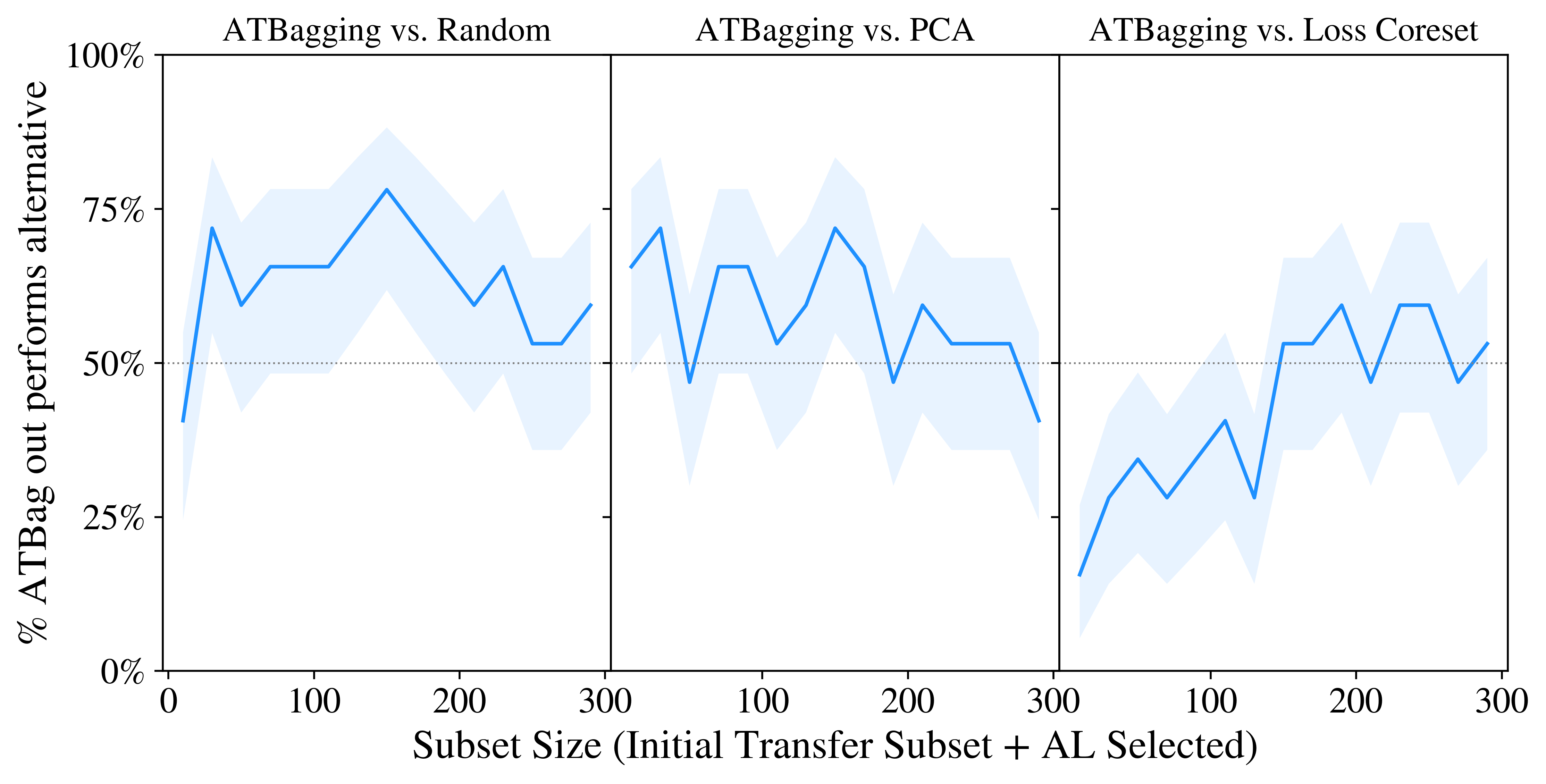}
    \end{subfigure}
    \caption{%
    Transfer and active learning performance for the seed subset creation methods on
    the Forbes 2000 dataset.
    \subref{fig:forbes_trans:a} The initial transfer performance (ITP) of RFR models
    trained on the seed subsets generated by ATBagging and the three alternative methods
    for seed subset sizes of 10, 50, and 100.
    \subref{fig:forbes_trans:b} Active learning curves starting from $n_\text{seed}$ of
    10 growing with $m_\text{collect}$ of 20 to a subset of size 290. Mean accuracy is
    shown as solid lines, with 90\% high density intervals shown by the shaded regions.
    The legend reports the NAULC (mean$\pm$std.dev.) of the curves. 
    \subref{fig:forbes_trans:c} Pairwise model comparisons, the percentage of trials
    from the AL trials in \subref{fig:forbes_trans:b} in which ATBagging outperformed
    the indicated alternative method. 90\% credible intervals are shown in the shaded
    regions.
    }
    \label{fig:forbes_trans}
\end{figure}

\subsubsection{Transfer Seed Subset Performance: Forbes}

The transfer task for the Forbes dataset involves using a model trained on a variety of
indicators about Western companies to select which Asian companies to acquire market
value to train a performant model for the Asian market. In this task there is no overlap
between the source and transfer datasets (no companies are both Western and Asian). This
was the dataset on which ATBagging performed the worst in the downselection task. The
ITP is quite similar between ATBagging and PCA subsets, while ATBagging slightly
outperforms the random subset generation at the two larger subset sizes, as shown in
\Cref{fig:forbes_trans:a}. Loss coreset subsets, on the other hand, is significantly
different at the smallest subset size, where it outperforms all other methods with
a mean accuracy of \textasciitilde{}0.7. However, this advantage is quickly lost at the
two larger subset sizes (\Cref{fig:forbes_trans:a}).


AL performance is dominated by the loss coreset subsets, which exhibit a NAULC of
0.87$\pm$0.04, compared to 0.83$\pm$0.07, 0.82$\pm$0.06, and 0.78$\pm$0.09 for the
ATBagging, PCA, and random subsets (for $n_\text{seed} = $ 10), respectively
(\Cref{fig:forbes_trans:b}). Consistent with the ITP, loss coreset and the random
subsets both initially outperformed those of ATBagging. However, after the first
iteration of active learning ATBagging outperforms the random subset for the rest of the
learning curve. On the other hand, the loss coreset subsets maintain their higher
performance over ATBagging in this dataset until $N_\text{tr} = $ 150 when both methods
converge in accuracy. While the learning curve indicates quite similar performance
between ATBagging and PCA subsets, there is a marginal pairwise performance advantage
for ATBagging until an $N_\text{tr}$ of approximately 200 after which their accuracy
converges.

\subsubsection{Transfer Seed Subset Performance: ERA5}

\begin{figure}[!t]
    \centering
    \begin{subfigure}{0.49\textwidth}
        \caption{}
        \label{fig:era_trans:a}
        \centering
        \includegraphics[width=1.0\textwidth]{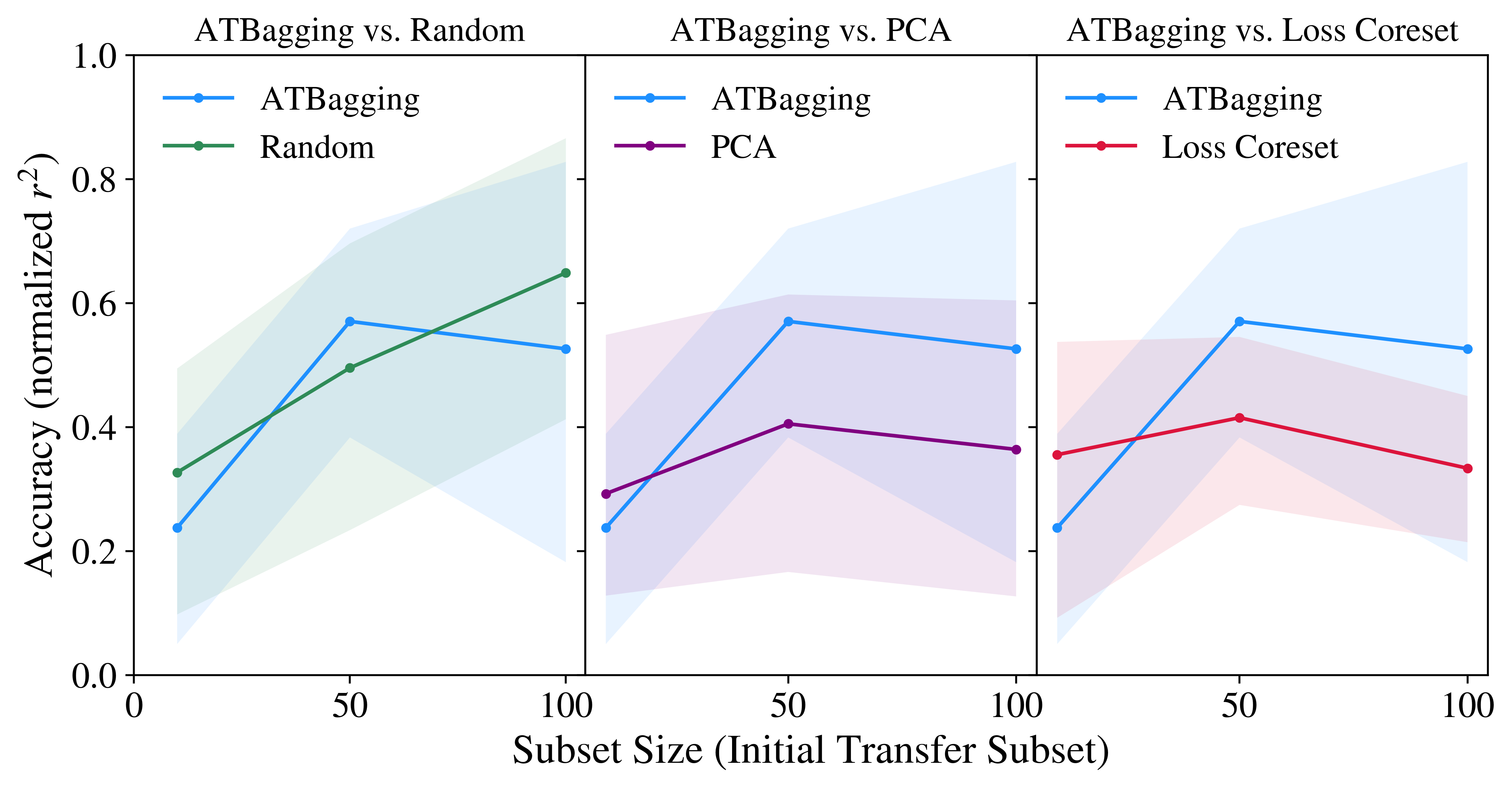}
    \end{subfigure}
    \begin{subfigure}[t]{0.49\textwidth}
        \caption{}
        \label{fig:era_trans:b}
        \centering
        \includegraphics[width=1.0\textwidth]{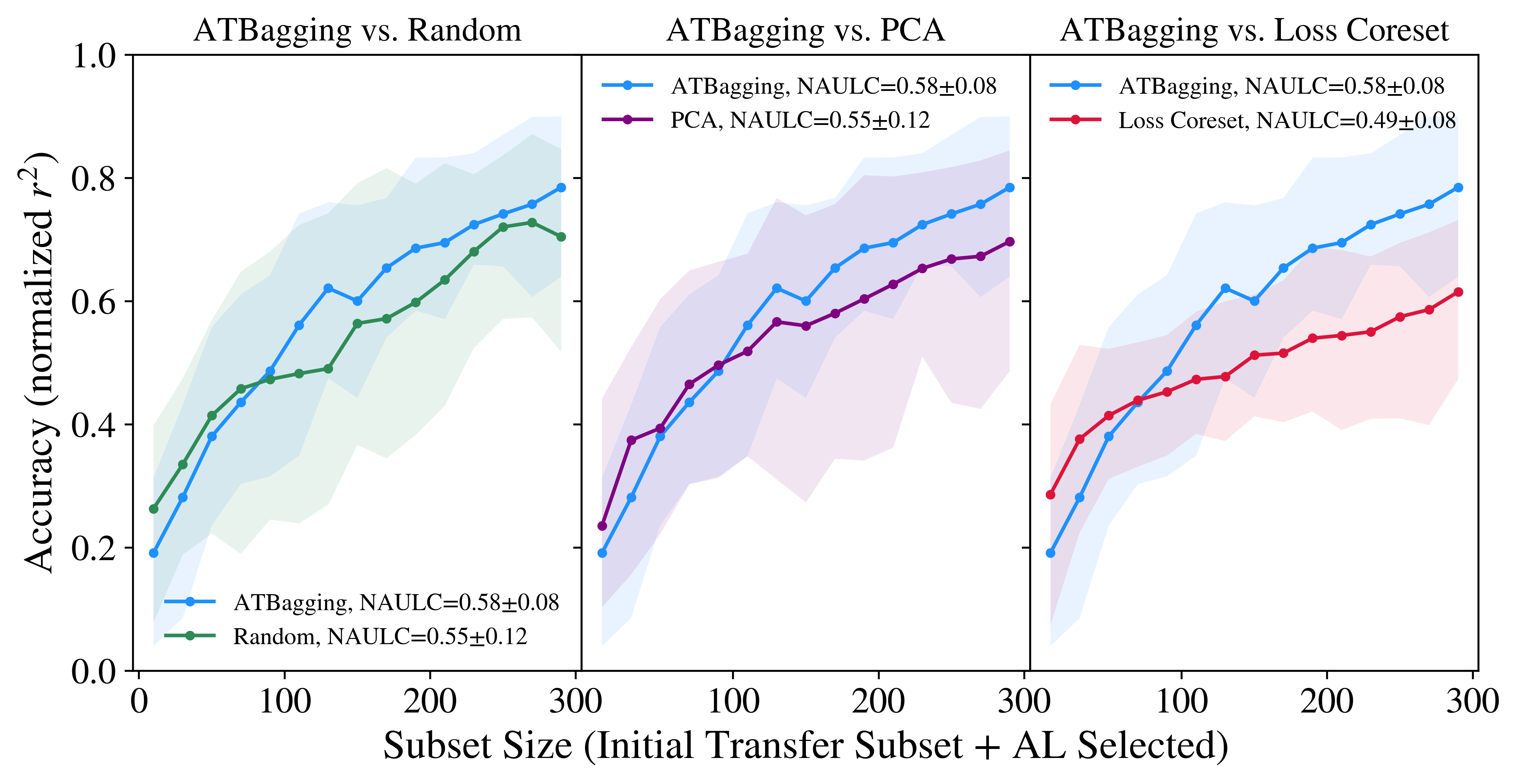}
    \end{subfigure}
    \begin{subfigure}[t]{0.49\textwidth}
        \caption{}
        \label{fig:era_trans:c}
        \centering
        \includegraphics[width=1.0\textwidth]{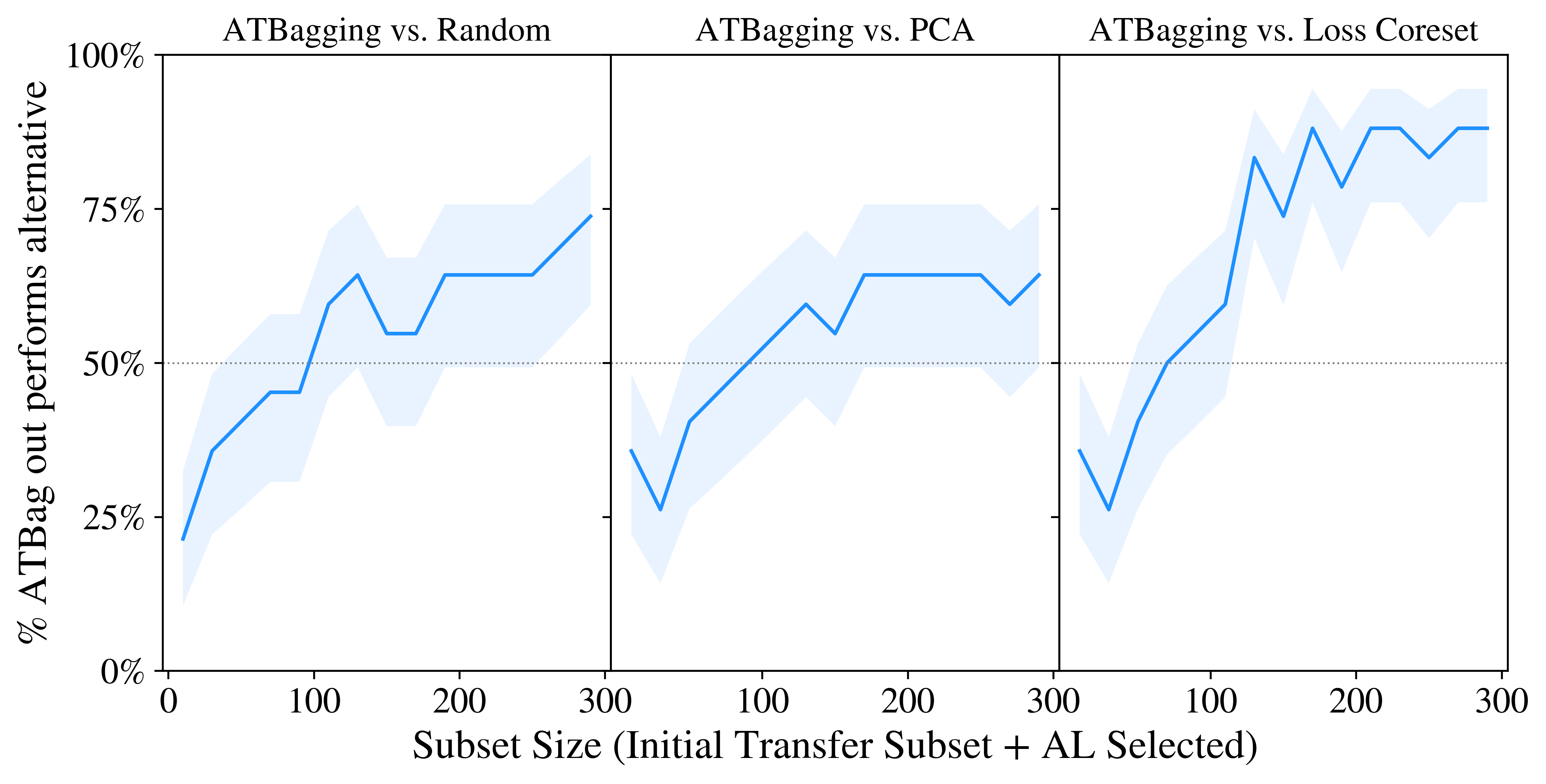}
    \end{subfigure}
    \caption{%
    Transfer and active learning performance for the seed subset creation methods on
    the ERA5 dataset.
    \subref{fig:era_trans:a} The initial transfer performance (ITP) of RFR models
    trained on the seed subsets generated by ATBagging and the three alternative methods
    for seed subset sizes of 10, 50, and 100.
    \subref{fig:era_trans:b} Active learning curves starting from $n_\text{seed}$ of 10
    growing with $m_\text{collect}$ of 20 to a subset of size 290. Mean accuracy is
    shown as solid lines, with 90\% high density intervals shown by the shaded regions.
    The legend reports the NAULC (mean$\pm$std.dev.) of the curves. 
    \subref{fig:era_trans:c} Pairwise model comparisons, the percentage of trials from
    the AL trials in \subref{fig:era_trans:b} in which ATBagging outperformed the
    indicated alternative method. 90\% credible intervals are shown in the shaded
    regions.
    }
    \label{fig:era5_trans}
\end{figure}

The transfer task for the ERA5 weather dataset is to use a dataset of spatial
meteorological data including a target related to precipitation runoff to select
locations for measuring a new precipitation-related target. In this task the source and
transfer domains are identical. In ITP, ATBagging subsets significantly outperform both
PCA and loss coreset subsets at $n_\text{seed} = $ 50 and 100, though it loses to all
three competitors when the transferred subset only has $n_\text{seed} = $ 10. At the
random subset outperforms ATBagging subsets at all but $n_\text{seed} = $ 50
(\Cref{fig:era_trans:a}).

AL performance initially mirrors that of ITP, where ATBagging starts with an accuracy of
\textasciitilde{}0.2, below all other methods. However, in all cases, ATBagging
surpasses the competitor methods in accuracy by the time the AL subset reaches
$N_\text{tr} = $ 100, after which it maintains superior performance in all three
comparisons, as shown in \Cref{fig:era_trans:b}. NAULC values reflect this trend,
with ATBagging subsets scoring the highest at 0.58$\pm$0.08, with random, PCA, and loss
coreset subsets following with 0.55$\pm$0.12, 0.55$\pm$0.12, and 0.49$\pm$0.08,
respectively (\Cref{fig:era_trans:b}). The pairwise comparisons
(\Cref{fig:era_trans:c}) demonstrate the superiority of ATBagging for this problem
type. The ATBagging subsets outperform those of the loss coreset method 60--85\% of the
time after $N_\text{tr} > $ 100, as shown in \Cref{fig:era_trans:c}. ATBagging
subsets have a lesser, but still notable, advantage against the random and PCA-generated
subsets of approximately 60\% after $N_\text{tr} > $ 100. 

\FloatBarrier
\subsection{Discussion}

Our analysis clearly shows the superior performance of ATBagging subsets for
transfer-active learning, especially in target-transfer types of problems (QM9 and ERA5
datasets). Notably, the advantage of our method does not end with its superior
downselection performance but extends to advantages in active transfer learning tasks as
well, with model accuracy consistently higher even after tens of AL iterations. On
feature shift problems, even when initial transfer performance is outclassed by other
methods, the ATBagging subset can match or out-perform the accuracy of other subsets
within as few as one AL iteration (e.g. Forbes dataset). 

\section{Conclusions}

\begin{table*}
	\caption{Comparison of Method Performances on Transfer Learning Seed Creation}
	\centering
    \begin{tabularx}{\textwidth}{
        >{\raggedright\arraybackslash}m{0.12\textwidth}| 
        >{\raggedright\arraybackslash}m{0.12\textwidth}| 
        >{\raggedright\arraybackslash}X                 
        >{\raggedright\arraybackslash}m{0.11\textwidth} 
        >{\raggedright\arraybackslash}m{0.11\textwidth} 
        >{\raggedright\arraybackslash}m{0.11\textwidth} 
        >{\raggedright\arraybackslash}m{0.11\textwidth} 
    }
		\toprule
        Dataset & Subset Size ($n_\text{seed}$) & Metric & Random & PCA & Loss Coreset
        & ATBagging \\
		\midrule
        QM9 &  10 & ITP   & 0.31 & 0.35 & 0.28 & \textbf{0.40} \\
            &     & NAULC & 0.61 & 0.65 & 0.63 & \textbf{0.66} \\
            &  50 & ITP   & 0.56 & 0.56 & 0.40 & \textbf{0.60} \\
            &     & NAULC & 0.70 & 0.68 & 0.65 & \textbf{0.71} \\
            & 100 & ITP   & 0.63 & 0.64 & 0.56 & \textbf{0.66} \\
            &     & NAULC & 0.67 & 0.67 & 0.60 & \textbf{0.68} \\
        PM\textsubscript{2.5} &  10 & ITP   & 0.18 & 0.23 & 0.23 & \textbf{0.28} \\
                              &     & NAULC & 0.68 & 0.68 & 0.61 & \textbf{0.72} \\
                              &  50 & ITP   & \textbf{0.52} & 0.40 & 0.33 & 0.36 \\
                              &     & NAULC & \textbf{0.72} & 0.64 & 0.69 & 0.71 \\
                              & 100 & ITP   & 0.68 & 0.67 & 0.63 & \textbf{0.71} \\
                              &     & NAULC & \textbf{0.80} & 0.60 & 0.56 & 0.66 \\
        Forbes &  10 & ITP   & 0.51 & 0.44 & \textbf{0.68} & 0.42 \\
               &     & NAULC & 0.72 & 0.76 & \textbf{0.81} & 0.77 \\
               &  50 & ITP   & 0.55 & 0.56 & 0.50 & \textbf{0.58} \\
               &     & NAULC & 0.61 & 0.61 & 0.59 & \textbf{0.63} \\
               & 100 & ITP   & 0.61 & 0.62 & 0.59 & \textbf{0.64} \\
               &     & NAULC & 0.62 & 0.64 & 0.64 & \textbf{0.64} \\
        ERA5 &   10 & ITP   & 0.33 & 0.29 & \textbf{0.36} & 0.24 \\
             &      & NAULC & 0.68 & 0.68 & 0.61 & \textbf{0.72} \\
             &   50 & ITP   & 0.50 & 0.41 & 0.42 & \textbf{0.57} \\
             &      & NAULC & \textbf{0.72} & 0.64 & 0.69 & 0.71 \\
             &  100 & ITP   & \textbf{0.65} & 0.36 & 0.33 & 0.53 \\
             &      & NAULC & \textbf{0.80} & 0.60 & 0.56 & 0.66 \\
        \midrule
        \multicolumn{3}{r}{Wins} & 6 & 0 & 3 & 15 \\
		\bottomrule
	\end{tabularx}
	\label{tab:result comparison}
\end{table*}

Herein we have developed and demonstrated a novel ATBagging method; by maximizing both
the informativeness and heterogeneity of data subsets in downselection and
transfer-active learning, ATBagging achieves superior performance in tests on four
real-world datasets in both feature-shift and target-transfer type tasks. 

Across the tests, the subsets produced by ATBagging were found to be more representative
of the source datasets than those of the alternative methods, which acted as more
performant seeds in transfer active learning. In downselection of the representative
subsets it retained higher accuracy, especially at small $N_\text{tr}$, than alternative
methods, indicating that the combination of informativeness \& feature-space
heterogeneity is a strong avenue for dataset distillation. 

In transfer-active learning, ATBagging-derived subsets generally demonstrated improved
ITP and NAULC, even in cases where it was not the best performing method with low data,
it overtook the alternatives within the first 100-150 acquisitions. This suggests that
not only does ATBagging generate small $n$ model subsets whose accuracy is generally
high, but that it provides excellent starting ground for active learning. Even when the
accuracy of predictions based on ATBagging seed data is lower than alternatives, the
uncertainty estimates still outperform those of the more accurate same-$n$ models in
determining which points to select next. 

Overall, these results show that ATBagging is a practical methodology for the
transfer-active learning task, especially in scenarios where the cost of dataset
labelling is expensive, forcing the problem into the low n regime where it performs most
favorably.

\FloatBarrier
\newpage

\section*{Funding Sources}

The Authors gratefully acknowledge several funding sources which supported the authors
contributing to this cross-project collaborative effort. VP acknowledges support from
the Science and Technologies for Phosphorus Sustainability (STEPS) Center, a National
Science Foundation Science and Technology Center (CBET-2019435). DR acknowledges support
from the National Alliance for Water Innovation (NAWI), funded by the U.S. Department of
Energy, Office of Energy Efficiency and Renewable Energy (EERE), Industrial Technologies
Office (ITO), under Funding Opportunity Announcement DE-FOA-0001905 and from the U.S.
National Science Foundation CBET Catalysis under award 2450869. ON acknowledges support
from the National Institute of Environmental Health Sciences of the National Institutes
of Health under Award Number P42ES030990 as part of the MEMCARE (Metals Mixtures:
Cognitive Aging, Remediation, and Exposure Sources) project and the National Science
Foundation Nanosystems Engineering Research Center for Nanotechnology Enabled Water
Treatment (NEWT EEC 1449500). SW acknowledge support from the U.S. Department of
Energy's Energy Efficiency \& Renewable Energy office under Award Number
DE-EEDE-EE0010732 and the U.S. Department of Energy, Office of Science under Federal
Award Identification Number DE-SC0024724. CM acknowledge support from the U.S.
Department of Energy, Office of Science, Office of Basic Energy Sciences, under Award
Number(s) DE-SC0024194. The content is solely the authors' responsibility and does not
necessarily represent the official views of the National Institutes of Health, US
Department of Energy, or the NSF. In addition, we acknowledge support from Research
Computing at Arizona State University for providing high-performance supercomputing
services.\citep{HPC:ASU23}

\section*{Code Availability}

A python package which implements a Scikit-learn compatible implementation of the
ATBagging methodology is available on github at:\\
\url{https://github.com/MuhichLab/active_transfer_bagging}

\section*{Corresponding Author Information}

Corresponding Authors: Christopher Muhich\\
Email: \texttt{cmuhich@asu.edu}\\
Telephone: 480-965-2673\\
Address: 551 E. Tyler Mall, ERC 257, Arizona State University, Tempe AZ\\

\newpage

\bibliographystyle{unsrtnat}

\newpage

\onecolumn

\appendix
\crefalias{section}{appendix}
\crefname{appendix}{appendix}{appendices}
\Crefname{appendix}{Appendix}{Appendices}

\section{In-Bag/Out-of-Bag Model Availability}
\label{sec:appendixA}

Following from the definition of a bagged ensemble model as a model
$\mathcal{M}=\{m_i\}_i^M$, comprised of $M$ submodels of weak learners, $m_i$, trained
on bootstrapped samples of the training dataset, the probability of a data point's
exclusion from a bootstrapped resampling is $\frac{1}{e}$; therefore, the probability of
both in-bag and out-of-bag models existing for every point in the dataset of size $N$ is
$\left(1-\left(1-\frac{1}{e}\right)^M\right)^N$. Thus, even assuming a massive
one-million-point data set, the generation of 50 models means that there is only a 1 in
10,000 chance of even a single data point not being present in at least one in- and one
out-of-bag model. If 100 models are constructed, a data set of 10 quadrillion points
would be necessary to have the same, still tiny, 1 in 10,000 chance of not having both
in- and out-of-bag models. 

\section{Information Gain via Bayesian Interpretation of Bagging Models}
\label{sec:appendixB}

Assuming a Bayesian context, where model parameters are treated as random variables with
distribution $p(\theta)$, the predictive distributions are expressed via the integral

\begin{equation*}
p(Y_* | X_*) = \int{p(Y_* | X_*, \theta)p(\theta)d\theta}
\end{equation*}

In regression, it is reasonable to partition uncertainty such that all epistemic
uncertainty is placed on the parameters, i.e. that the distribution over parameters
represents the uncertainty in their true values, and all aleatoric uncertainty is placed
on the model predictions, i.e. that the data generation process is noisy. This
partitioning is achieved by imposing a Gaussian noise model over model predictions via
$p(Y_*|X_*,\theta)\sim\text{MVN}(m_\theta(X_*),\sigma^2 I_n)$, where the mean vector is
the vector of model predictions over the input set $X_*$ and $\sigma^2$ is the
observational noise, a hyperparameter. The integral then becomes,

\begin{equation*}
p(Y_*|X_*) = \int{\text{MVN}(m_\theta(X_*),\sigma^2)p(\theta)d\theta}
\end{equation*}

With a bagged ensemble, the model parameters for in-bag models, $\theta_\text{ib}$, are
approximately drawn from the posterior parameter distribution, $\theta_\text{ib}\sim
p(\theta|x,y)$, while model parameters for out-of-bag models $\theta_\text{oob}$ are
approximately drawn from the prior distribution, $\theta_\text{oob}\sim p(\theta)$.
Thus, the above integrals may be approximated via a Monte Carlo expectation

\begin{equation*}
\begin{aligned}
p(Y_*|X_*) &\approx \frac{1}{|\mathcal{M}_\text{oob}|} 
\sum_{m\in\mathcal{M}_\text{oob}}{\text{MVN}(m(X_*),\sigma^2 I_n)} \\
p(Y_*|X_*,x,y) &\approx \frac{1}{|\mathcal{M}_\text{ib}|} 
\sum_{m\in\mathcal{M}_\text{ib}}{\text{MVN}(m(X_*),\sigma^2 I_n)} \\
\end{aligned}
\end{equation*}

These approximations are both mixtures of multivariate Gaussians, which makes
calculating their KL divergence intractable. Therefore, the mixture distributions are
approximated via a single multivariate Gaussian with matched moments,  

\begin{equation*}
\begin{aligned}
p(Y_* | X_*) &\approx \frac{1}{|\mathcal{M}_\text{ib/oob}|}
    \sum_{m\in\mathcal{M}_\text{ib/oob}}{\text{MVN}(m(X_*),\sigma^2 I)}
    \approx \text{MVN}(\mu,\Sigma) \\
\mu &= \frac{1}{|\mathcal{M}_\text{ib/oob}|} 
    \sum_{m\in\mathcal{M}_\text{ib/oob}}{m(X_*)} \\
\Sigma &= \sigma^2 I_n + \frac{1}{|\mathcal{M}_\text{ib/oob}|} 
\sum_{m\in\mathcal{M}_\text{ib/oob}}{\left(m(X_*)-\mu\right)\left(m(X_*)-\mu\right)^\top} \\
\end{aligned}
\end{equation*}

With this approximation, the KL divergence and thus information gain upon including data
point $(x,y)$ is expressible analytically as: 

\begin{equation*}
\text{KL}\left(p(Y_*|X_*,x,y) \parallel p(Y_*|X_*)\right)
=
\frac{1}{2}\lbrack
    \tr(\Sigma^{-1}_\text{oob}\Sigma_\text{ib})
    + (\mu_\text{oob} - \mu_\text{ib})^\top
    \Sigma^{-1}_\text{oob}
    (\mu_\text{oob}-\mu_\text{ib})
    - n
    - \ln{\frac{\det{\Sigma_\text{oob}}}{\det{\Sigma_\text{ib}}}}
\rbrack
\end{equation*}

This KL divergence is calculated for every point in the training set, the magnitude of
which defines its informativeness.

\section{Approximate Sampling of Ill-Conditioned $k$-DPPs}
\label{sec:appendixC}

In applications where our approach is attractive it is assumed that acquiring labels for
the seed subset is either expensive or will take a large amount of time, and therefore
budgeting subset acquisition in terms of a predetermined subset size is desirable. This
poses a problem with utilizing DPPs for the subset creation, as a sample from a DPP is
a subset of random size. This problem of fixed sample size is addressed via $k$-DPPs,
a method of sampling from a DPP conditional on a fixed sample size $k$.
\citep{Kulesza2011} However, as the size of the dataset increases problems may arise
with the use $k$-DPPs, since for rank-deficient $L$-matrices sampling from a $k$-DPP may
fail for even small $k$, while the sampling process itself, which depends on an
eigendecomposition of the $L$-matrix, may become computationally prohibitive.

The latter of these problems has been addressed by fast DPP sampling algorithms which
exploit the RFF decomposition of the kernel from which the $L$ matrix is constructed,
which can be applied for both $k$-DPP and normal DPP sampling.\citep{Barthelm2019} The
first of these issues, the rank deficiency of $L$ which is observed in many real
applications such as those utilized as test cases in this work, remains an issue.

To address these issues, we propose the following extension to Algorithm 3 of
\citet{Tremblay2018} and Algorithm 1 of \citet{Kulesza2012}, where we scale the $L$
matrix such that the expectation of the sample size of the $L$-ensemble is equal to our
desired sample size. This is done by solving for the scale factor, $a$, in the equation,

\begin{equation*}
\mathbb{E}[|\mathcal{S}|] = \sum_i{\frac{a\lambda_i}{1+a\lambda_i}}
\end{equation*}

with the expected sample size $\mathbb{E}[|\mathcal{S}|]$ set to the desired sample
size, where $\lambda_i$ are the eigenvalues of the $L$ matrix. This must be done
numerically, which we do via Brent’s root finding method.\citep{Brent1973} With the
scaled eigenvectors of $L$, they are sampled according to Algorithm 3 and used to
construct a projective $L$-ensemble which may be sampled using a standard technique.

This approach is robust to ill-conditioned $L$ matrices but does come with some
drawbacks. As the subset size is still random, there is no guarantee that the first
attempt will generate a sample of the desired size, and therefore multiple samples are
sometimes required. Additionally, uniform scaling of the $L$ matrix may modify the
marginal inclusion probabilities of each point in complicated ways. In practice this was
not seen to be an issue, but an improved $k$-DPP sampling method which can handle
ill-conditioned $L$ matrices is desired for future research.

\newpage
\section{Supllemental Transfer Learning Figures}
\label{sec:appendixD}

\subsection*{Transfer Performance -- ERA5: Subset Size 50}

\begin{figure}[H]
    \centering
    \begin{subfigure}[t]{0.49\textwidth}
        \centering
        \includegraphics[width=1.0\linewidth]{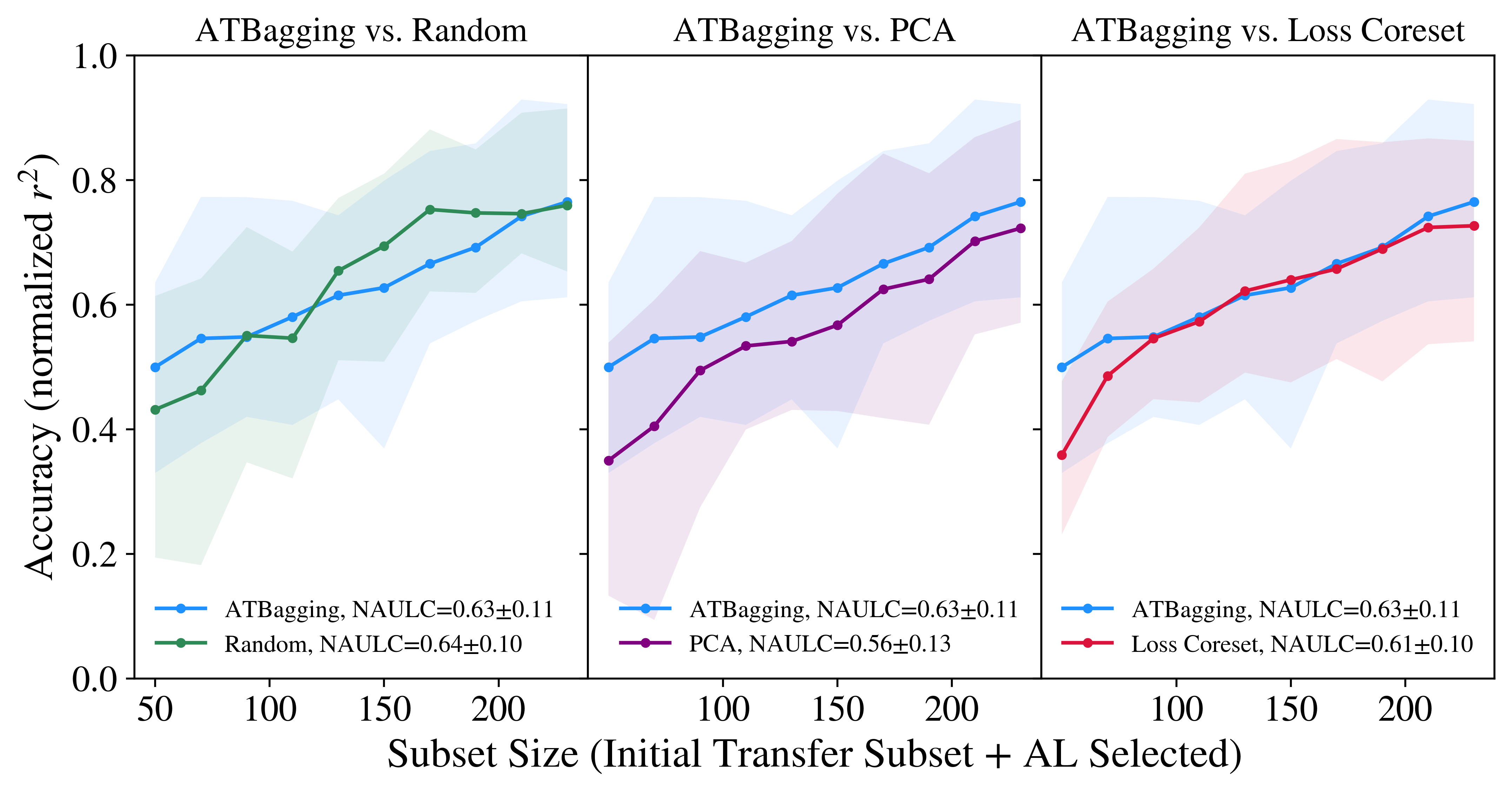}
        \caption{ERA5 dataset transfer performance with initial subset size of 50.}
        \label{fig:era_trans_acc_50}
    \end{subfigure}\hfill
        \begin{subfigure}[t]{0.49\textwidth}
        \centering
        \includegraphics[width=1.0\linewidth]{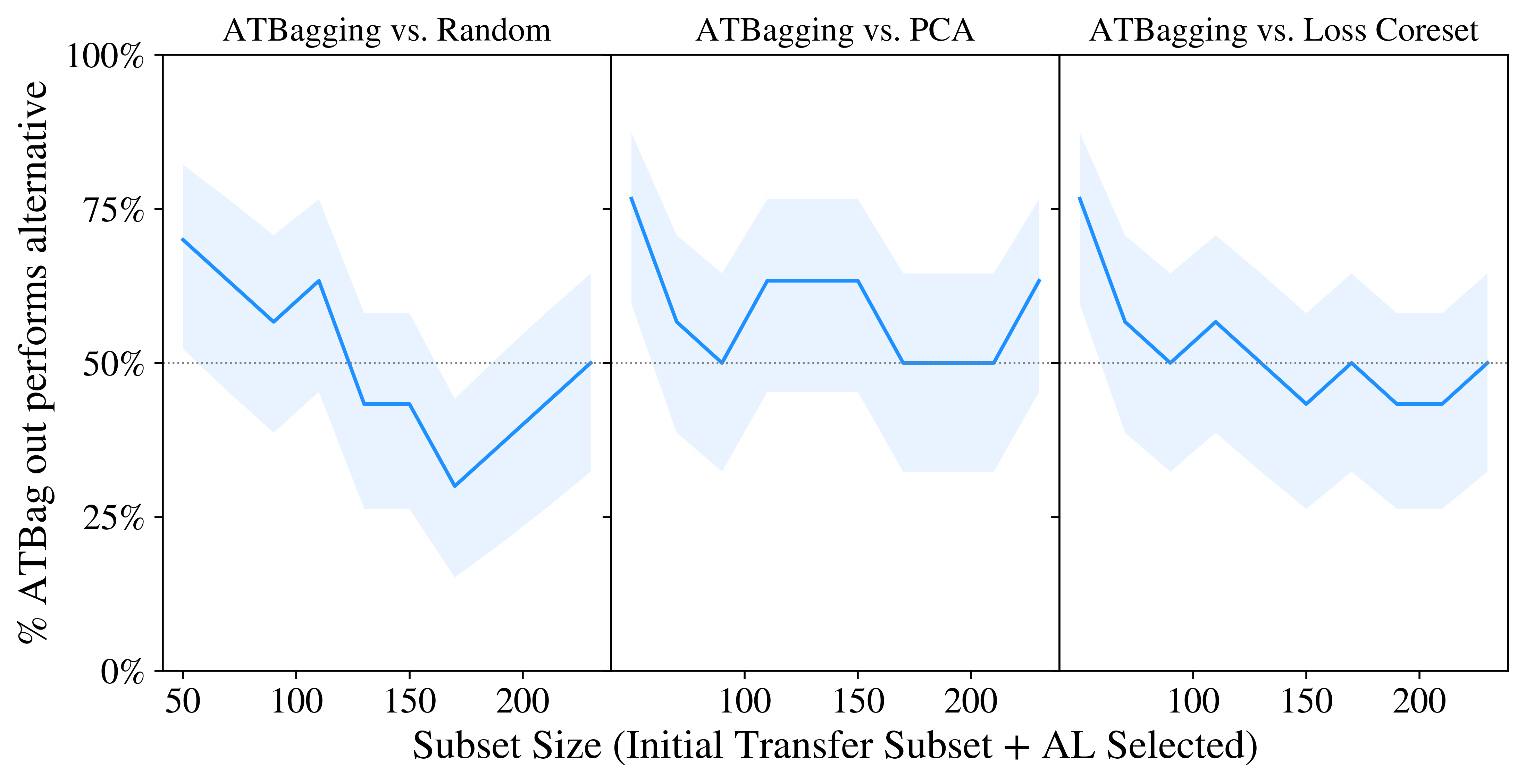}
        \caption{%
        ERA5 dataset transfer performance pairwise method comparisons with initial
        subset size of 50.
        }
        \label{fig:era_trans_perc_50}
    \end{subfigure}
\end{figure}

\subsection*{Transfer Performance -- ERA5: Subset Size 100}

\begin{figure}[H]
    \centering
    \begin{subfigure}[t]{0.49\textwidth}
        \centering
        \includegraphics[width=1.0\linewidth]{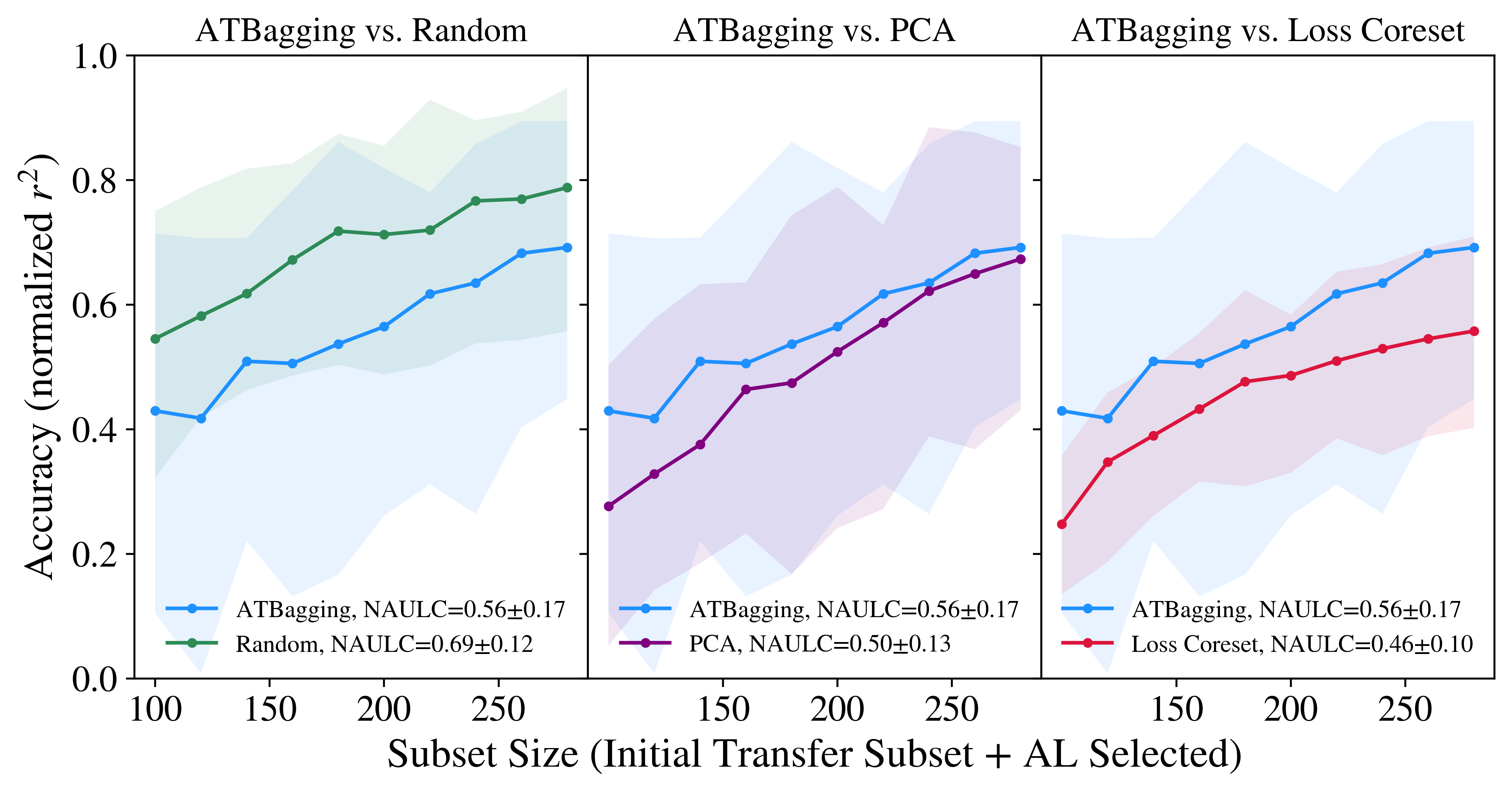}
        \caption{ERA5 dataset transfer performance with initial subset size of 100.}
        \label{fig:era_trans_acc_100}
    \end{subfigure}\hfill
        \begin{subfigure}[t]{0.49\textwidth}
        \centering
        \includegraphics[width=1.0\linewidth]{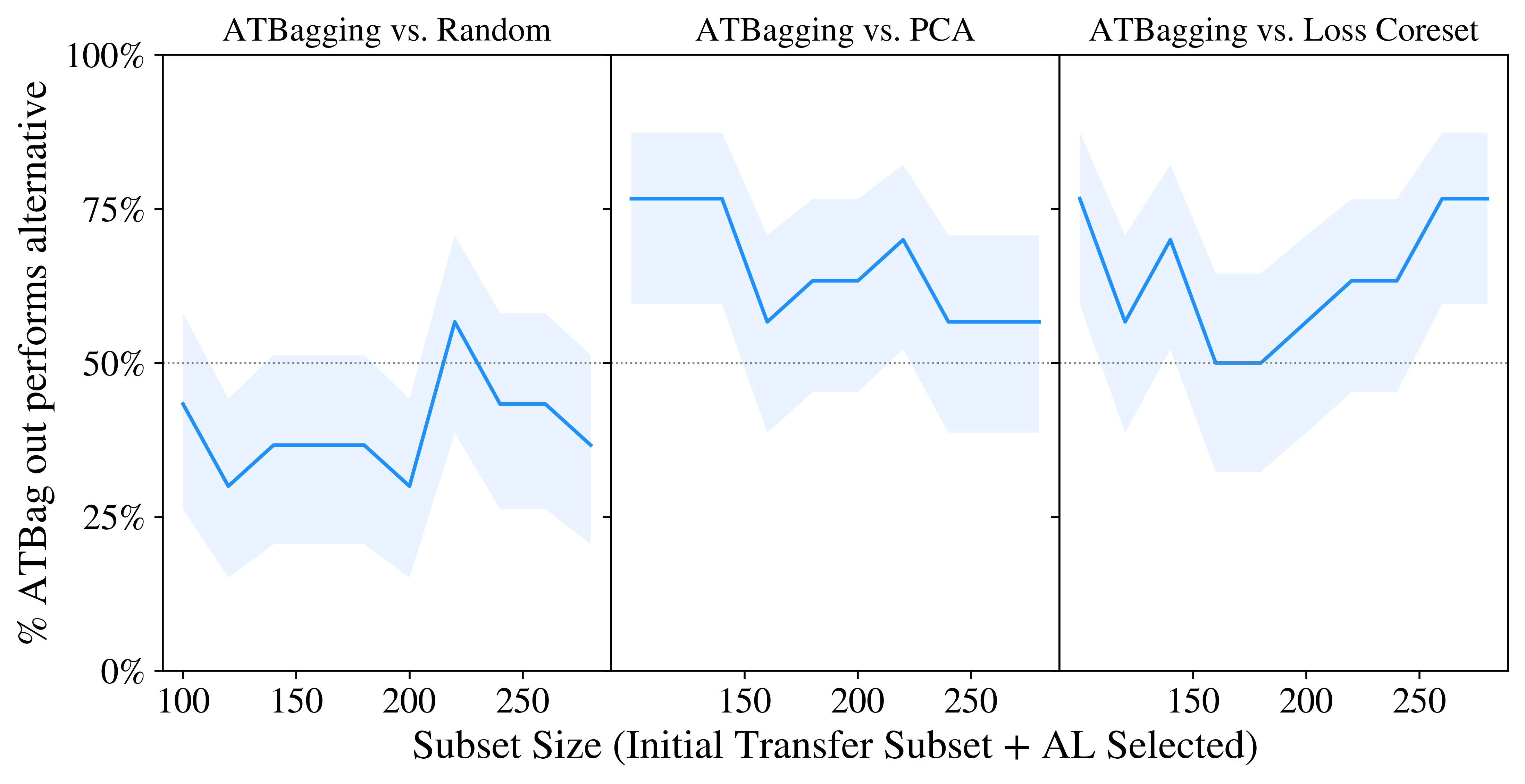}
        \caption{%
        ERA5 dataset transfer performance pairwise method comparisons with initial
        subset size of 100.
        }
        \label{fig:era_trans_perc_100}
    \end{subfigure}
\end{figure}

\subsection*{Transfer Performance -- QM9: Subset Size 50}

\begin{figure}[H]
    \centering
    \begin{subfigure}[t]{0.49\textwidth}
        \centering
        \includegraphics[width=1.0\linewidth]{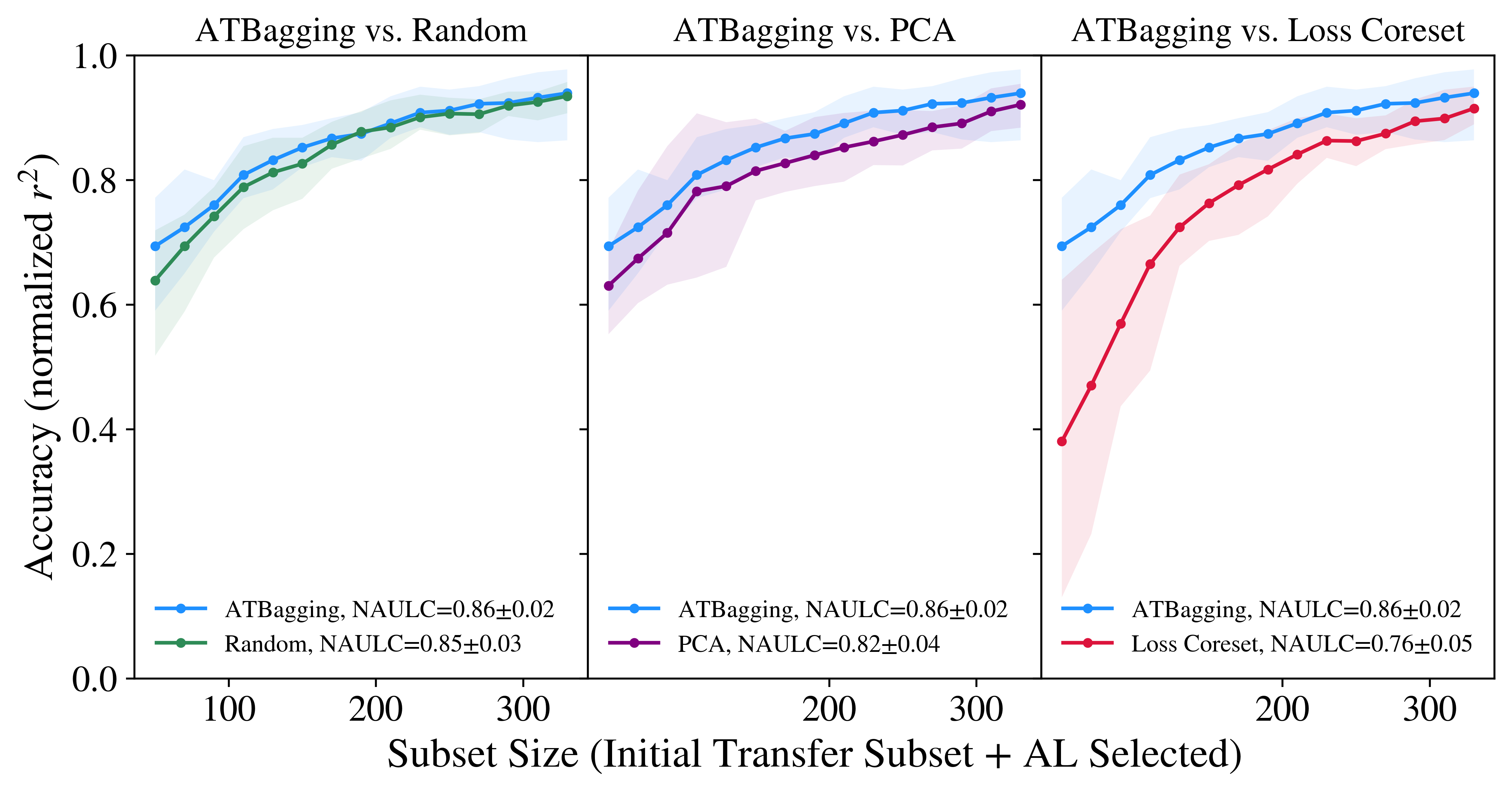}
        \caption{QM9 dataset transfer performance with initial subset size of 50.}
        \label{fig:qm9_trans_acc_50}
    \end{subfigure}\hfill
        \begin{subfigure}[t]{0.49\textwidth}
        \centering
        \includegraphics[width=1.0\linewidth]{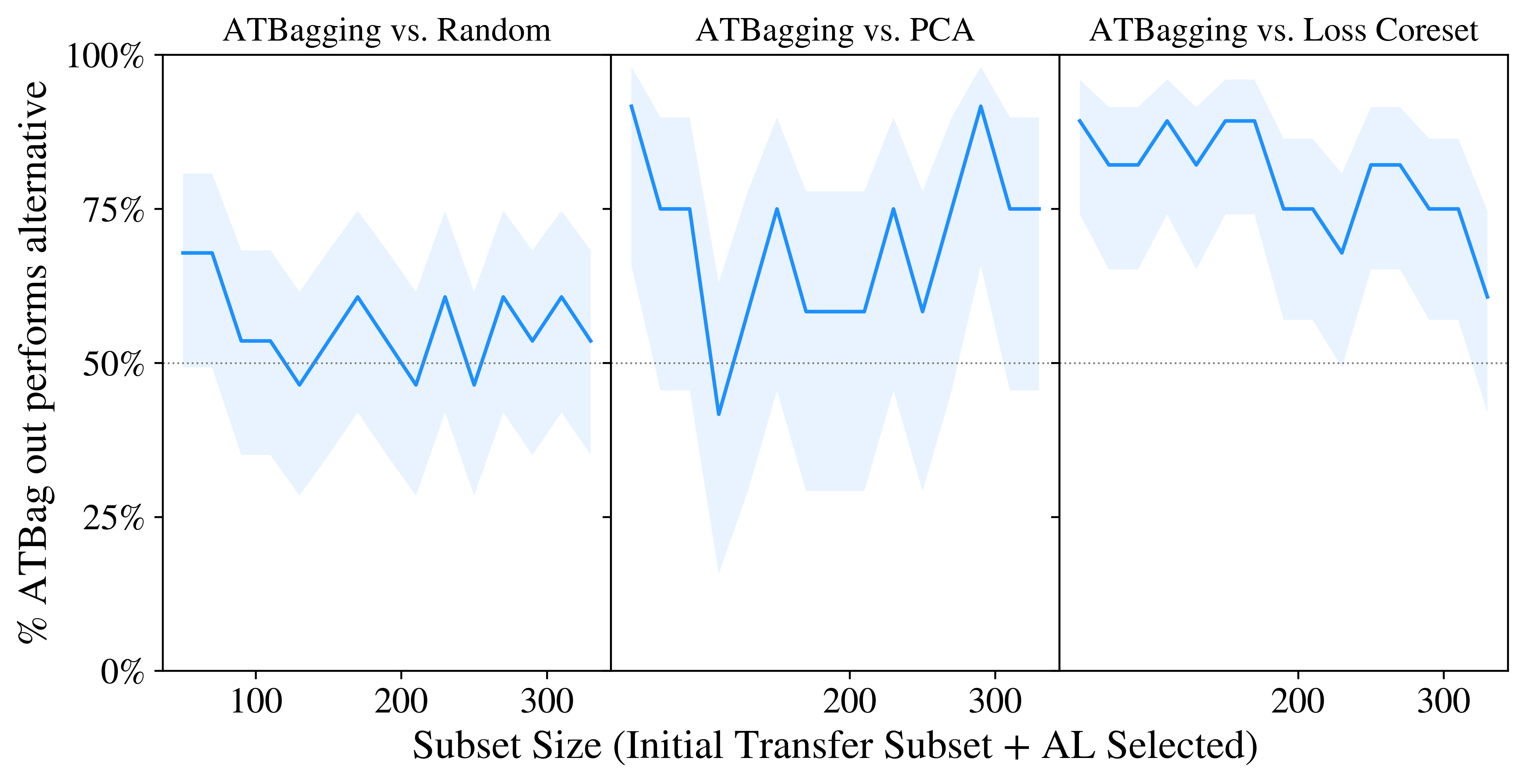}
        \caption{%
        QM9 dataset transfer performance pairwise method comparisons with initial
        subset size of 50.
        }
        \label{fig:qm9_trans_perc_50}
    \end{subfigure}
\end{figure}

\subsection*{Transfer Performance -- QM9: Subset Size 100}

\begin{figure}[H]
    \centering
    \begin{subfigure}[t]{0.49\textwidth}
        \centering
        \includegraphics[width=1.0\linewidth]{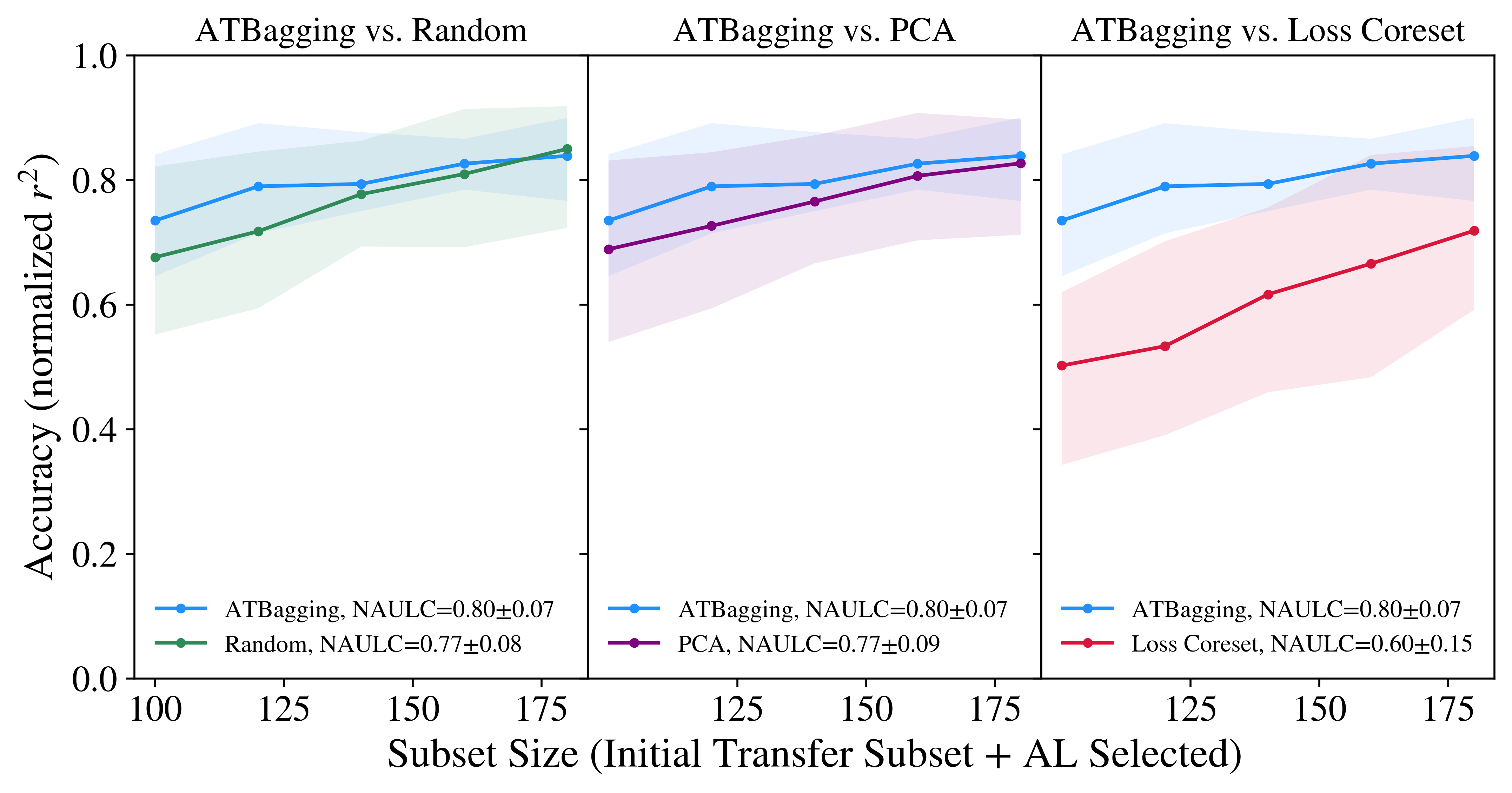}
        \caption{QM9 dataset transfer performance with initial subset size of 100.}
        \label{fig:qm9_trans_acc_100}
    \end{subfigure}\hfill
        \begin{subfigure}[t]{0.49\textwidth}
        \centering
        \includegraphics[width=1.0\linewidth]{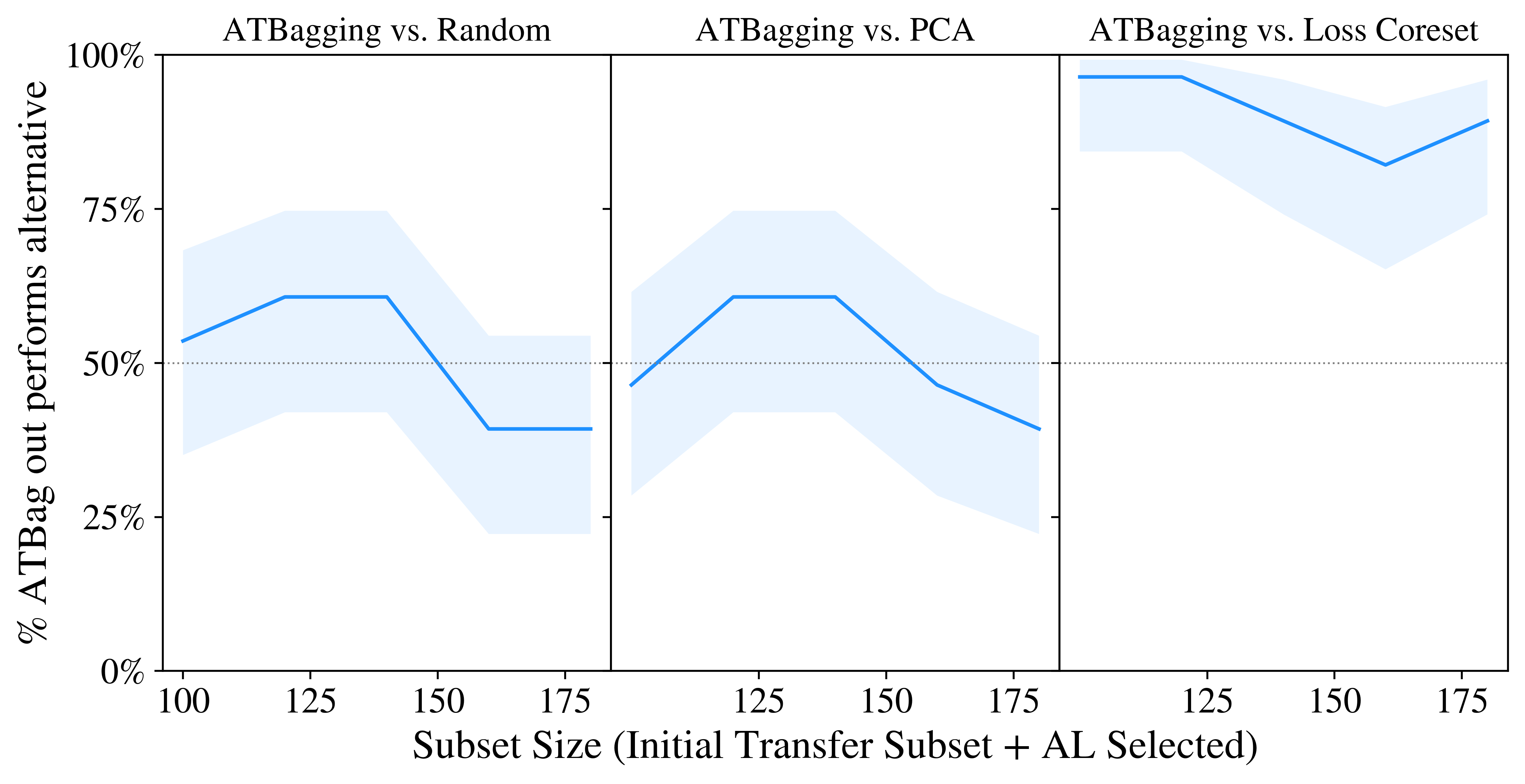}
        \caption{%
        QM9 dataset transfer performance pairwise method comparisons with initial
        subset size of 100.
        }
        \label{fig:qm9_trans_perc_100}
    \end{subfigure}
\end{figure}

\subsection*{Transfer Performance -- PM\textsubscript{2.5}: Subset Size 50}

\begin{figure}[H]
    \centering
    \begin{subfigure}[t]{0.49\textwidth}
        \centering
        \includegraphics[width=1.0\linewidth]{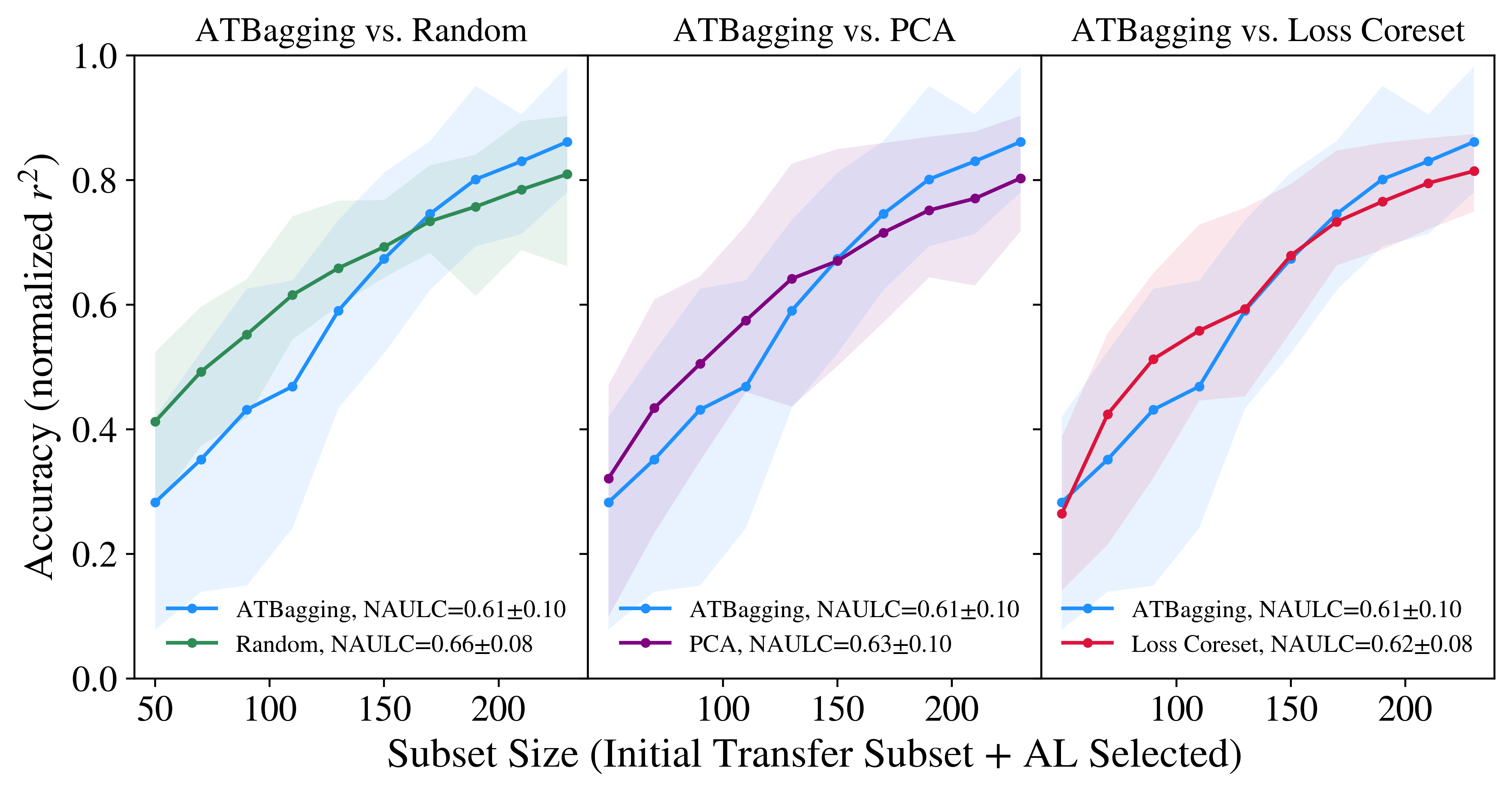}
        \caption{
        PM\textsubscript{2.5} dataset transfer performance with initial subset size
        of 50.
        }
        \label{fig:pm25_trans_acc_50}
    \end{subfigure}\hfill
        \begin{subfigure}[t]{0.49\textwidth}
        \centering
        \includegraphics[width=1.0\linewidth]{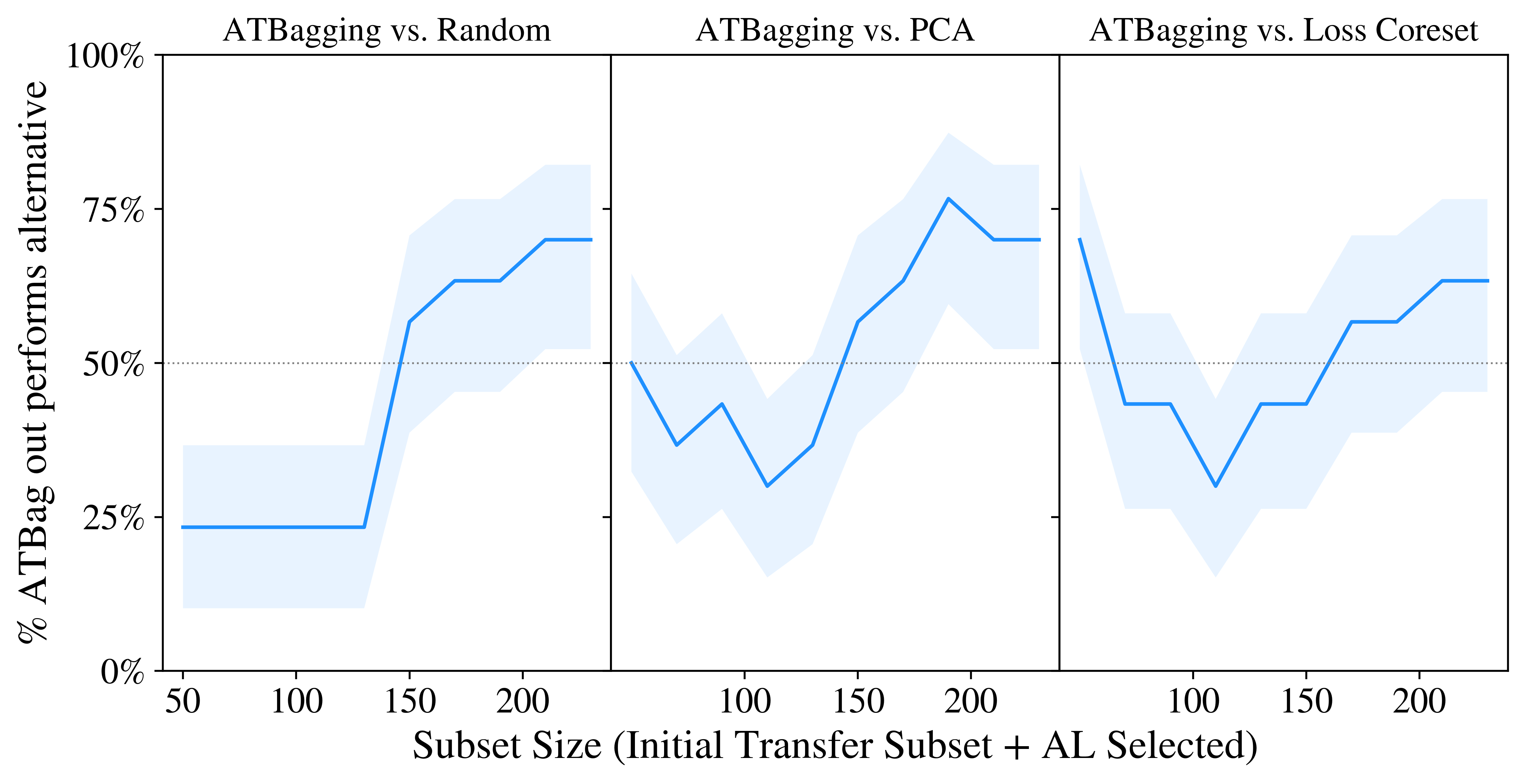}
        \caption{%
        PM\textsubscript{2.5} dataset transfer performance pairwise method comparisons
        with initial subset size of 50.
        }
        \label{fig:pm25_trans_perc_50}
    \end{subfigure}
\end{figure}

\subsection*{Transfer Performance -- PM\textsubscript{2.5}: Subset Size 100}

\begin{figure}[H]
    \centering
    \begin{subfigure}[t]{0.49\textwidth}
        \centering
        \includegraphics[width=1.0\linewidth]{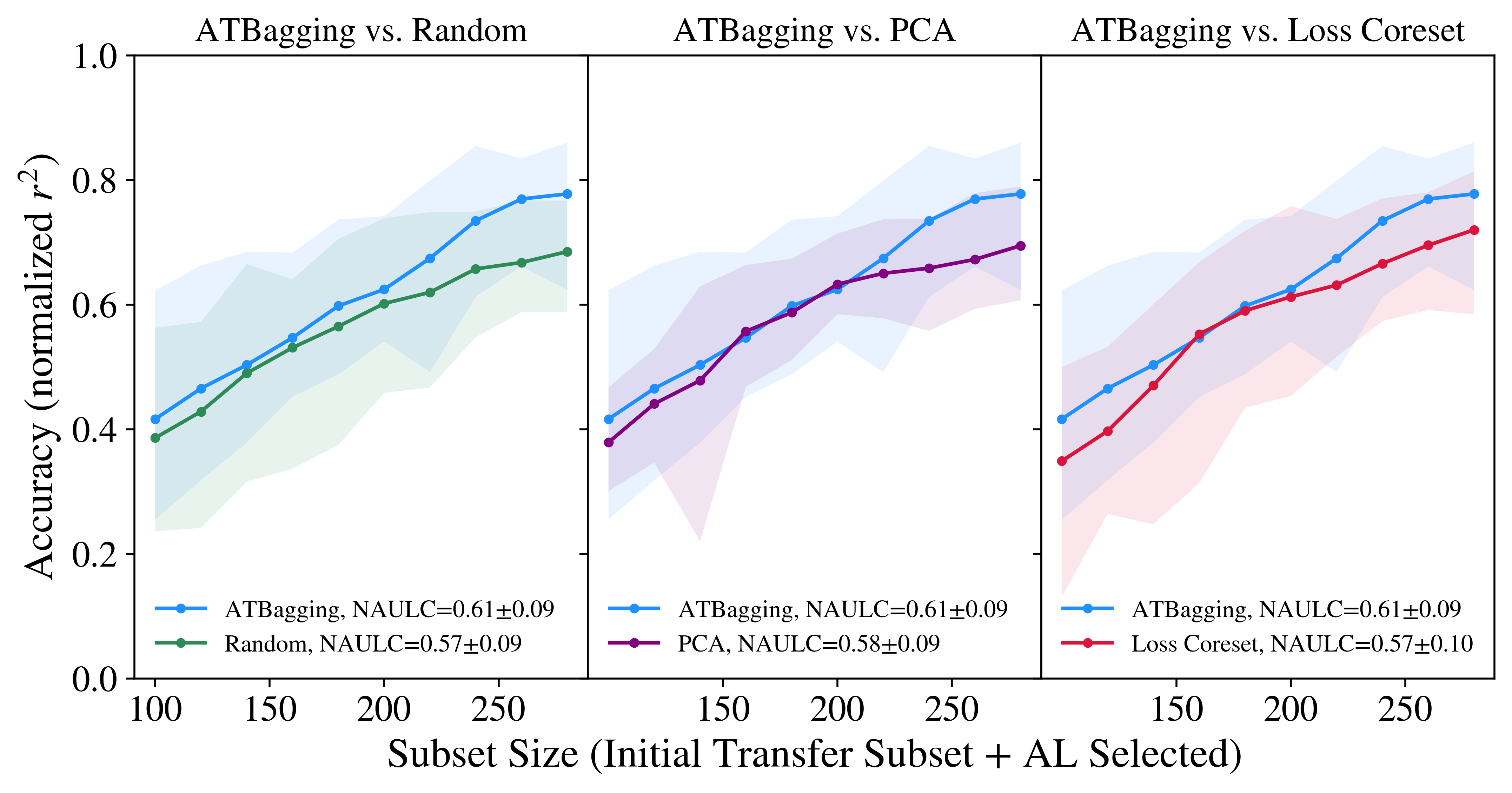}
        \caption{
        PM\textsubscript{2.5} dataset transfer performance with initial subset size
        of 100.
        }
        \label{fig:pm25_trans_acc_100}
    \end{subfigure}\hfill
        \begin{subfigure}[t]{0.49\textwidth}
        \centering
        \includegraphics[width=1.0\linewidth]{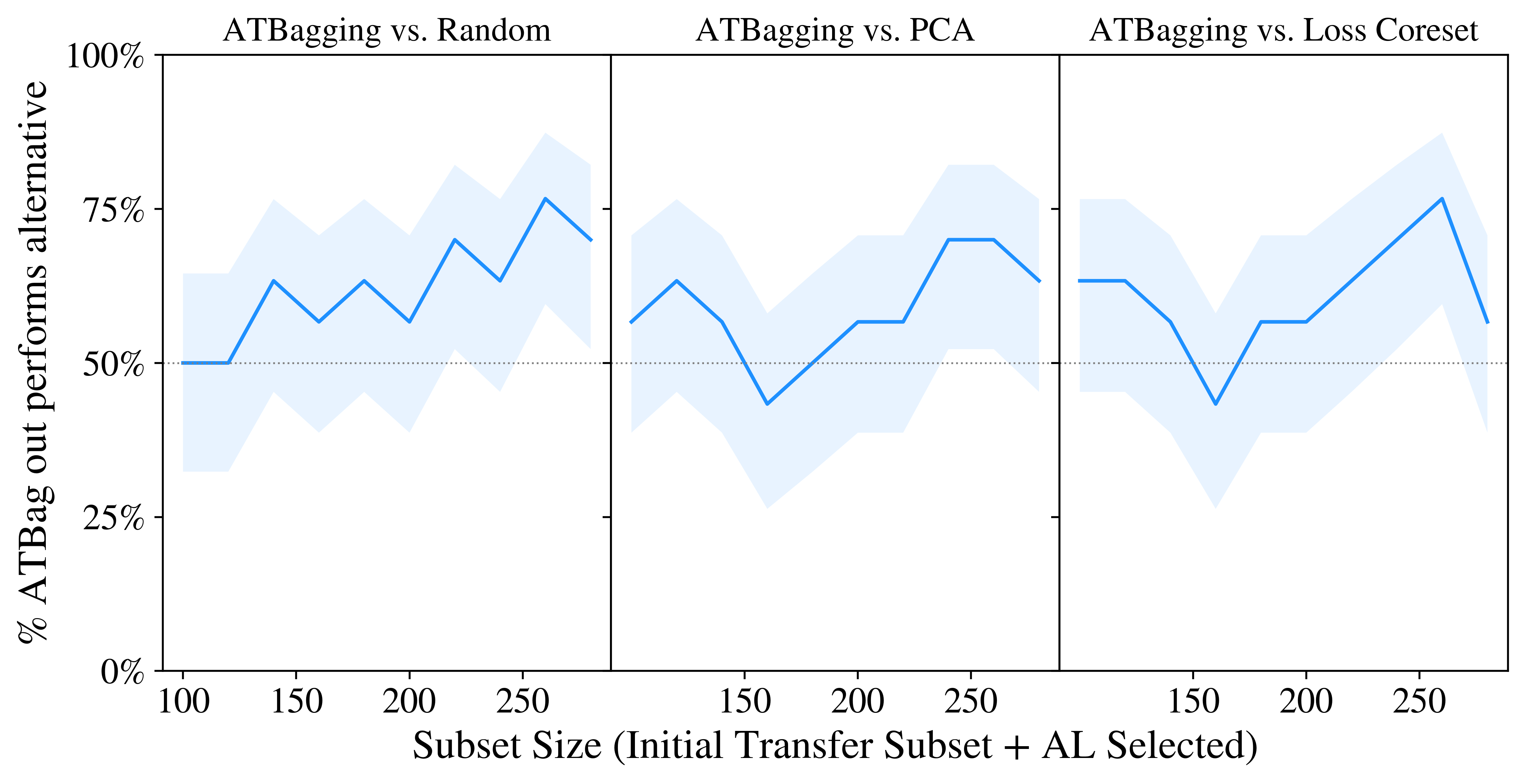}
        \caption{%
        PM\textsubscript{2.5} dataset transfer performance pairwise method comparisons
        with initial subset size of 100.
        }
        \label{fig:pm25_trans_perc_100}
    \end{subfigure}
\end{figure}

\subsection*{Transfer Performance -- Forbes: Subset Size 50}

\begin{figure}[H]
    \centering
    \begin{subfigure}[t]{0.49\textwidth}
        \centering
        \includegraphics[width=1.0\linewidth]{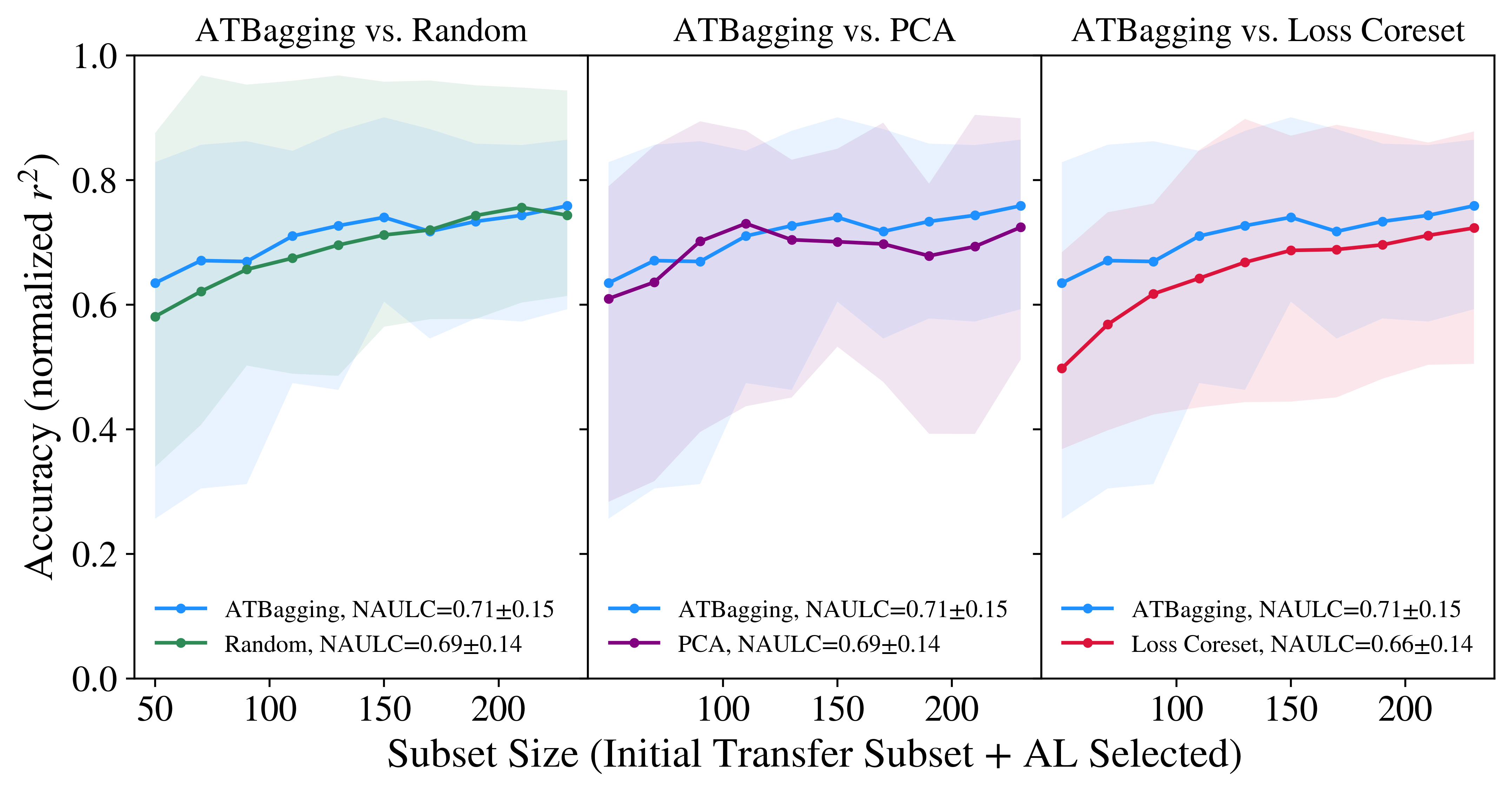}
        \caption{
        Forbes 2000 dataset transfer performance with initial subset size
        of 50.
        }
        \label{fig:forbes_trans_acc_50}
    \end{subfigure}
        \begin{subfigure}[t]{0.49\textwidth}
        \centering
        \includegraphics[width=1.0\linewidth]{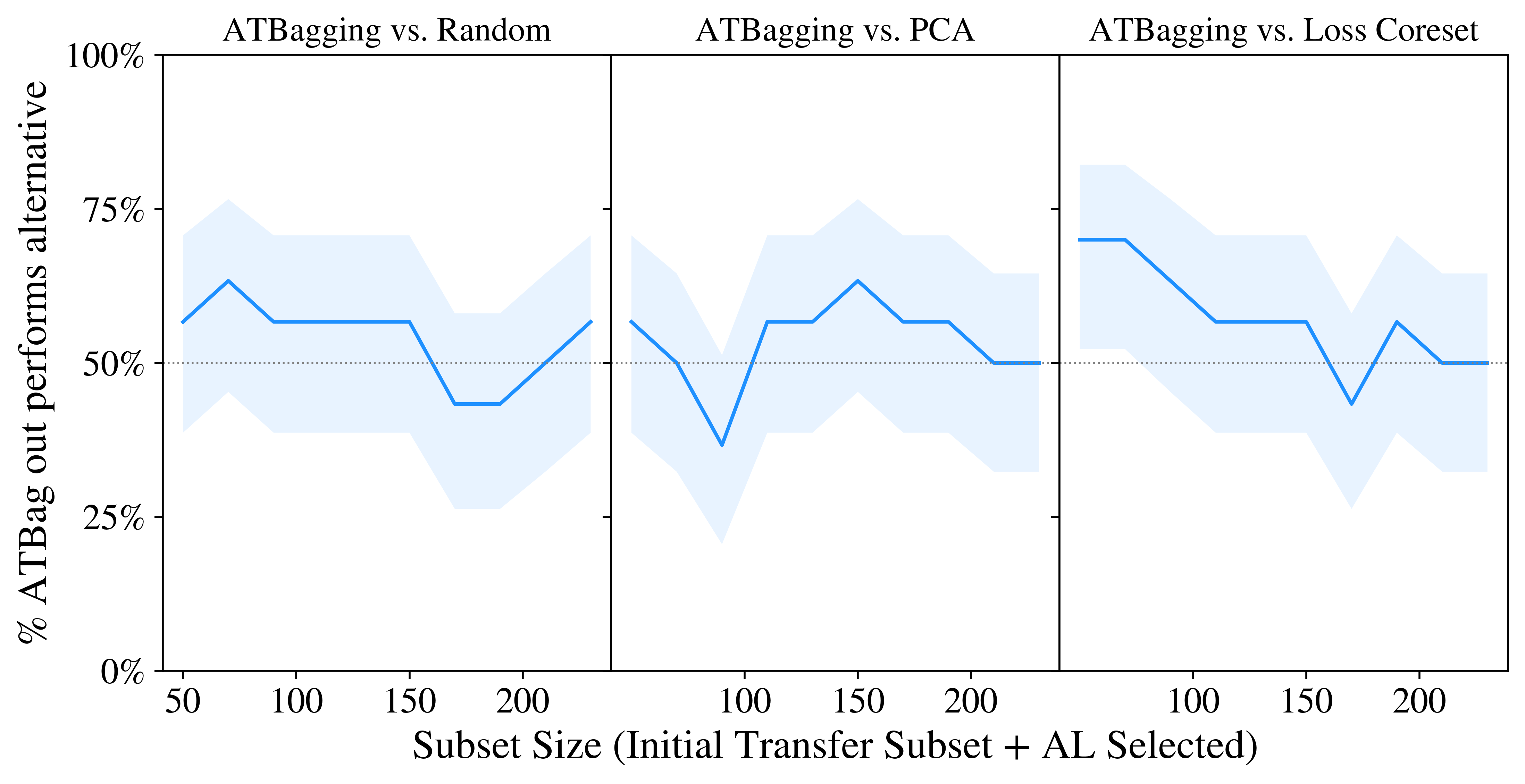}
        \caption{%
        Forbes 2000 dataset transfer performance pairwise method comparisons
        with initial subset size of 50.
        }
        \label{fig:forbes_trans_perc_50}
    \end{subfigure}
\end{figure}

\subsection*{Transfer Performance -- Forbes: Subset Size 100}

\begin{figure}[H]
    \centering
    \begin{subfigure}[t]{0.49\textwidth}
        \centering
        \includegraphics[width=1.0\linewidth]{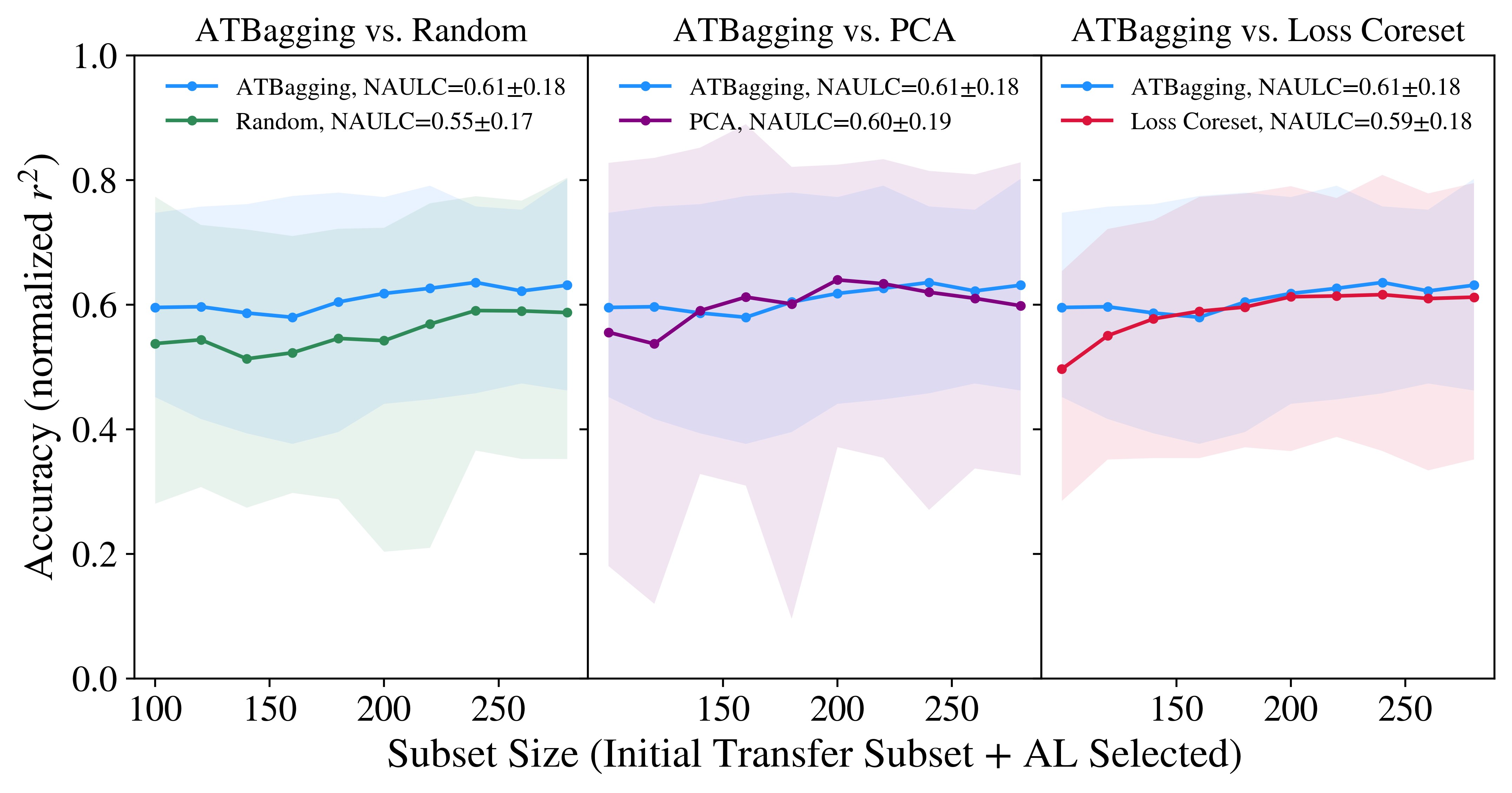}
        \caption{
        Forbes 2000 dataset transfer performance with initial subset size
        of 100.
        }
        \label{fig:forbes_trans_acc_100}
    \end{subfigure}
        \begin{subfigure}[t]{0.49\textwidth}
        \centering
        \includegraphics[width=1.0\linewidth]{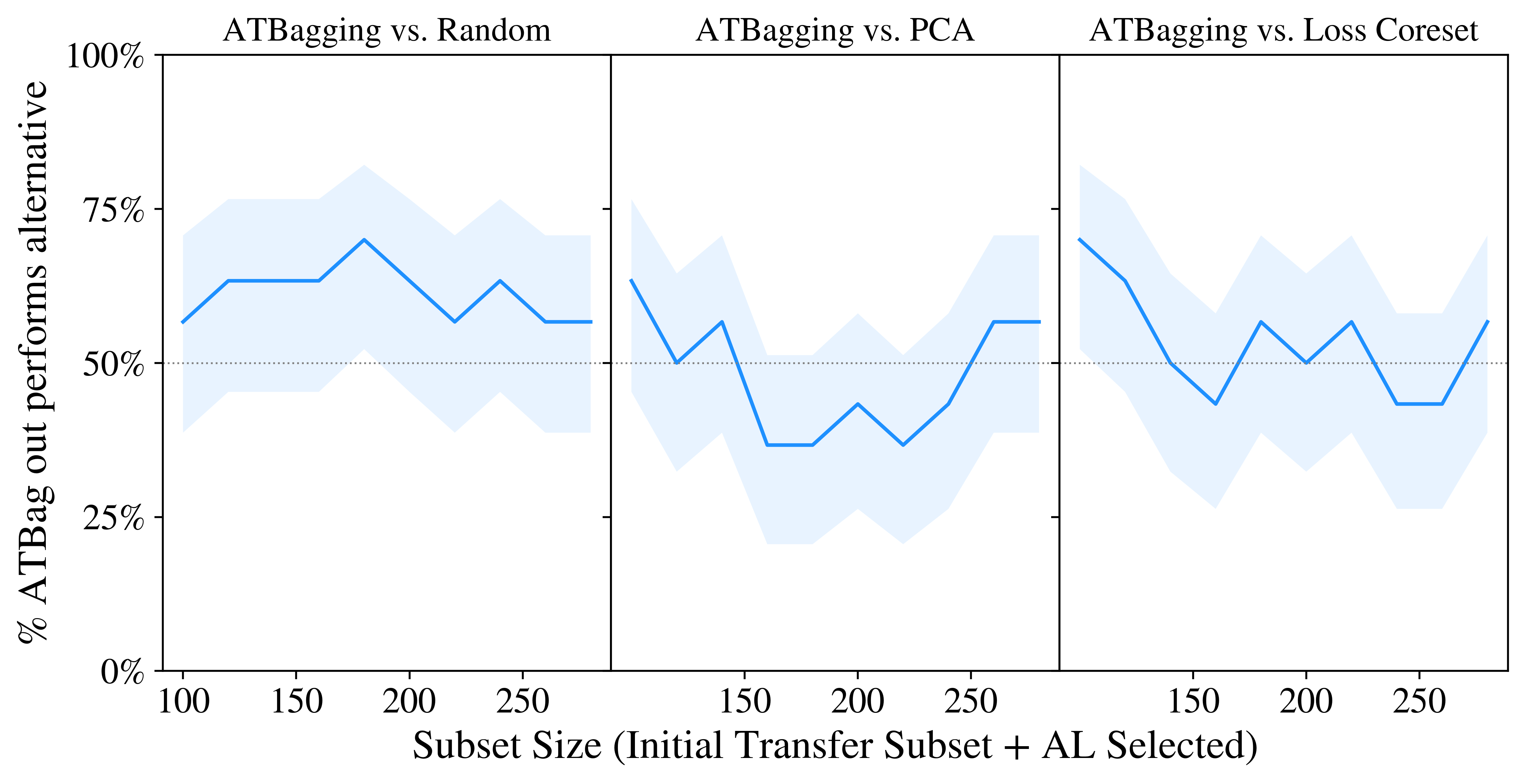}
        \caption{%
        Forbes 2000 dataset transfer performance pairwise method comparisons
        with initial subset size of 100.
        }
        \label{fig:forbes_trans_perc_100}
    \end{subfigure}
\end{figure}

\end{document}